\documentclass{article}

\usepackage{iclr2027_conference,times}
\iclrfinalcopy

\usepackage[utf8]{inputenc}
\usepackage[T1]{fontenc}
\usepackage{amsmath}
\usepackage{amssymb}
\usepackage{booktabs}
\usepackage{graphicx}
\usepackage{xcolor}
\usepackage[expansion=false]{microtype}
\usepackage{multirow}
\usepackage{caption}
\usepackage{etoolbox}
\usepackage{array}
\AtBeginEnvironment{longtable}{\scriptsize\setlength{\tabcolsep}{3pt}}
\usepackage{adjustbox}
\usepackage{longtable}
\usepackage{wrapfig}
\usepackage{pifont}
\usepackage{url}

\providecommand{\tightlist}{\setlength{\itemsep}{0pt}\setlength{\parskip}{0pt}}

\title{Localization Does Not Imply Steerability: \\ A Mechanistic Study of Sign Errors in LLM Mathematics}

\author{Shradha Agarwal\\
Department of Computer Science \&\\Department of Nuclear Engineering and Radiation Science\\
Missouri University of Science and Technology\\
\texttt{sabrc@mst.edu}
\And
Tariq J.\\
CMPP, NED University of Engineering\\and Technology, Karachi\\
Missouri University of Science and Technology
\And
Deepak Rajbhar\\
Missouri University of Science and Technology}

\begin{document}

\maketitle
\lhead{Preprint}

\begin{abstract}
When large language models do multi-step mathematics, a characteristic failure recurs: the model writes one wrong sign token mid-computation --- a cofactor sign, the minus of a recursive integration-by-parts step --- then executes the rest of the arithmetic correctly, and the answer comes out simply wrong. We ask where inside the network that sign is written. Across six open models spanning 14B to 675B parameters, a per-layer readout shows the wrong sign is committed late --- by MLP layers in the last quarter of the network (70--100\% per model and task), a pattern already present, where tested (Qwen), in the base model before instruction tuning. The mechanism varies: some models carry fixed ``minus-stamp'' layers that push a minus regardless of the right answer, others carry amplifiers of whichever sign was already chosen, and Qwen adds a downstream rescue layer that opposes the stamp. The circuit is causal: subtracting a single sign direction from the residual stream flips 13 of 14 held-out minus-written integration-by-parts errors on Llama-3.3-70B, with flips replicating on Qwen, Gemma, and Mistral-Small (24--70\% per cell), and with representation-level specificity --- the same dose breaks correctly-minus answers while sparing correct-plus answers where the direction is surgical (0/72 on Llama; Mistral, the outlier, breaks every cohort), and matched random-direction and control-layer interventions are null. Core claims were preregistered with pass criteria frozen in git before held-out data was seen: 23 of 29 checks passed, all misses reported. But causality is not uniform: Phi-4 implements an equally well-localized circuit at the same late depth, yet the identical intervention flips almost nothing --- the sign is written redundantly across $\sim$15 layers with margins several times larger (median 17.8 logits). Localization, we conclude, does not imply steerability: two models can share a circuit location yet differ enormously in how overturnable it is.

\end{abstract}

\section{Introduction}

\textbf{One wrong token.} Ask Qwen-2.5-72B for a 4\ensuremath{\times}4 determinant and watch it work. Every product is right. Every sum is right. Yet the model answers 323 when the truth is 83 --- because midway through, at exactly one position, it wrote \guillemotleft{}\ensuremath{-}\guillemotright{} where the algorithm demands \guillemotleft{}+\guillemotright{}, and every step after that mistake ran correctly on the corrupted expression. One character separates a perfect solution from a wrong one. This case, which we dissect throughout the paper, is no fluke: sign errors are structural, not noise --- the leading failure class wherever models still execute the algorithm {[}1{]}. That is the failure mode this paper studies: in multi-step mathematics, a single wrong sign token, mid-computation, that decides the final answer. A model that fails at exactly one token is not failing to reason; some specific part of the network wrote that token. This paper locates that part and tests it causally, in five model families across two tasks. (Beginner background: Appendix~A.1.)

\textbf{The question.} A central working assumption of mechanistic interpretability is a two-step program: first find \emph{where} in the network a behavior is decided, then intervene \emph{there} to change it. The first step is \emph{localization} --- naming the layers, heads, or directions that carry the decision. The second is \emph{steerability} --- showing that an intervention at the named site actually changes what the model does. In practice the two steps are rarely separated: a circuit is found, an edit at the circuit's site changes the behavior, and the localization is thereby declared ``validated.'' The unstated premise underneath this practice is that localization \emph{implies} steerability --- that knowing where a decision is written also tells you that you can change it there. This paper puts that premise to a test it cannot dodge: the same localization pipeline and the same intervention, applied identically and with preregistered predictions, across five models and two tasks. The implication does not hold: in four of the five models the identified direction is a usable lever, flipping up to 93\% of held-out errors, but localization cannot say in advance which model resists. Sign errors make the test fair: the place the behavior lives is one token --- one sign, at one identifiable character offset in the model's own text --- embedded in an otherwise-intact computation, an unusually clean target for circuit analysis.

\textbf{Contributions.} (i)~\emph{Localization:} in five model families and two tasks, the written sign is committed in the last quarter of the network --- errors and corrects alike, one fixed pipeline (\S{}3; Appendices~C--F). (ii)~\emph{Mechanism:} the identified layers split into fixed \emph{stamps} versus adaptive \emph{amplifiers} --- same localization, different mechanism (\S{}4; Appendices~B, F, G). (iii)~\emph{Causal battery:} single-layer, joint, and per-head ablations with random-set nulls; dose--response steering with a magnitude-matched control; a three-condition specificity test (\S{}5; Appendices~H--I). (iv)~\emph{The dissociation:} across ten model\ensuremath{\times}task cells, localization is uniformly high while steerability spans 4--93\% --- the best-localized model steers least (\S{}6, Figure 5). (v)~\emph{Preregistration:} frozen per-model predictions, pass bars, and denominators, scored once on held-out data, misses reported (\S{}6; Appendices~I, J).

\textbf{Related work.} Our tools are standard: the logit lens {[}2, 3{]} as readout, direct logit attribution in the transformer-circuits framework {[}4{]} for layer credit; the locate-then-intervene program we stress-test runs through causal mediation {[}5{]}, the indirect-object-identification circuit {[}6{]}, and mechanistic accounts of arithmetic {[}7, 8{]}. The closest prior is Hase et~al. {[}9{]}: causal-tracing localization does not predict where weight \emph{editing} succeeds. We extend the dissociation to activation-level steering, make it cross-model and cross-task, tie it to an identified mechanism difference (stamps versus amplifiers, redundancy, rescue layers), and preregister it --- noting the refinement cuts both ways: the direction is a working lever in four of five models, but only a causal test certifies it, and the predictive information sits in mechanism, not location. The redundancy we find echoes the self-repair of McGrath et~al. {[}10{]}; our intervention belongs to the activation-steering family {[}11--13{]}, whose implicit premise --- a well-localized direction is a usable lever --- is exactly what the Phi result bounds; the steering-reliability literature reports the same gap behaviorally {[}15{]}, and our per-model, mechanism-level predictor answers its call for predictors of steering success. Dataset methodology builds on LinAlg-Bench {[}1{]}.

\section{Models, tasks, and data}

\textbf{Models and tasks.} Five open model families appear in this paper: Gemma-3-27B, Mistral-Small-24B, Phi-4-14B, Llama-3.3-70B-Instruct, and Qwen-2.5-72B-Instruct. All five receive the identical treatment --- the full localization readout, the causal battery, and the preregistered evaluation --- on two tasks: 4\ensuremath{\times}4 matrix determinants (``det'', linear algebra) and integration by parts (``IBP'', calculus). That is five models \ensuremath{\times} two tasks: ten model\ensuremath{\times}task combinations, which we call \emph{cells}. A sixth, far larger model --- Mistral-Large-675B --- appears behaviorally only, in an unfiltered localization run on 5\ensuremath{\times}5 determinants (\S{}3; Appendix~C).

\textbf{What a sign error is.} The sites of the failure shown in \S{}1 are position-determined: in a cofactor expansion the term for entry \emph{a}\textsubscript{1\emph{j}} carries (\ensuremath{-}1)\textsuperscript{1+\emph{j}}; in integration by parts every recursive application of \ensuremath{\int}u\,dv~=~uv~\ensuremath{-}~\ensuremath{\int}v\,du plants a new minus inside the previous step. A sign error is a solution whose arithmetic is correct but in which at least one such sign enters wrong, flipping the final answer or an intermediate value. (Full background in Appendix~A: where sign errors live and why they matter in A.1; the problem banks, harvest funnels, and judge protocol behind every case in A.2--A.3.)

\textbf{Where the problems come from.} The determinant problems trace back to LinAlg-Bench {[}1{]} (open source, arXiv:2605.16675), a benchmark of SymPy-certified linear-algebra problems in which sign errors were first isolated as a structural failure mode: across 6,600 graded outputs, sign errors led every error class where models still execute the algorithm --- magnitude right, sign alone wrong --- and at 4\ensuremath{\times}4 their recursion-driven parity subtypes surge to 61\%, fixing our task size. The benchmark bank alone yields too few sign errors for causal testing, so each model received its own larger bank, generated with the benchmark's released generator to the same specification: integer entries in {[}\ensuremath{-}5, 5{]}, exact symbolic ground truth, deduplicated against all prior corpora. The more accurately a model computes, the more problems it takes to harvest its errors --- Gemma needed a bank of 4,050 problems, Qwen only 675 (full funnel in Appendix A.2, Table A1).

\textbf{The second task.} Integration by parts is not in the benchmark; we built that bank ourselves to test whether the findings extend beyond linear algebra. A custom generator produced 1,294 poly-trig integrals per model, each constructed so the parts formula must be applied at least once --- forcing the model to write its minus sign explicitly, the exact site of the IBP sign error (generator details in Appendix A.2).

\textbf{Harvesting the two pools.} Every model answered every problem in its bank at temperature 0, instructed to show all steps and box the final answer; one exact-match check against the symbolic ground truth sorts every response into an error pool and a correct pool. Wrong answers then pass the benchmark's released three-judge pipeline, which keeps a case only if the magnitude is right and the sign alone is wrong and sends ambiguous cases to a human (Appendix A.3); an annotator marks the exact character position of the wrong sign, and a deterministic check verifies it. Correct solutions need no judge: each is matched by \emph{kind} of sign --- probed at a cofactor or parts-formula operator sign, here written right --- located and verified by the same procedure (Appendix A.3.2).

\textbf{Two batches, four cohorts, three rules.} The data arrived in two batches: a discovery batch of 1,000 cases, used to choose the layers, the sign direction, and the pass bars, and a held-out batch of 2,253 new cases the discovery step never saw, reserved for the causal tests (screening and split protocol in Appendix A.4). Both batches contain both kinds of case: sign-error solutions and correct solutions. Every case is then tagged by the sign the model actually wrote at the probed position, \guillemotleft{}+\guillemotright{} or \guillemotleft{}\ensuremath{-}\guillemotright{}. Crossing the two labels gives the four cohorts every figure uses: error wrote-\guillemotleft{}\ensuremath{-}\guillemotright{}, error wrote-\guillemotleft{}+\guillemotright{}, correct wrote-\guillemotleft{}\ensuremath{-}\guillemotright{}, and correct wrote-\guillemotleft{}+\guillemotright{}. Table 1 reports each cell's totals, batch by batch, and unless a figure says otherwise, results combine both batches (per-batch accounting in Table C1, Appendix C, and Table E1, Appendix E). No stage of the funnel selects on the written sign, yet minus errors outnumber plus errors about two to one across the ten cells (1,037 vs 491; Table A4, Appendix A.4) --- a purely behavioral asymmetry that \S{}4 will locate mechanistically. Where circuit-level claims are commonly demonstrated on a single model and a few prompts, every claim here is cohort-level --- 3,253 cases across the ten cells, a bootstrap confidence interval behind every verdict --- and three rules govern the causal claims: every effect is read against matched random and negative controls, never against zero; every intervention is also run on correct answers, with failures reported at full strength; and the design is preregistered --- targets and pass bars frozen in git before any held-out case was scored, with held-out cells the headline numbers throughout (protocol in Appendix I).

\begin{wraptable}{r}{0.50\textwidth}
\vspace{-0.9\baselineskip}
\caption{The master data table. Each row is one model\ensuremath{\times}task cell (model versions: text above): read 218 (92+126) as 92 discovery + 126 held-out cases, and 63/79 as wrote \guillemotleft{}+\guillemotright{}\,/\,wrote \guillemotleft{}\ensuremath{-}\guillemotright{} in that batch; the causal tests are scored on the held-out batch.}
\label{tab:1}
\centering\footnotesize\setlength{\tabcolsep}{2.5pt}
\begin{adjustbox}{max width=\linewidth}
\begin{tabular}{@{}lllll@{}}
\toprule
Cell & Errors & Corrects & Disc. & Held \\
 & (disc+held) & (disc+held) & \guillemotleft{}+\guillemotright{}/\guillemotleft{}\ensuremath{-}\guillemotright{} & \guillemotleft{}+\guillemotright{}/\guillemotleft{}\ensuremath{-}\guillemotright{} \\
\midrule
Gemma det & 218 (92+126) & 188 (50+138) & 63/79 & 130/134 \\
Mistral-S det & 237 (95+142) & 200 (78+122) & 41/132 & 50/214 \\
Phi det & 211 (61+150) & 199 (50+149) & 17/94 & 36/263 \\
Llama det & 208 (16+192) & 210 (80+130) & 27/69 & 122/200 \\
Qwen det & 206 (59+147) & 188 (47+141) & 41/65 & 92/196 \\
Gemma IBP & 143 (70+73) & 200 (50+150) & 60/60 & 44/179 \\
Mistral-S IBP & 109 (39+70) & 199 (43+156) & 20/62 & 76/150 \\
Phi IBP & 117 (40+77) & 196 (50+146) & 13/77 & 61/162 \\
Llama IBP & 43 (22+21) & 70 (23+47) & 7/38 & 52/16 \\
Qwen IBP & 36 (10+26) & 75 (25+50) & 0/35 & 0/76 \\
\bottomrule
\end{tabular}
\end{adjustbox}
\vspace{-0.9\baselineskip}
\end{wraptable}
\textbf{Conditions of the experiments.} Every measurement in this paper, observational or causal, runs under one condition: \emph{teacher-forced replay} --- the model's own written solution replayed token-by-token, exactly as it wrote it, listening or intervening only at the \emph{error-site sign token}, the one mid-solution position where the wrong sign is written (on corrects, the matched position where the right sign is written). The replay keeps the crime scene untouched; the probe decides nothing. Two readouts and four interventions share this condition, all in one fixed pipeline of ten numbered stages s00--s09: arithmetic control (s00), readouts (s01--s02), target derivation (s03), projections (s04), interventions (s05--s08), judge tests (s09) --- identical for all five models, the s03 target list frozen and human-reviewed before any causal stage runs (stage definitions and complete per-stage results in Appendix B). The readouts: the logit lens (s01) decodes the residual stream into token scores at every layer, and the quantity tracked everywhere is the \emph{sign lead} --- the written sign's logit minus the alternative's; direct logit attribution (DLA, s02) reads each layer's own contribution to that lead, and late layers whose contribution tracks the true sign rather than the written one are checking layers (``judges''). The interventions --- ablation (s05), direction subtraction (s06), patching (s07), and steering (s08) --- are defined where they run, in \S{}5, with full protocols in Appendices H--K.
\section{The sign is decided late}

\textbf{What Figure 1 measures.} Each case is the model's own written solution, replayed as in \S{}2. In a determinant expansion, \textbackslash det(A) = \ensuremath{-}3\ensuremath{\cdot}\textbackslash det(A\_\{11\}) \ensuremath{-} (\ensuremath{-}4)\ensuremath{\cdot}\textbackslash det(A\_\{12\}) {[}\ensuremath{-}{]} \ldots, the bracketed sign is the error: the model wrote \guillemotleft{}\ensuremath{-}\guillemotright{} where the true sign is \guillemotleft{}+\guillemotright{}. We probe the token just before it --- here \guillemotleft{}\})\guillemotright{}, the close of \textbackslash det(A\_\{12\}), which the tokenizer holds as a single token (probe location is offset-verified per case; this very case is traced end to end in Appendix C.2). At that position the logit lens (s01) asks every layer the same question --- if the model had to answer from here, which sign? --- and records the written sign's lead, layer by layer. The layer where the lead is largest is the case's peak: where the network pushes the written sign hardest. Models differ in layer count, so peaks are compared as relative depth, layer / (n \ensuremath{-} 1): 0 the embedding, 1 the final layer. One forward pass per case, one peak per case, one point in Figure 1. Only confirmed cases enter --- final-layer lead \ensuremath{\geq} 2 logits; an excluded case is not evidence against a late decision, it carries no locatable decision (full recipe and accounting, Appendix C).

\begin{wrapfigure}{l}{0.58\textwidth}
\vspace{-0.9\baselineskip}
\centering
\includegraphics[width=\linewidth]{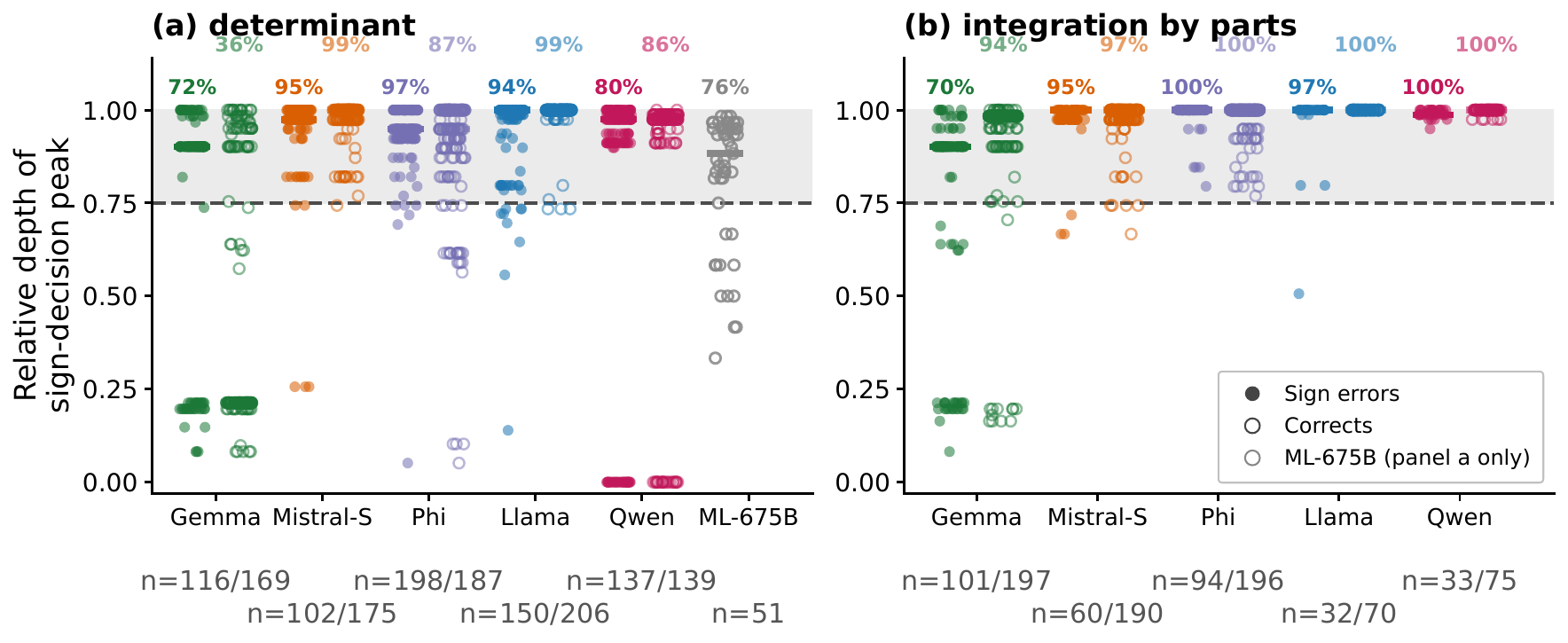}
\caption{The sign is decided late --- in both tasks: (a) determinant, (b) integration by parts, measured identically. One point per confirmed case's peak depth, both batches pooled; bar: median; shaded band: late window (depth \ensuremath{\geq} 0.75); percentages: late share (bold: errors, solid points; lighter: corrects, open points); n = errors/corrects per panel. Exclusions: Table C1. Grey circles, (a) only: unfiltered Mistral-Large-675B run (\S{}3).}
\label{fig:1}
\vspace{-0.7\baselineskip}
\end{wrapfigure}
\textbf{Two kinds of peaks.} Figure 1 shows where the decision peaks, and in both panels the answer is the same: the great majority of points sit in the shaded band, past depth 0.75. Determinant: 72--97\% of confirmed errors per model (Gemma 72\%, Qwen 80\%, Llama 94\%, Mistral-Small 95\%, Phi 97\%); IBP: 70--100\%. Correct answers commit equally late, so lateness characterizes sign-writing itself, not failure --- with one exception: Gemma's determinant corrects decide early (36\% late, median depth 0.21), and even that vanishes on IBP (94\% late). An exploratory behavioral run on a sixth, far larger model --- Mistral-Large-675B, 51 sign errors on the 5\ensuremath{\times}5 task, unfiltered, the causal battery computationally prohibitive at that scale (accounting in Appendix C) --- is consistent: 39 of 51 (76\%) are late. The arithmetic control closes the loop: 48 plain subtraction problems, answered almost always correctly (237 of 240 completions), every case confirmed --- and four of the five models still peak past depth 0.75, with Gemma again the exception (22 of 48 late, split bimodally between an early band and the end) --- so late sign-writing belongs neither to the algebra tasks nor to errors (Appendix D). The figure also leaves a small early minority, concentrated in two models, taken up next.

\textbf{The early minority.} The peak alone can mislead --- a curve can spike early, collapse below zero through the whole middle of the network, and commit only at the very end --- so Appendix C asks a second question of every early case: from which layer does the lead turn positive and stay positive (the crossing, Appendix C.4)? Qwen's early points are all artifacts of a token-table lookup at layer 0: its dominant probe token leans \guillemotleft{}\ensuremath{-}\guillemotright{} before any layer has run, the lean collapses mid-network, and every such case locks in its sign only at 95--99\% depth. Gemma's are the real exception: an early band (layers 5--13) copies the minus ink of the model's own recent text --- the source of its corrects exception in Figure 1 --- and it vanishes on IBP. Across all five models, genuinely early error commitments are 26 of 703 confirmed cases: 3.7\% --- the limit of the late-decision claim, stated plainly (case-by-case census in Table C2, Appendix C).

\textbf{The full path.} The complete median trajectories (Appendix E, Figures E1--E2) seal the story: in every model the lead sits inside the noise band, or below zero, through the middle of the network and exits steeply only past depth 0.75. An inherited error, computed earlier and carried forward, would be visible mid-network; nothing is.

\section{Habit layers: which layers write the sign, and how}

\textbf{From when to where.} \S{}3 located the sign decision in time; this section locates it in the architecture. The instrument is per-layer direct logit attribution (DLA): each layer's MLP output projected onto the final-layer probe direction --- its direct contribution, in logits, to the written sign's lead (protocol: Appendix B, stage s02; pooled verdict table: Appendix E.6). MLPs are the unit of account, not a verdict on attention: attention is tested causally in \S{}5, where no single head accounts for the push (per-head grid: Appendix G). In every model the credit concentrates on a handful of late layers --- the \emph{habit layers}: Gemma L58/L60, Mistral L37--L39, Phi L37/L39, Llama L77/L79, Qwen L75/L78 --- frozen on discovery data (Appendix I.2, Table I.1), one set per model, the same for errors and corrects, because targets name layers, not outcomes. Figure 2 pools both phases, so a tall pooled bar is not the selection criterion (Phi's L34 --- visible in the pooled view, outside the frozen set --- is the clearest example; full-depth view: Appendix E, Figures E3/E4). The reading is simple: a handful of named late layers write the sign --- Mistral's L37--L39 in the last quarter of 40 layers, Llama's L77/L79 in the last few of 80 --- and the error and correct bars rise at the same layers: the same pen writes both the right and the wrong sign. These named layers are the exact targets of every causal experiment in \S{}5. One column already breaks the pattern: Phi's credit is not a handful of spikes but a nearly even band across its whole late window --- the spread that \S{}5 and \S{}6 show is exactly why its circuit resists steering (Appendix~K).

\begin{figure}[t!p]
\centering
\includegraphics[width=0.78\textwidth]{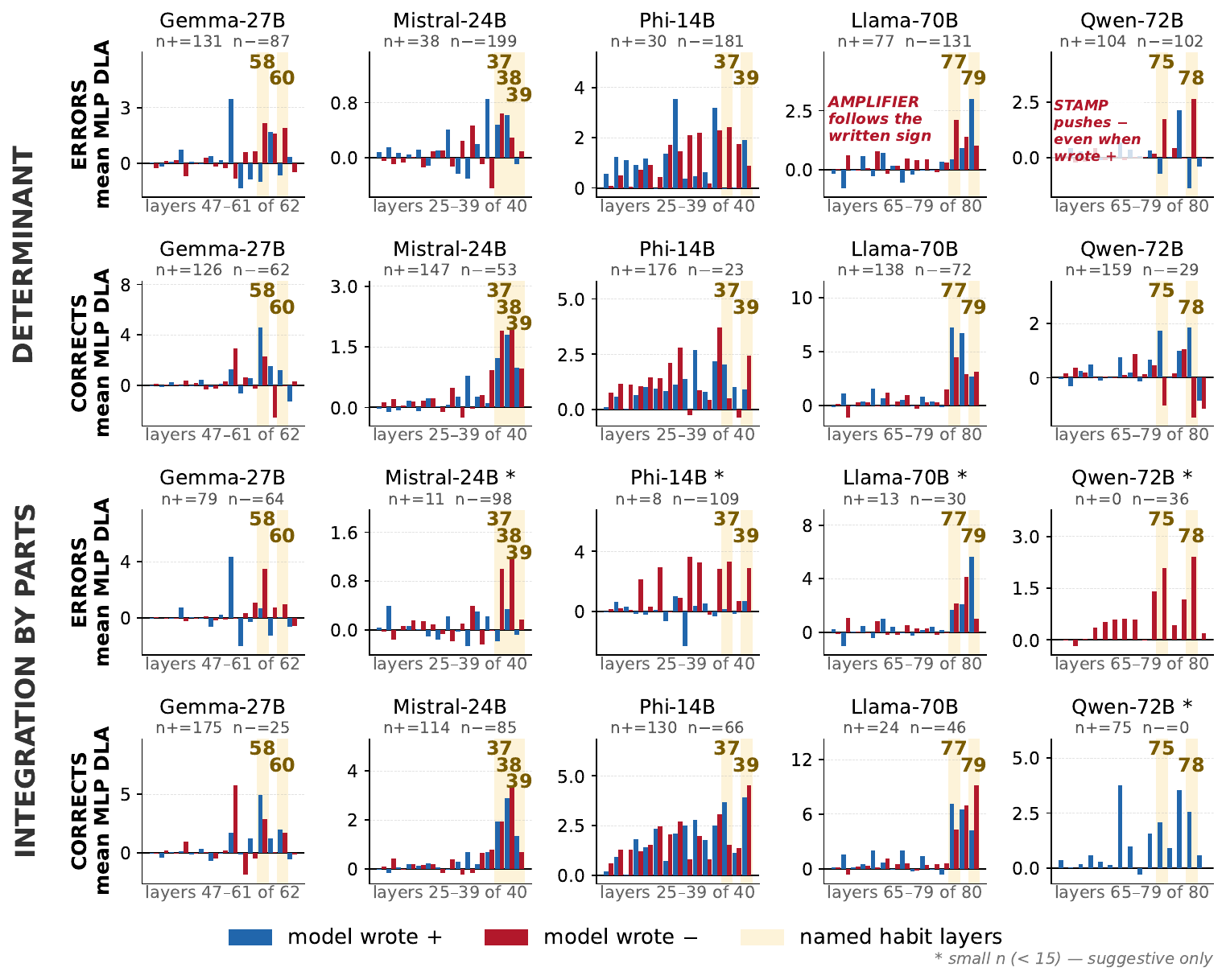}
\caption{Per-layer MLP DLA split by written sign --- determinant (top) and IBP (bottom), all five models, both batches. Blue: wrote \guillemotleft{}+\guillemotright{}; red: wrote \guillemotleft{}\ensuremath{-}\guillemotright{}; gold bands: habit layers. Same-direction bars in both groups: amplifier; opposite: stamp; a late layer pushing back toward the correct sign on errors is a judge. * small n (\textless{} 15), suggestive only (Mistral/Phi/Llama IBP \guillemotleft{}+\guillemotright{} errors, n = 11/8/13); Qwen IBP has no blue bars (0/36 errors, 0/75 corrects) \textbf{--- a property of the model, not the data: Qwen never writes \guillemotleft{}+\guillemotright{} at the IBP sign site.}}
\label{fig:2}
\end{figure}

\begin{wraptable}{r}{0.62\textwidth}
\vspace{-0.9\baselineskip}
\caption{Layer roles at a glance --- verdict first, evidence in brackets. Each cell: the verdict and the mean MLP DLA (logits) toward the written sign on wrote-\guillemotleft{}+\guillemotright{} cases, pooled discovery + held-out. CI fully below 0 = stamp; fully above 0 = amplifier; crossing 0 = indeterminate; * = n \textless{} 15, suggestive only. Full 95\% bootstrap CIs, case counts, and the wrote-\guillemotleft{}\ensuremath{-}\guillemotright{} sanity columns: Table E2 (Appendix E.6); held-out-only preregistered scoring: Appendix I.}
\label{tab:2}
\centering\footnotesize\setlength{\tabcolsep}{3.0pt}
\begin{adjustbox}{max width=\linewidth}
\begin{tabular}{@{}lllll@{}}
\toprule
\textbf{Model (layer)} & \textbf{Det errors} & \textbf{Det corrects} & \textbf{IBP errors} & \textbf{IBP corrects} \\
\midrule
Gemma (L58) & \textbf{stamp (-0.98)} & amplifier (+2.30) & amplifier (+0.66) & amplifier (+2.88) \\
Mistral (L39) & indet. (-0.14) & weak amp. (+0.99) & indet.* (-0.28) & weak amp. (+0.32) \\
Phi (L39) & amplifier (+1.56) & amplifier (+2.12) & amplifier* (+0.62) & amplifier (+3.89) \\
Llama (L79) & amplifier (+2.99) & amplifier (+3.38) & amplifier* (+5.66) & amplifier (+8.93) \\
Qwen (L75) & \textbf{stamp/judge (-0.72)} & \textbf{stamp (-1.02)} & --- no \guillemotleft{}+\guillemotright{} cases & --- no \guillemotleft{}+\guillemotright{} cases \\
\bottomrule
\end{tabular}
\end{adjustbox}
\vspace{-0.9\baselineskip}
\end{wraptable}
\textbf{Stamps versus amplifiers: the diagnostic.} Naming the layers is not the end of the question: two models can write the sign at the same relative depth by different means, and the distinction --- a fixed \emph{stamp} versus an adaptive \emph{amplifier} --- is not cosmetic: it predicts how each model responds to intervention in \S{}5. The push Figure 2 measures travels along a single \emph{sign direction} in the residual stream (construction in Appendix H.2; validation, the random-direction control, in H.3). Each bar is the mean DLA contribution of one layer to the written sign's lead, computed separately for the wrote-\guillemotleft{}+\guillemotright{} and wrote-\guillemotleft{}\ensuremath{-}\guillemotright{} groups; because the lead is a contest between exactly two tokens, a downward bar is not ``indifference'' --- it is an active vote for the \emph{opposite} sign, in the very cases where the model wrote this one. The diagnostic is a single comparison per layer: bars pointing the same way in both groups mark an \emph{amplifier} --- the layer supports whatever the model is writing; bars pointing opposite ways mark a \emph{stamp} --- one fixed sign, whatever was written.

\textbf{The verdicts.} Table 2 condenses the one-comparison rule into a verdict per model and domain (full bootstrap CIs, case counts, and sanity columns in Table E2, Appendix E.6). \emph{Qwen-2.5-72B:} L75 and L78 push \guillemotleft{}\ensuremath{-}\guillemotright{} no matter which sign the model is writing --- clean fixed-minus \emph{stamps}, active even on the correct plus answers it gets right; those survive because L77, a \emph{rescue layer}, pushes hardest exactly for the plus-writers: Qwen errs when its standing minus default goes unopposed. \emph{Llama-3.3-70B:} at L79 both groups' bars point up --- a pure \emph{amplifier}, supporting whichever sign the model has already decided. \emph{Gemma-3-27B}'s det L58/L60 lean stamper-like and \emph{Phi-4-14B} is mixed (L37 stamp-like, L39 amplifier-like): the models do not share one mechanism --- localization is universal, the mechanism is not. The corrects confirm it (a downward bar on a correct answer is the strongest evidence of a stamp, and the stamp layers show exactly that), and the IBP panels of Figure 2 replay the comparison in the second domain.

\textbf{Only minus gets stamped.} One regularity in Figure 2 is an absence: if stamps were a generic failure mode, some model somewhere should carry a fixed \emph{plus} stamp, and none does --- wherever the two bars disagree, it is always the wrote-\guillemotleft{}\ensuremath{-}\guillemotright{} bar that points up, so every fixed stamp at the habit layers pushes \guillemotleft{}\ensuremath{-}\guillemotright{}, never \guillemotleft{}+\guillemotright{} (the one plus-leaning bar, Gemma's untagged L55, sits outside the habit set and is outvoted by its minus-pushing neighbours). The asymmetry has a plausible origin in written mathematics --- a negative result is marked explicitly (``\ensuremath{-}67'') while a positive one is written bare, so text statistics teach a model to expect minus --- and three independent fingerprints support the reading: the error pool's own two-to-one minus skew (Table A4, Appendix A.4), Qwen's layer-0 embedding lean (Appendix C.4.3), and a base-versus-instruct comparison finding the L78 minus stamp already present before any chat training (Appendix F). The practical consequence shapes everything downstream: since only minus habits exist where the circuit acts, the steering experiment of \S{}5 (Figure 3) is a minus-removal design --- one can only erase a habit that exists.

\textbf{Roles are properties of the layers, not of error cases.} The correct panels are already in Figure 2; Table 2 condenses the verdict test on correct answers --- same layers, same rule, both domains. Qwen's minus-stamp, Llama's L79 amplifier and Phi's L39 amplifier all appear on corrects too. Mistral's errors are indeterminate; its corrects lean amplifier, weakly. One divergence: Gemma's determinant L58 --- stamper on errors, amplifier on corrects; its determinant corrects commit early (median depth 0.21), before L58's window. Across domains, the dependence is selective: Gemma L58 changes role; Phi L39 and Llama L79 do not. Qwen's role test is not computable on IBP --- no wrote-+ cases in either class. What separates errors from corrects is the sign the machinery is applied to, not the machinery. And the timing never moves: in both domains, both classes commit in the last few percent of depth. Localization claims must name the task; the late-commitment claim need not.

\section{Causal tests: the layers write the sign, redundantly --- and steering it is another matter}

\S{}3 and \S{}4 located the sign decision and typed its mechanisms; both rest on correlational evidence. This section runs the interventions declared in \S{}2, under the same replay condition as every measurement so far, in two directions: necessity --- does removing the habit layers' contribution destroy the sign? --- and sufficiency for control --- does subtracting the sign direction change what the model writes? The targets were frozen before held-out scoring (\S{}2), so the interventions are valid on both halves of the data; we report both, with held-out as the confirmatory half.

\textbf{Necessity: ablation.} DLA credit is correlational, so we ablate. No single writer is necessary: removing the sign-direction component from any \emph{one} layer is near-null except at Llama's final amplifiers, and zeroing candidate attention heads one at a time costs at most 0.46 logits --- attention is not the carrier (Appendix G). The \emph{set} is necessary --- in some models: removing the five habit layers jointly collapses the wrong sign's lead in Llama (det \ensuremath{-}4.25, IBP \ensuremath{-}7.09 held-out), Qwen det (\ensuremath{-}2.52), and Gemma on discovery, with the three large held-out collapses each exceeding all 40 matched random five-layer sets (protocol and full grid: Appendix G). Near-null single layers plus a large joint effect is the signature of \emph{redundant writing}: several layers push the sign along the same direction, and any one can be removed while the rest compensate --- the self-repair of McGrath et al. {[}10{]}, confirmed at scale by {[}16{]}, here preregistered and cross-model. The split has a second face: in Mistral and Phi the joint ablation is near-null even on discovery, and ablating Phi's entire late window L24--39 moves its det lead by only \ensuremath{-}1.28, \emph{and cumulatively mean-ablating its top fifteen DLA layers --- every credited writer --- reaches just \ensuremath{-}1.34 against a median margin of 17.8 (Appendix~K)} --- a necessity failure that foreshadows \S{}6. A stronger combined removal (the \ensuremath{\alpha}~=~1 arm of the protocol below: subtract the sign direction at the habit-layer MLPs, zero the amplifier attention) bites in all five models on held-out data (full grid: Table G7, Appendix G.7) --- but Phi's large drops move the lead without flipping the written sign: its oversized margins are part of the story of \S{}6.

\textbf{Sufficiency: steering.} If the habit layers write the sign along one direction, subtracting that direction at those layers should \emph{change what the model writes}. Figure 2 showed that every fixed stamp at the habit layers pushes ``\ensuremath{-}'', so the intervention is minus-removal: subtract the sign direction \ensuremath{\hat{d}} (construction in Appendix H.2) at each model's habit-layer steering sites (listed per model in Appendix H.4) at doses \ensuremath{\alpha} = 1, 2, 3, on minus-written error cases from both phases (pooled; Figure 3), against CTRL\_MAG --- a magnitude-matched random direction that separates direction-specific effects from generic disturbance.

\begin{figure}[t!p]
\centering
\begin{minipage}[t]{0.485\textwidth}
\centering
\includegraphics[width=0.84\textwidth]{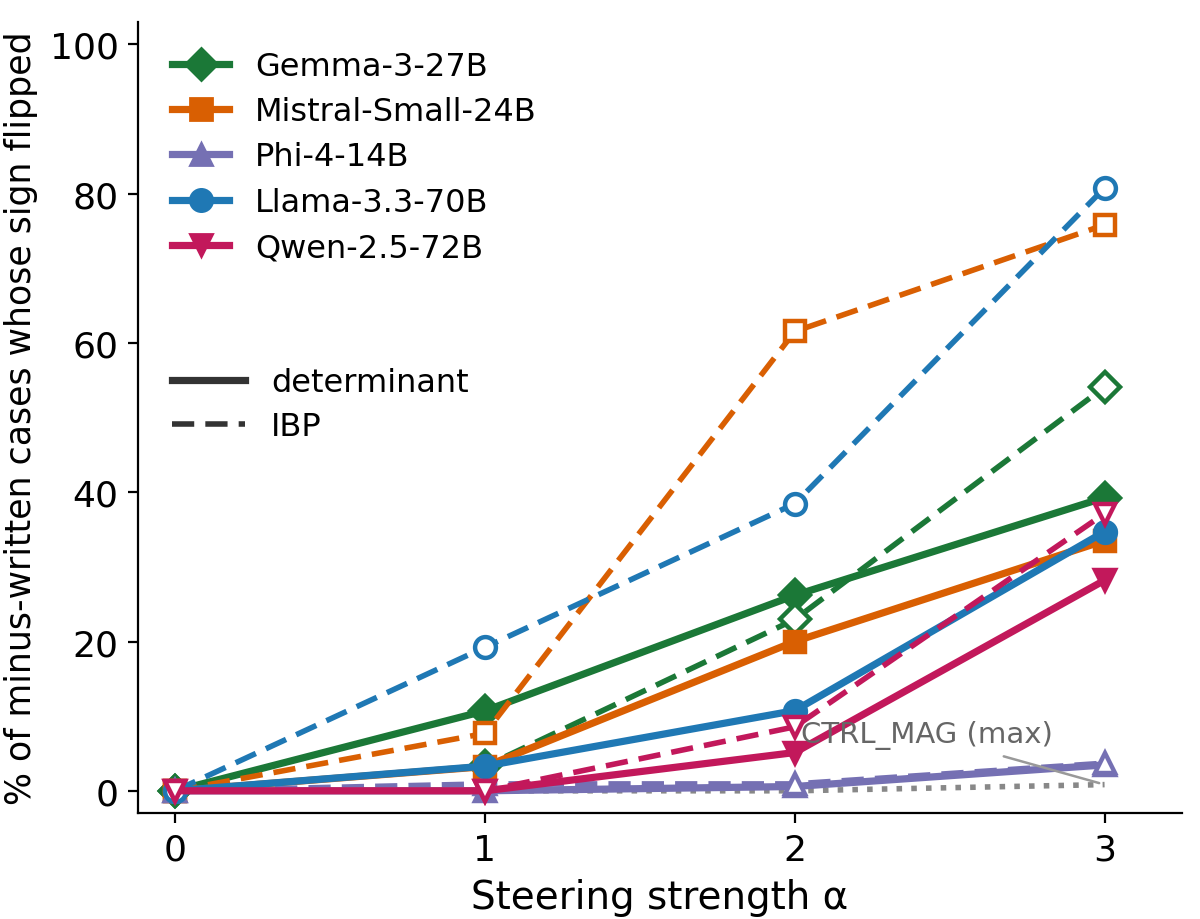}
\captionof{figure}{\textbf{Minus-removal steering --- dose--response, both domains in one panel (C2, pooled discovery + held-out).} Solid: determinant; dashed: IBP; one color per model; grey dotted: CTRL\_MAG, a magnitude-matched random direction (max across models). Subtraction of \ensuremath{\hat{d}} at the habit-layer sites (protocol: Appendix H; Llama det held-out only). The \ensuremath{\alpha} = 3 endpoints (k/n) are the top row of Figure 4. Not every flip is a repair: Llama det 42 = 33 corrective + 9 harmful; Mistral det 52 + 10, IBP 67 + 2; all other flips corrective.}
\label{fig:3}
\end{minipage}\hfill
\begin{minipage}[t]{0.485\textwidth}
\centering
\includegraphics[width=0.84\textwidth]{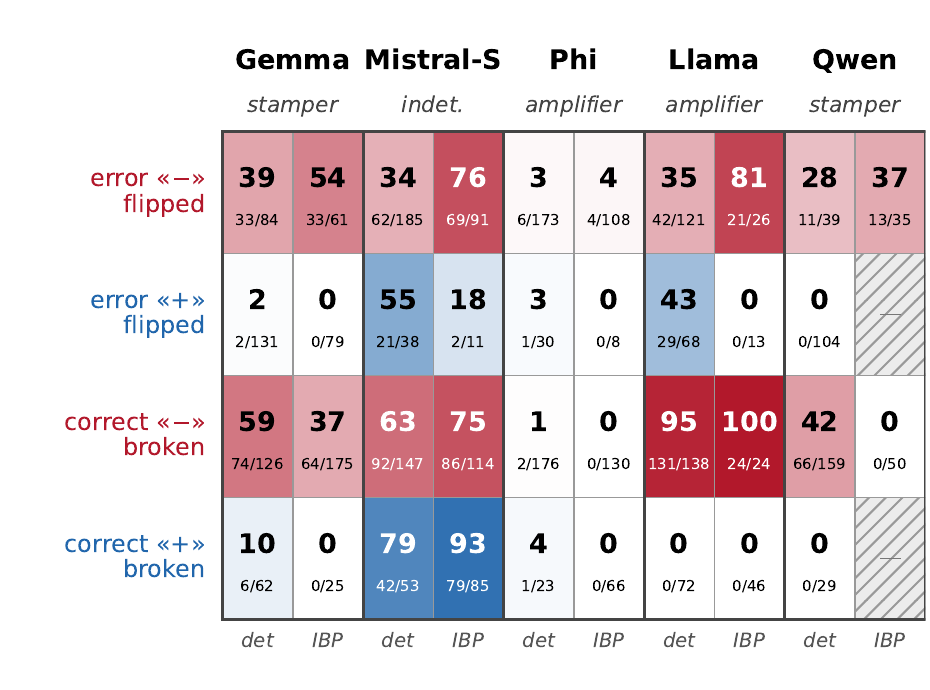}
\captionof{figure}{\textbf{The specificity fingerprint --- what minus-removal at \ensuremath{\alpha} = 3 does to the four cohorts, per model.} Each cell split: left det, right IBP; large number = \% flipped (errors) or broken (corrects), small = k/n. Red rows: minus-written; blue: plus-written. Hatched: cohort does not exist (Qwen wrote no \guillemotleft{}+\guillemotright{} in IBP). Prediction and verdicts: text (lead shifts: Appendix H.4; held-out-only rates: Appendix I).}
\label{fig:4}
\end{minipage}
\end{figure}

\textbf{The dose--response.} Each line is one model on one task; moving right along the x-axis means subtracting the minus direction more strongly. Read it in three steps. First, the grey dashed floor: a random direction of exactly the same size flips 1 of 923 cases in the entire pooled experiment --- whatever the colored lines do above it is specific to the minus direction, not to disturbance in general. Second, the slopes: four of the five models rise with dose --- removing the minus habit fixes wrongly-written minus signs, and fixes more of them the harder it is removed; Mistral climbs to 69/91 on IBP and Llama to 21/26. The IBP task responds more strongly than the determinant in every model. Third, the exception: Phi barely moves on either task (6/173 det, 4/108 IBP) --- the same intervention, at equally well-localized layers, produces almost no causal purchase. That gap is the paper's central dissociation, and Figure 5 plots it directly. Why Phi resists is itself a measured result, not a mystery: oversized error-site margins (det median 17.8 logits) and writing spread over $\sim$15 late MLP layers (visible directly in Phi's column of Figure 2) mean even per-case calibrated dosing flips only about half the cases --- Phi is unsteerable \emph{at calibrated dose} (exploratory deep-dive: Appendix~K).

\textbf{What the lever moves: minus-ness, not errors.} A flip rate is only interpretable next to a specificity test: does the intervention repair \emph{wrongness}, or remove \emph{minus-ness} wherever it lives? Applying the identical direction and dose to the correct answers settles it, and Figure 4 is that test drawn as a picture: one column per model, one row per cohort --- the two red rows minus-written, the two blue rows plus-written. Figure 2 turns the stamp/amplifier split into a row prediction: removing a standing \guillemotleft{}\ensuremath{-}\guillemotright{} vote should color the red rows and leave the blue rows empty; removing an amplifier should spill into every minus row, correct or wrong. At \ensuremath{\alpha} = 3 the matrix obeys: in Llama's column the correct-minus row breaks at 95\% (131/138) while the correct-plus row stays blank (0/72); in Qwen's, 42\% of correct-minus (66/159) and 0/29 correct-plus; CTRL\_MAG colors nothing anywhere. That row pattern --- fix wrong minus, break right minus, spare right plus --- is the signature of a direction whose content is ``this answer is minus, with no component separating correct-minus from wrong-minus: steering acts on representation content, not correctness (full protocol, per-dose tables, and the bridge to all five models: Appendix H).

\textbf{What each model is} (Figure 4, column by column). \textbf{Qwen --- the purest stamp:} a fixed \guillemotleft{}\ensuremath{-}\guillemotright{} vote even on corrects; every det-error flip corrective (11/11), none of 104 plus-written cases touched; wrote no \guillemotleft{}+\guillemotright{} in IBP at all. \textbf{Gemma --- stamp on errors, amplifier on corrects:} just as surgical --- the minus cohorts collapse, plus-written survive 129/131 with growing \guillemotleft{}+\guillemotright{} leads. \textbf{Llama --- the pure amplifier:} a minus-ness lever, not a repair --- 43\% of plus-written det errors flip too, 95--100\% of correct-\guillemotleft{}\ensuremath{-}\guillemotright{} break, correct-\guillemotleft{}+\guillemotright{} spared (0/72). \textbf{Mistral --- grips everything, specific to nothing:} weakest verdict in Table 2, strongest steering in Figure 3, color in every cohort (correct-\guillemotleft{}+\guillemotright{}: 42/53). \textbf{Phi --- the dissociation in one model:} best-localized, an amplifier by DLA, immune to the lever --- the preregistered null (Appendix I).

\textbf{The reverse arrow: injection.} The patching stage (s07) and prediction P5 ask the opposite question: at the moment before the sign is written, swap the message one layer passes forward --- replace its activation at that single token with a mean message recorded from other runs --- and see which sign the model writes now. The pattern across every arm is one line, and all four scored models confirmed their registered predictions: handing an error the healthy message from correct runs repairs almost nothing (a preregistered null --- at the stamp layers the healthy and the error message are the same, so there is nothing error-specific to copy in), while handing a correct answer the error message breaks it (Mistral 79/107 at \ensuremath{\alpha} = 3; Qwen's check layer reads the transplanted content exactly as registered; Llama's margins absorb the lead shift, 0/130 broken; Phi is immune in this arrow too, which rules out a miscalibrated dose or direction and leaves the mechanism itself as the explanation). The signal can be written in, not copied out; controls stay flat throughout (mid-network swaps and CTRL\_MAG \ensuremath{\approx} 0); per-model protocols and full tables are in Appendix J, the frozen verdicts in Appendix I.

\section{The dissociation, preregistered}

\begin{wrapfigure}{l}{0.42\textwidth}
\vspace{-12pt}
\centering
\includegraphics[width=\linewidth]{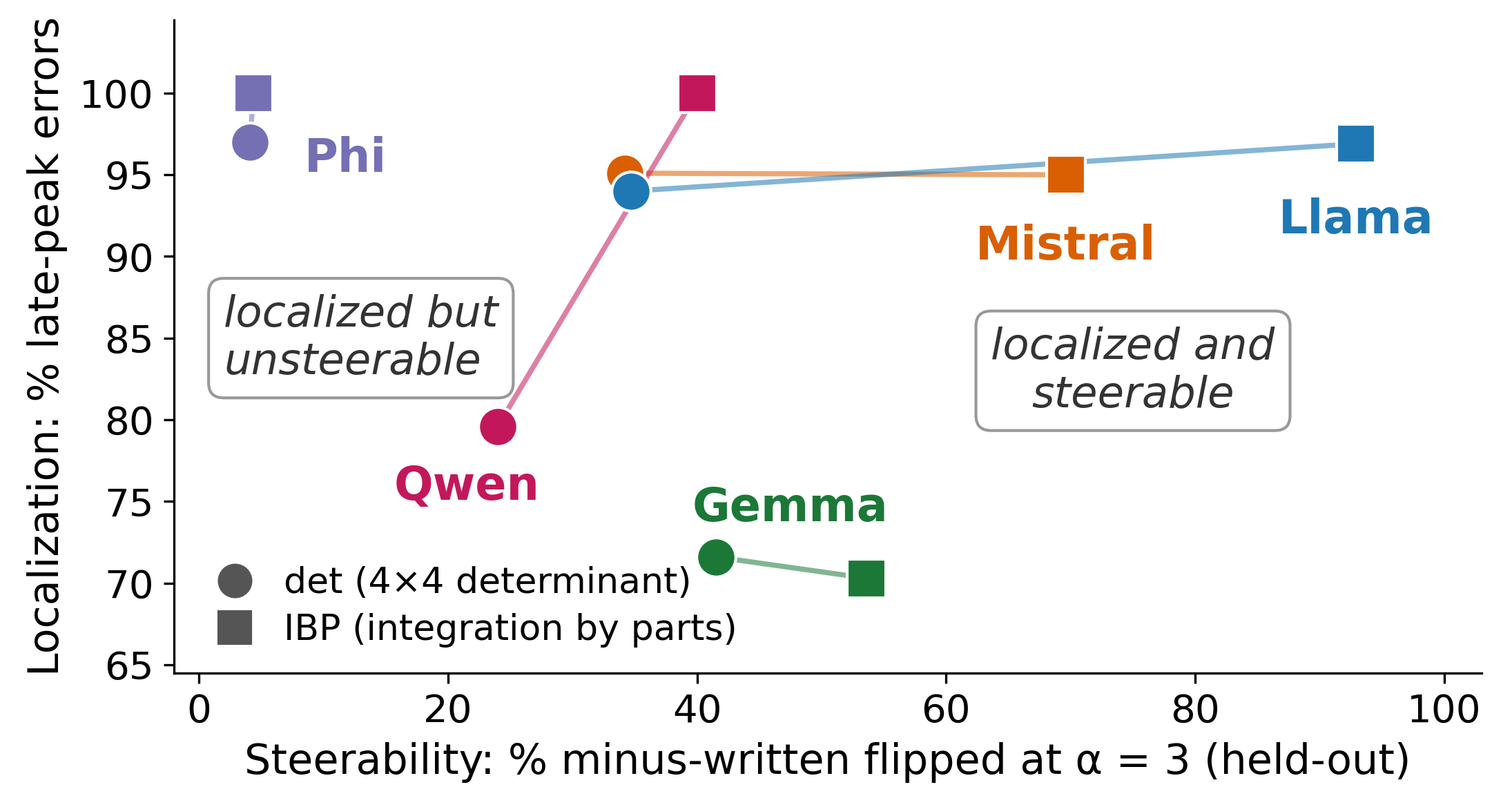}
\caption{\textbf{The dissociation --- localization does not imply steerability.} Two points per model, joined by a thin line; axes and markers as labeled. Filters and exclusions match Figures 1 and 3; a flip is not always a repair (corrective/harmful split in the Figure 3 caption).}
\label{fig:5}
\vspace{-10pt}
\end{wrapfigure}

\textbf{Two axes, one verdict.} Figure 5 plots, for each of the ten model\ensuremath{\times}task cells, localization --- the percent of confirmed errors peaking in the last quarter of depth --- against steerability --- the percent of held-out minus-written cases flipped at \ensuremath{\alpha}~=~3. If localization implied steerability, the points would climb a diagonal. They do not: all ten cells sit high on localization (70--100\%) yet scatter across nearly the full width on steerability (4--93\%), and Phi breaks the implication outright --- the sharpest localization of all five models (97\% det, 100\% IBP) and the least steerable (4.1\% and 4.4\%), on both tasks at once. The thin lines add one more fact: IBP is more steerable than the determinant for every model, yet Phi's pair barely moves --- the task effect cannot rescue a model that resists. Every held-out number was scored under criteria frozen in advance: six predictions per model (P1--P6) with numeric pass bars, exclusions, and denominators fixed before any held-out data existed (freeze dates, amendment policy, per-cell scoring, and each value's source JSON: Appendix I).

\textbf{The scorecard.} 23 of 29 scored predictions confirm; the misses are the informative ones (scorecard figure and per-model tables: Appendix I). Gemma's near-misses are artifacts of the frozen scoring sites, not the effect (Appendix I.5); Qwen's one miss (P3) is a sign-convention reversal, restored one layer later at L79 (77.9\%); Phi's fully green column is preregistered failure confirmed --- P4--P6 predicted the null, and it held. Among the frozen predictions: that the best-localized model would steer least. It did.

\section{Discussion}

\textbf{What the dissociation is, mechanistically.} Four identified properties open the gap between localization and steerability. First, \emph{redundancy}: the sign is written by several layers along a shared direction, so an intervention that misses part of the set is compensated by the rest (\S{}5; Appendix~G) --- self-repair {[}10{]}. Second, \emph{opposing structure}: Qwen's minus-stamps are opposed by rescue layers that steering does not touch (Figure E5; Appendix H), consistent with its partial break rate on correct-minus answers. Third, \emph{mechanism type}: Phi's habit layers are a stamp/amplifier mixture, and the same subtraction that moves four models produces almost no purchase there. What predicts steerability is not where the sign is written but how: layers that stamp a fixed sign yield to subtraction; layers that amplify an already-chosen sign resist it. Fourth, asymmetry by answer state: the same direction moves errors more than matched correct answers --- a difference of magnitude, not presence (\S{}4). None of these properties is visible in a localization map alone; the domain matters too --- in the calculus task the minus-stamp is cleaner and flip rates higher in every model (Figure 5). The practical reading for intervention-based methods {[}11--13{]}: a sharply localized site is not yet a lever, and validating a circuit by editing at its site {[}5, 6, 9, 14{]} measures the conjunction of localization \emph{and} the network's cooperation, not localization alone.

Throughout the paper the comparisons are between plus and minus even though the object of study is the error. Errors and corrects run the same layers, in the same roles, on the same timing, differing only in the magnitude of the per-layer contributions (\S{}4), and the error pool itself is minus-skewed two to one (Table A4), so the experiments naturally work in sign coordinates: an error is the same sign-writing machinery serving the wrong sign.

\textbf{Limitations.} Steering verdicts are scored at the answer token; a free-generation probe (the generation arm of s08) agrees on small n, but full solutions run \ensuremath{\approx} 4,000 characters (\ensuremath{\approx}1--1.5k tokens) and often end without a parseable final answer at high dose, so generation could not be scored at scale (Appendix H). Flips are not always repairs: at \ensuremath{\alpha}~=~3 some flipped cases invert an already-correct sign (Figure 3 caption). A residual 26 of 703 confirmed errors (3.7\%) genuinely peak early (Appendix C.4). The origin probe of Appendix F is one model and one small completion set. And all results concern one failure family in one task-domain pair: the dissociation is demonstrated, not exhaustively mapped.

\textbf{Future work.} Two follow-ups are specified by the paper's own loose ends. The corpus probe behind the origin hypothesis: measure explicit minus versus plus tokens at answer positions in pretraining-scale mathematical text and check whether that asymmetry predicts stamp strength across models --- Appendix F's base-versus-instruct result (the stamp predates chat-training) is the first data point such a probe must fit. The repair path: a small parameter-efficient update (LoRA- or ReFT-style; cf. AxBench {[}17{]}) restricted to the identified habit layers, with the frozen battery as the pass/fail screen --- steer-aware localization: certifying a site as a lever, not merely a location, before intervening --- with stamp versus amplifier as a testable steerability predictor for new models. And one open direction: whether any direction in the discovered sign space separates correct-minus from wrong-minus --- the direction our steering shows the current space does not contain.
\section{Conclusion}

Five models, one pipeline, one intervention. In every model the written sign is decided late, at nameable layers, along a single direction --- and yet the same intervention at the same layers spans 4\% to 93\% steerability, an outcome we predicted before measuring it. The gap has mechanistic content --- redundant writers, rescue layers, and stamps that carry minus-ness rather than wrongness --- and it is not domain-bound: the same late decision, the same direction, and the same dissociation appear in both linear algebra and calculus. This paper shows what the gap {[}9, 10{]} is made of. Finding a circuit is the beginning of control, not the end of it: localization does not imply steerability. All code, per-case results, frozen predictions, and preregistration documents are in a public repository: \url{https://github.com/signcircuits-experiments/sign-circuit-scaling}.

\section*{References}
{\small

\begin{enumerate}
\def\labelenumi{[\arabic{enumi}]}
\tightlist
\item
  S. Agarwal et~al. LinAlg-Bench: A Forensic Benchmark Revealing Structural Failure Modes in LLM Mathematical Reasoning. arXiv:2605.16675, 2026. \url{https://arxiv.org/abs/2605.16675}
\item
  nostalgebraist. Interpreting GPT: the logit lens. LessWrong, 2020.
\item
  N. Belrose, Z. Furman, L. Smith, D. Halawi, I. Ostrovsky, L. McKinney, S. Biderman, and J. Steinhardt. Eliciting latent predictions from transformers with the tuned lens. arXiv:2303.08112, 2023.
\item
  N. Elhage, N. Nanda, C. Olsson, T. Henighan, N. Joseph, B. Mann, A. Askell, et~al. A mathematical framework for transformer circuits. Transformer Circuits Thread, Anthropic, 2021.
\item
  J. Vig, S. Gehrmann, Y. Belinkov, S. Qian, D. Nevo, Y. Singer, and S. Shieber. Investigating gender bias in language models using causal mediation analysis. NeurIPS, 2020.
\item
  K. Wang, A. Variengien, A. Conmy, B. Shlegeris, and J. Steinhardt. Interpretability in the wild: a circuit for indirect object identification in GPT-2 small. ICLR, 2023.
\item
  N. Nanda, L. Chan, T. Lieberum, J. Smith, and J. Steinhardt. Progress measures for grokking via mechanistic interpretability. ICLR, 2023.
\item
  A. Stolfo, Y. Belinkov, and M. Sachan. A mechanistic interpretation of arithmetic reasoning in language models using causal mediation analysis. EMNLP, 2023.
\item
  P. Hase, M. Bansal, B. Kim, and A. Ghandeharioun. Does localization inform editing? Surprising differences in causality-based localization vs. knowledge editing in language models. NeurIPS, 2023.
\item
  T. McGrath, M. Rahtz, J. Kramar, V. Mikulik, and S. Legg. The Hydra effect: emergent self-repair in language model computations. arXiv:2307.15771, 2023.
\item
  A. Turner, L. Thiergart, D. Udell, G. Leech, U. Mini, and M. MacDiarmid. Activation addition: steering language models without optimization. arXiv:2308.10248, 2023.
\item
  K. Li, O. Patel, F. Vi\'egas, H. Pfister, and M. Wattenberg. Inference-time intervention: eliciting truthful answers from a language model. NeurIPS, 2023.
\item
  A. Zou, L. Phan, S. Chen, J. Campbell, P. Guo, R. Ren, A. Pan, et~al. Representation engineering: a top-down approach to AI transparency. arXiv:2310.01405, 2023.
\item
  K. Meng, D. Bau, A. Andonian, and Y. Belinkov. Locating and editing factual associations in GPT. NeurIPS, 2022.
\item
  D. Tan, D. Chanin, A. Lynch, D. Kanoulas, B. Paige, A. Garriga-Alonso, and R. Kirk. Analysing the generalisation and reliability of steering vectors. NeurIPS, 2024.
\item
  C. Rushing and N. Nanda. Explorations of self-repair in language models. arXiv:2402.15390, 2024.
\item
  Z. Wu, A. Arora, A. Geiger, Z. Wang, J. Huang, D. Jurafsky, C. D. Manning, and C. Potts. AxBench: steering LLMs? Even simple baselines outperform sparse autoencoders. ICML, 2025.
\end{enumerate}
}

\clearpage
\appendix
\raggedbottom
\section*{Appendix A --- Dataset construction and provenance}
\addcontentsline{toc}{section}{Appendix A --- Dataset construction and provenance}

\subsection*{Appendix A.1 Background --- sign errors as a structural
failure
mode}

\textbf{The LinAlg-Bench benchmark.} Agarwal et al. {[}1{]} introduced
LinAlg-Bench, a diagnostic benchmark of 660 linear-algebra problems
(nine task types; 220 problems each at 3\ensuremath{\times}3, 4\ensuremath{\times}4, and 5\ensuremath{\times}5 matrix sizes),
every one certified by SymPy, a symbolic math library that computes
exact ground truth. Ten frontier models produced 6,600 outputs, all
graded; a three-stage judge pipeline then traced each of the 1,156
failures back to its first wrong step and gave it one of ten primary
error tags, validated against 593 human labels (92.6\% agreement). The
headline finding: model failure on this kind of math is not random noise
--- it is structured, and it changes shape predictably with matrix size.

\textbf{Where the error lives.} A sign error, in this taxonomy, is one
wrongly written sign token in the middle of the computation. The wrong
sign appears at many different sites, not one: the alternating operator
in a cofactor expansion (+\ensuremath{-}+\ensuremath{-} lost midway); the parity factor
(\ensuremath{-}1)\^{}(i+j) dropped from a cofactor term; a negative matrix entry
mishandled, so \ensuremath{-}(\ensuremath{-}4) is written as \ensuremath{-}4; a product of two negatives
written negative; and in integration by parts, the minus written when
the formula is applied recursively. The tag is assigned at the first
wrong value, and only when the magnitude at that step is correct and the
sign alone is wrong --- this rule is what separates a pure sign error
from an arithmetic error. Everything downstream then runs correctly on
the corrupted expression.

\textbf{How large is the class, and how does it grow?} The taxonomy has
ten primary tags --- four execution errors (sign, arithmetic,
carry-down, memory loss), two structural (hallucination, method
failure), and four artifacts --- and sign error is a major class. At 3\ensuremath{\times}3
it leads every tag: 33.0\% of all failures. As difficulty rises, its
count nearly quadruples, 33 \ensuremath{\rightarrow} 123 \ensuremath{\rightarrow} 114 sign errors at 3\ensuremath{\times}3 / 4\ensuremath{\times}4 / 5\ensuremath{\times}5
(270 total), and the recursion-driven parity subtypes surge from 3.0\%
to 61.0\% of sign errors at 4\ensuremath{\times}4. Its share of failures falls to 17.2\%
at 5\ensuremath{\times}5 only because models stop computing at all there --- hallucination
(fabricated or abandoned computation) rises from 17.0\% to 47.1\% and
takes over. So sign errors dominate exactly where models still execute
the algorithm.

\textbf{Which models make them.} The ten models evaluated span three
accuracy tiers on the determinant task: Tier 1 --- OpenAI-o1,
Gemini-3.1-Pro, DeepSeek-V3, and Qwen3-235B (\ensuremath{\ge}74\% accuracy); Tier 2 ---
GPT-5.2 and Mistral-Large (36--54\%); Tier 3 --- Claude-4.5-Sonnet,
Qwen2.5-72B, Llama-3.3-70B, and GPT-4o (0--12\%). Sign errors are not
evenly spread across them: top-tier Gemini-3.1-Pro commits zero at any
size, while the four Tier-3 models account for 221 of the 270 (81.9\%).
Yet the class persists across tiers, consistent with reports that
transformers systematically struggle with signed arithmetic. The two
models this paper studies mechanistically --- Qwen2.5-72B and
Llama-3.3-70B --- sit in Tier 3. They are also the only models on that
list a mechanistic study can realistically open: most of the ten are
closed-weight (OpenAI-o1, Gemini-3.1-Pro, GPT-5.2, Claude-4.5-Sonnet,
GPT-4o --- an API exposes no internal activations to probe), and the
remaining open ones are too large to probe layer-by-layer (DeepSeek-V3,
Qwen3-235B). Mistral-Large-675B was also attempted; it enters this paper
only as an exploratory behavioral run, since the causal battery
(ablation, patching, dose-response steering) is computationally
prohibitive at that scale (Appendix A.5, C.5). To test that the findings
are not particular to the two 70B models, we therefore added three
further open models at smaller sizes --- Phi-4 (14B), Mistral-Small
(24B), and Gemma-3 (27B).

\textbf{Where we probe --- the token picture.} Every measurement in this
paper sits at one position inside the model's own written solution ---
precisely, at the token immediately before the sign token (position \ensuremath{-}1):
the hidden state there is what the model uses to write the sign, so the
layer-by-layer readout at that position shows the sign being decided.
The error site differs by subtype:

\begin{center}
\fbox{\begin{minipage}{0.94\linewidth}\vspace{5pt}
\textbf{det --- alternating drift} \emph{(real traced case; probed sign in brackets)}\\[2pt]
\ldots{} det(A) = \ensuremath{-}3\ensuremath{\cdot}det(A11) \ensuremath{-} (\ensuremath{-}4)\ensuremath{\cdot}det(A12) \textbf{{[}\ensuremath{-}{]}} (\ensuremath{-}4)\ensuremath{\cdot}det(A13) \ensuremath{-} \ldots\\
\ensuremath{\uparrow} should be ``+'' --- alternation broken\\[7pt]
\textbf{det --- double negative} \emph{(illustrative form)}\\[2pt]
\ldots{} term = \ensuremath{-}(\ensuremath{-}4)\ensuremath{\cdot}M13 = \textbf{{[}\ensuremath{-}{]}}4\ensuremath{\cdot}M13 \ldots\\
\ensuremath{\uparrow} should be ``+4'' --- the two minuses not resolved\\[7pt]
\textbf{IBP --- recursive step} \emph{(illustrative form)}\\[2pt]
\ldots{} = x\ensuremath{^2}e\ensuremath{^x} \ensuremath{-} 2(xe\ensuremath{^x} \ensuremath{-} e\ensuremath{^x}) = x\ensuremath{^2}e\ensuremath{^x} \ensuremath{-} 2xe\ensuremath{^x} \textbf{{[}\ensuremath{-}{]}} 2e\ensuremath{^x} \ldots\\
\ensuremath{\uparrow} should be ``+'' --- the recursive minus misdistributed\\[7pt]
\textbf{Downstream:} correct arithmetic on the corrupted expression \ensuremath{\rightarrow} final answer a wrong number (traced det case: 323; truth 83).\\[3pt]
\textbf{Probe:} hidden state at the token before the bracketed sign, replayed teacher-forced; everything before it is held exactly as the model wrote it.
\vspace{5pt}\end{minipage}}
\end{center}

\textbf{What this study probes.} The taxonomy splits sign errors into
three structural families {[}1, App. F{]}. Across every subtype the
surface signature is identical --- a single wrongly written sign token
--- and that shared signature is the object this paper studies
mechanistically. Every subtype has a concrete written form:

Parity family --- the (\ensuremath{-}1)\^{}(i+j) bookkeeping of cofactor
expansion; 61.0\% of 4\ensuremath{\times}4 sign errors:

\begin{itemize}
\item
  \textbf{Alternating drift:} the +\ensuremath{-}+\ensuremath{-} operator pattern is tracked
  correctly for the first terms, then lost midway --- det = a11\ensuremath{\cdot}M11 \ensuremath{-}
  a12\ensuremath{\cdot}M12 \ensuremath{-} a13\ensuremath{\cdot}M13 ... (third operator should be +). This is the traced
  case in the diagram above.
\item
  \textbf{Cofactor neglect:} a minor is computed correctly but used raw,
  with the parity factor never applied --- +a32\ensuremath{\cdot}M32 written where
  position (3,2) demands \ensuremath{-}a32\ensuremath{\cdot}M32.
\item
  \textbf{Parity sign error:} the parity factor is applied, but with the
  wrong sign or applied twice --- C13 = \ensuremath{-}M13 where (\ensuremath{-}1)\^{}(1+3) = +1.
\end{itemize}

Product family --- signed multiplication anywhere in the
computation; 75.8\% of 3\ensuremath{\times}3 sign errors:

\begin{itemize}
\item
  \textbf{Product sign error:} a product has correct magnitude, wrong
  sign --- (\ensuremath{-}3)\ensuremath{\cdot}5 = 15.
\item
  \textbf{Double negative trap:} two negatives meet and are not resolved
  --- \ensuremath{-}(\ensuremath{-}4) written as \ensuremath{-}4 instead of +4.
\item
  \textbf{Rule interference:} a negative entry is treated as positive
  throughout --- entry \ensuremath{-}2 enters the expansion as 2.
\end{itemize}

\textbf{Standalone ---} an isolated flip orthogonal to both: operation
direction (adding where the algorithm subtracts, or vice versa) and
silent sign flip (a signed value appears with no working shown for that
one step, while surrounding steps show their work).

The same taxonomy carries over to IBP. The clean recursive case is a
double negative trap on the by-parts minus: distributing it, x\ensuremath{^2}e\ensuremath{^x} \ensuremath{-}
2(xe\ensuremath{^x} \ensuremath{-} e\ensuremath{^x}) is written as x\ensuremath{^2}e\ensuremath{^x} \ensuremath{-} 2xe\ensuremath{^x} \ensuremath{-} 2e\ensuremath{^x} --- the minus-times-minus
never resolved to +. In the IBP sign errors this study uses, silent sign
flip is the most common subtype, followed by the double negative trap
and alternating drift.

\textbf{Why this matters.} A sign error is the smallest possible failure
--- one token --- inside an otherwise correct computation, produced even
by models that execute everything else right. It is structured
(predictable subtypes, predictable growth with recursion depth), it
dominates failures where models still compute, and it survives in
current frontier systems. A one-token, structurally regular failure is
exactly the kind interpretability can hope to localize and remove ---
which is what this paper attempts. Appendix C.2 walks through one
concrete case end to end.

\subsubsection*{\textbf{A.2 Problem
generation}}

Determinant problems. The determinant data began with the 4\ensuremath{\times}4
problems of LinAlg-Bench {[}1{]}, but the benchmark bank alone could not
supply enough sign errors, so each of the five core models ---
Gemma-3-27B, Llama-3.3-70B, Mistral-Small-24B, Phi-4-14B and
Qwen-2.5-72B --- was given its own larger bank, generated with the
benchmark's released generator to the same specification: integer
matrices with entries drawn uniformly from {[}\ensuremath{-}5, 5{]}, ground truth
computed symbolically so every answer is an exact integer, and every new
matrix deduplicated against all prior determinant corpora by exact
content match. Bank sizes per model are the first row of Table A1 ---
Gemma 4050, Llama 2850, Mistral 2050, Phi 2050, Qwen 675; the more
accurately a model computes, the more problems it takes to harvest
enough errors from it.

Integration-by-parts problems. The IBP problems are not part of
LinAlg-Bench. We built this bank ourselves: a custom poly-trig
integration-by-parts generator, created specifically to test whether the
sign circuit extends beyond linear algebra to a second domain. Each of
the same five models received 1,294 problems (train + validation
combined; Table A3), and every problem forces at least one application
of the parts formula, so the minus of \ensuremath{-}\ensuremath{\int}v du must be written
explicitly.

\subsubsection*{\textbf{A.3 Harvesting and
grading}}

Every model answered every problem in its bank at temperature 0,
under a fixed instruction to show all computation steps and box the
final answer. A response is graded correct if and only if the boxed
answer equals the symbolic ground truth exactly --- no tolerance. This
single check sorts every response into the two pools the rest of the
paper uses: wrong answers, which may contain a sign error (A.3.1), and
correct answers, which supply the matched controls (A.3.2).

\subsubsection*{\textbf{A.3.1 Error
cases}}

Wrong answers pass through the LinAlg-Bench judge pipeline
{[}1{]}, used exactly as released; its agreement with expert human
labels is reported in {[}1{]} and quoted in A.1. A truncation pre-check
first sets aside any response without a complete boxed answer --- an
incomplete response is never classified. A classification judge (an LLM
judge at temperature 0) then reads the full response against the ground
truth, locates the first wrong value, and applies the magnitude rule:
the case is a sign error only if that value has the correct absolute
magnitude and the wrong sign; if the magnitude is also wrong, the case
is tagged arithmetic instead. A second, independent validation judge
re-examines every proposed sign error and can reject it --- no case is
kept on the classification judge's word alone. A sign error runs in
either direction: the model may write ``\ensuremath{-}'' where the truth is ``+'', or
``+'' where the truth is ``\ensuremath{-}'' --- the wrote-+ / wrote-\ensuremath{-} columns of
Table 1 count both. The pipeline's third stage is not fully automatic:
any case the validation judge flags as ambiguous goes to manual review
--- a human, not a model, is the final judge --- as described in
{[}1{]}.

Every experiment needs the character-level position of the
wrongly written sign in the model's own response. An LLM annotator marks
the position and copies ten characters of verbatim context on each side;
a deterministic check then verifies that the character at the recorded
offset is exactly the annotated sign and that it lies inside the model's
answer rather than the prompt. A manual check is involved at each stage
as well: we verified samples by hand at every stage, the manual and
automated results agreed statistically, and only then did cases go into
the final experiment. A case that survives every stage --- graded wrong,
confirmed a sign error by the judges, its sign located and
offset-verified --- is experiment-ready: a case the paper's experiments
can actually run on. Per-model counts at every stage are in Table A1.
All probing then happens at the token immediately before the sign,
teacher-forced, with everything before it held exactly as the model
wrote it.

A concrete case: NC\_4x4\_det\_201 (Qwen-2.5-72B, held-out;
ground truth det(A) = 83, model answer 323, family Alternating\_Drift).
In the cofactor expansion the model wrote det(A) = \ensuremath{-}3\ensuremath{\cdot}det(A11) \ensuremath{-}
(\ensuremath{-}4)\ensuremath{\cdot}det(A12) \ensuremath{-} (\ensuremath{-}4)\ensuremath{\cdot}det(A13) \ensuremath{-} (\ensuremath{-}3)\ensuremath{\cdot}det(A14), then summed 345 \ensuremath{-} 160 +
120 + 18 = 323. The A13 term is the slip: the cofactor sign is
(\ensuremath{-}1)\^{}(1+3) = +1, so the correct contribution is \ensuremath{-}4 \ensuremath{\times} 30 = \ensuremath{-}120; the
model's extra minus made it +120, adding 240 and turning 83 into 323.
The first wrong value is that third cofactor term --- magnitude right,
sign flipped --- so the pipeline tags it a sign error. Appendix C.2
traces this same case through the logit lens.

\subsubsection*{\textbf{A.3.2 Correct
cases}}

A correct case enters the pool by the exact-match check alone;
no judge is involved. We do not control which problems a model gets
right, so the wrote-+ / wrote-\ensuremath{-} split among corrects is simply the
models' own behavior --- that is why the positive and negative counts
differ so much between models. To compare a correct case against an
error case we must probe the same kind of sign. An error case is probed
at its wrongly written sign; a correct solution has no wrong sign, so we
probe the sign at the same kind of step, written correctly --- in det(A)
= 0\ensuremath{\cdot}C11 \ensuremath{-} 5\ensuremath{\cdot}C12, the minus before 5\ensuremath{\cdot}C12 is a cofactor-expansion operator
sign, exactly the type that goes wrong in error cases, here written
right. The annotator-plus-deterministic-check procedure of A.3.1 locates
that matched sign and verifies its offset the same way. Only verified
rows enter any experiment; per-model counts are in Table A2.

\subsubsection*{\textbf{A.4 The data funnel and the
discovery / held-out
split}}

Tables A1---A3 give the full funnel --- bank \ensuremath{\rightarrow} graded \ensuremath{\rightarrow}
validated sign errors \ensuremath{\rightarrow} experiment-ready \ensuremath{\rightarrow} discovery / held-out --- for
the five core models. Rows not marked file-verified are counts noted at
harvest time; the file-verified rows are recomputed from the released
per-case files and are the counts every experiment in this paper uses
(A.6).

The split rule. All the questions producing errors and their
correct counterparts are split into discovery and held-out sets ---
separate batches, not a random split of one pool. The discovery set was
used to identify the layers and the sign direction; the held-out set is
a distinct batch of new problems the experiments never saw. Before any
held-out measurement, overlap with the discovery set is checked at both
the ID level and the content level (exact matrix-entry comparison) and
overlapping rows are dropped --- for Qwen, 206 received rows lost 59 to
content overlap, leaving 147. Held-out data is touched exactly once: a
technical crash may be restarted, a disliked result may not.

\textbf{Table A1. Determinant 4\ensuremath{\times}4 error funnel, five core models.}

\begin{longtable}[]{@{}>{\raggedright\arraybackslash}p{0.351\textwidth}>{\raggedright\arraybackslash}p{0.099\textwidth}>{\raggedright\arraybackslash}p{0.117\textwidth}>{\raggedright\arraybackslash}p{0.153\textwidth}>{\raggedright\arraybackslash}p{0.081\textwidth}>{\raggedright\arraybackslash}p{0.108\textwidth}@{}}
\toprule
\textbf{Stage} & \textbf{Gemma-3-27B} & \textbf{Llama-3.3-70B} &
\textbf{Mistral-Small-24B} & \textbf{Phi-4-14B} &
\textbf{Qwen-2.5-72B}\tabularnewline
\midrule
\endhead
Bank size & 4050 & 2850 & 2050 & 2050 & 675\tabularnewline
PASS (correct) & 3555 & 2306 & 438 & 223 & 133\tabularnewline
FAIL (wrong / no answer) & 495 & 544 & 1612 & 1827 & 542\tabularnewline
Accuracy & 87.8\% & 80.9\% & 21.4\% & 10.9\% & 19.7\%\tabularnewline
Sign errors as \% of FAILs & 54\% & 43\% & 16\% & 20\% &
43\%\tabularnewline
SIGN\_ERROR working set & 266 & 234 & 263 & 361 & 233\tabularnewline
Experiment ready & 219 & 208 & 200 & 200 & 206\tabularnewline
Experiment-ready errors (file-verified) & 218 & 208 & 237 & 211 &
206\tabularnewline
Discovery errors (file-verified) & 92 & 16 & 95 & 61 & 59\tabularnewline
Held-out errors (file-verified) & 126 & 192 & 142 & 150 &
147\tabularnewline
\bottomrule
\end{longtable}

Table A2. Determinant 4\ensuremath{\times}4 correct sets (matched controls),
discovery / held-out, file-verified.

\begin{longtable}[]{@{}llll@{}}
\toprule
\textbf{Model} & \textbf{Discovery (file-verified)} & \textbf{Held-out
(file-verified)} & \textbf{Total}\tabularnewline
\midrule
\endhead
Gemma-3-27B & 50 & 138 & 188\tabularnewline
Llama-3.3-70B & 80 & 130 & 210\tabularnewline
Mistral-Small-24B & 78 & 122 & 200\tabularnewline
Phi-4-14B & 50 & 149 & 199\tabularnewline
Qwen-2.5-72B & 47 & 141 & 188\tabularnewline
\bottomrule
\end{longtable}

\textbf{Table A3. IBP funnel: custom poly-trig bank, 1,294 problems per
model, train + validation combined.}

\begin{longtable}[]{@{}>{\raggedright\arraybackslash}p{0.328\textwidth}>{\raggedright\arraybackslash}p{0.103\textwidth}>{\raggedright\arraybackslash}p{0.122\textwidth}>{\raggedright\arraybackslash}p{0.159\textwidth}>{\raggedright\arraybackslash}p{0.084\textwidth}>{\raggedright\arraybackslash}p{0.112\textwidth}@{}}
\toprule
\textbf{Stage} & \textbf{Gemma-3-27B} & \textbf{Llama-3.3-70B} &
\textbf{Mistral-Small-24B} & \textbf{Phi-4-14B} &
\textbf{Qwen-2.5-72B}\tabularnewline
\midrule
\endhead
Problem bank (train + validation) & 1,294 & 1,294 & 1,294 & 1,294 &
1,294\tabularnewline
PASS (correct inference) & 396 & 395 & 215 & 284 & 435\tabularnewline
FAIL (wrong / no answer / N/A) & 898 & 899 & 1,079 & 1,010 &
859\tabularnewline
Working set (sign-error candidates) & 133 & 43 & 116 & 120 &
155\tabularnewline
Experiment ready & 128 & 43 & 109 & 117 & 132\tabularnewline
\bottomrule
\end{longtable}

Notes. The FAIL row of Table A3 includes truncated or nil
responses (Gemma 2, Llama 7, Mistral 40, Phi 0, Qwen 23). Working set
567 \ensuremath{\rightarrow} experiment-ready 529: 12 rows were withdrawn for a corrupt byte in
the prompt prefix (Qwen 7, Mistral 5); the remainder dropped in later
filtering or manual review. Per-model IBP accuracy at harvest: Qwen
33.6\%, Gemma 30.6\%, Llama 30.5\%, Phi 21.9\%, Mistral 16.6\%.

\textbf{The sign the model wrote --- an uncontrolled outcome.} Nothing
in this funnel selects on the sign of the error. The pools were
harvested, not designed: 11,675 determinant problems and 6,470 IBP
problems (18,145 in all) were graded at temperature 0, producing 5,020
determinant and 4,745 IBP failures; the judge pipeline kept a case only
for where the error is (magnitude right, sign alone wrong, offset
verified) --- never for which sign was written. Yet the surviving pool
is skewed: minus errors outnumber plus errors about two to one. Table A4
gives the breakdown. The skew is stronger in IBP (75.2\% minus) than in
determinants (64.8\%), reaches roughly nine in ten cases for Phi and
Mistral-Small, and reverses only in Gemma. The main text returns to this
asymmetry in \S{}4 (``Only minus gets stamped'') and \S{}7.

\textbf{Table A4.} Written sign of the error --- every file-verified
error case in the logit-lens pool (discovery + held-out).

\begin{longtable}[]{@{}lllll@{}}
\toprule
\textbf{Model} & \textbf{det \ensuremath{-} / +} & \textbf{IBP \ensuremath{-} / +} & \textbf{Total
\ensuremath{-} / +} & \textbf{\% minus}\tabularnewline
\midrule
\endhead
Gemma-3-27B & 87 / 131 & 64 / 79 & 151 / 210 & 41.8\%\tabularnewline
Llama-3.3-70B & 131 / 77 & 30 / 13 & 161 / 90 & 64.1\%\tabularnewline
Mistral-Small-24B & 199 / 38 & 98 / 11 & 297 / 49 &
85.8\%\tabularnewline
Phi-4-14B & 181 / 30 & 109 / 8 & 290 / 38 & 88.4\%\tabularnewline
Qwen-2.5-72B & 102 / 104 & 36 / 0 & 138 / 104 & 57.0\%\tabularnewline
All five & 700 / 380 & 337 / 111 & 1,037 / 491 & 67.9\%\tabularnewline
\bottomrule
\end{longtable}

\textbf{Note.} Counts are the lens pool of Tables A1/A3 (for Gemma and
Qwen IBP, the subset of the experiment-ready set with logit-lens runs).
The exploratory Mistral-Large-675B 5\ensuremath{\times}5 run (A.5) adds 21 minus / 30
plus.

\subsubsection*{\textbf{A.5 The Mistral-Large-675B
run}}

Most of the data for the exploratory Mistral-Large-675B run came
directly from LinAlg-Bench {[}1{]}; a small number of additional samples
were generated to reach the exploratory discovery count. The model
enters this paper only as a behavioral, unfiltered run on the 5\ensuremath{\times}5
determinant task (Figure 1); the causal battery used on the five core
models --- ablation, patching, dose-response steering --- is not applied
to it, since those operations are computationally prohibitive at 675B
scale. The discovery count is 51 sign-error cases --- the same 51
plotted in Figure 1 --- verified against the released per-case files.

\subsubsection*{\textbf{A.6
Provenance}}

Every count marked file-verified in this appendix can be recomputed from
the released repository
(\url{https://github.com/signcircuits-experiments/sign-circuit-scaling}):
it contains the experiment-ready case files, all judge outputs, the
sign-offset metadata, the generation code, and every per-case experiment
output (s01--s09 JSONs), under an md5 checksum manifest. A verification
script (paper/appendix/verify\_all.py) re-derives the seventeen headline
numbers of the paper directly from those JSONs, printing pass or fail
for each against the entry-by-entry ledger in verification\_ledger.md.
The harvest-time counts were recorded as the data was collected; the raw
harvest files behind them are not part of the release and are available
on request.

\clearpage
\section*{Appendix B --- The s00--s09 pipeline}
\addcontentsline{toc}{section}{Appendix B --- The s00--s09 pipeline}

\subsection*{B.1 What the pipeline is}

Every mechanistic number in this paper comes out of one fixed, numbered
pipeline of stages s00--s09, run with the same code on all five models
(pipeline\_release/stages/s00\_arith\_control.py \ldots{} s09\_judge.py
--- the code is the methodology described below, one script per stage).
Two descriptive stages (s01, s02) read the model without touching it and
locate where the sign is decided --- what the main text calls the
written-sign lead (\S{}3), which the appendices also call the logit-diff:
the gap between the model's logit for the correct sign and its logit for
the sign it actually wrote. A derivation stage (s03) turns those
readouts into a frozen list of target layers by a fixed rule; the causal
stages (s04--s09) may intervene only at layers named in that list. The
list is written with a review flag set to false, and the PI checks the
automatic picks against the per-domain evidence before setting the flag
to true --- this is what keeps the causal tests honest.

One case, end to end. To show how the stages hand off to one another,
follow a single sign-error case through the whole pipeline (the specific
case, its layers and its counts are reported where each stage's own
results live --- Appendix A and the per-stage appendices --- not here).
Stage s01 locates the layer where that case's logit-diff peaks. Stage
s02 reads the same case and attributes that peak to specific MLP and
attention components. Stage s03 takes both readouts, together with every
other case in the same model/domain cell, and derives one frozen set of
target layers for the cell. From there the case enters the causal
battery in a fixed order: s04 characterizes its geometry against the
minus direction, s05/s05b/s05c test whether the target layers are
necessary, s06 tests the minus direction itself by subtraction, s07
validates that result by patching, s08 steers with it, and s09 asks
whether the model's own check layer can be made to override it. Each
stage writes one verdict per case to its own results file for that
model, domain and split.

\subsection*{B.2 What each stage measures}

\textbf{s00 --- arithmetic control. Goal:} build a cross-model baseline
uninfluenced by the sign-error domains. \textbf{Method:} every model
completes the sign on the same 48 shared arithmetic problems; the
layer-credit profile on this set is the baseline referenced in Appendix
D.

\textbf{s01 --- logit lens. Goal:} find where the written-sign lead
(logit-diff) is decided. \textbf{Method:} at every layer, compute the
logit-diff between the correct and written sign token at the sign
position.

\emph{peak\_layer = argmax\textsubscript{L} logit-diff(L) flip\_layer =
first L where logit-diff(L) \textgreater{} noise floor}

The split into embedding-peak (peak\_layer \ensuremath{\le} 5) and deeper-peak is
exhaustive by construction: a case's peak layer is a single number, and
that number is either \ensuremath{\le}5 or it isn't --- there is no third class. In our
data every case lands in exactly one of these two buckets, with nothing
in between. This is the recap of the procedure behind Figures 1--2 and
Appendix C's walked case. Feeds s03.

\textbf{s02 --- direct logit attribution. Goal:} assign layer-wise
credit for the logit-diff. \textbf{Method:} decompose the logit-diff at
the probe token into the embedding term plus each layer's MLP and
attention contribution.

\emph{DLA(component) = component\_output \ensuremath{\cdot} \ensuremath{\hat{d}} (embedding, per-layer MLP,
per-layer attention, and per-head over the last 13 layers)}

A layer whose mean MLP DLA pushes toward the written sign is a habit
pusher, one pushing against it a corrector. Feeds s03; stands behind
Figure 2 and Appendix E.6.

\textbf{s03 --- target derivation. Goal:} turn s01/s02 into a frozen,
reviewable target list. \textbf{Method:} restrict to the late window
(last 25\% of layers).

\emph{habit\_layers = top 2 late-window layers by mean MLP DLA toward
the written sign}

\emph{check\_layer = argmin\textsubscript{L} (mean MLP DLA toward the
written sign)}

\emph{control\_layers = round(0.375 \ensuremath{\cdot} N), round(0.625 \ensuremath{\cdot} N)}

Human-reviewed before any causal stage runs.

\textbf{s04 --- projections. Goal:} test whether one direction separates
minus- from plus-writers (prediction P3). \textbf{Method:} at each
capture layer, project the hidden state onto the minus direction and
record its norm.

\emph{proj\_d = h\textsubscript{L} \ensuremath{\cdot} \ensuremath{\hat{d}} norm\_h = \ensuremath{\|}h\textsubscript{L}\ensuremath{\|}}

\textbf{s05 --- mean ablations. Goal:} the necessity test (Appendix G).
\textbf{Method:} replace a target MLP's output with its mean, one site
at a time (habit, check, control, top-attribution), then jointly for up
to 5 sites.

\emph{\ensuremath{\Delta}logit-diff = logit-diff\textsubscript{post} \ensuremath{-}
logit-diff\textsubscript{baseline}}

\textbf{s05b / s05c --- per-head and joint-head ablations. Goal:} rule
out a single-head account of the sign decision. \textbf{Method:} s05b
mean-ablates individual attention heads; s05c ablates small groups
(pusher-pair, pushers+corrector, correctors, control group), using the
same \ensuremath{\Delta}logit-diff measure as s05. Run as an exploratory add-on on Gemma,
Mistral-Small-24B and Phi (discovery error domains). \emph{\textbf{Note
the Llama exception:}} Llama's per-head check covers a single cell only
--- held-out determinant errors --- not IBP or discovery; no per-head
ablation file exists for Qwen, consistent with Qwen's own
pre-registration making s05b conditional on an s02 head-level screen
that the record does not show firing.

\textbf{s06 --- direction subtraction. Goal:} a donor-free causal test
of the habit direction (Appendix H; prediction P4). \textbf{Method:}
subtract, in place, the projection of the habit-layer MLP output onto
the minus direction, then remeasure the logit-diff. Control layers
receive the same subtraction as a null.

\emph{MLP\textsubscript{out}\ensuremath{'} = MLP\textsubscript{out} \ensuremath{-} \ensuremath{\alpha}
(MLP\textsubscript{out} \ensuremath{\cdot} \ensuremath{\hat{d}}) \ensuremath{\hat{d}}, \ensuremath{\alpha} \ensuremath{\in} \{1, 2, 3\}}

(\ensuremath{\alpha} matches Figure 3's dose grid; \ensuremath{\hat{d}} is the sign/minus direction of \S{}5;
\ensuremath{\Delta}logit-diff as defined under s05.)

\textbf{s07 --- activation patching. Goal:} methodological validation of
s06, independent of the direction-subtraction machinery.
\textbf{Method:} replace a target layer's activation at the probe token
with a donor-set mean, in three conditions.

\emph{h\textsubscript{L} \ensuremath{\leftarrow} mean over donor set (h\textsubscript{L})}

Donor sets: the correct-set mean patched into error runs; the error-set
mean patched into correct runs (the reverse); and a sign-matched
control. Outcome measured as \ensuremath{\Delta}logit-diff and as a flip count (did the
sign reverse). Results: Appendix J.K.

\textbf{s08 --- steering. Goal:} the C2 intervention of \S{}5, Figure 3.
\textbf{Method:} apply the s06 subtraction simultaneously at both habit
layers, rather than one at a time.

\emph{MLP\textsubscript{out,h1}\ensuremath{'}, MLP\textsubscript{out,h2}\ensuremath{'} (both habit
layers, same subtraction as s06)}

Compared against a magnitude-matched random-direction control
(prediction P6), a C2-on-correct arm (Appendix H), and optional
full-answer generation after steering. Results: Appendix I.

\textbf{s09 --- judge tests. Goal:} ask whether the model's own
check-layer ``judge'' can be forced to override the habit.
\textbf{Method:} two arms, opposite directions:

\textbf{ArmB (on error cases):} \emph{MLP\textsubscript{check}\ensuremath{'} = \ensuremath{\gamma} \ensuremath{\cdot}
MLP\textsubscript{check}, \ensuremath{\gamma} \ensuremath{\in} \{1.5, 2.0, 3.0\} --- does boosting the
judge flip the written sign?}

\textbf{ArmA (on correct cases):} \emph{MLP\textsubscript{check} \ensuremath{\leftarrow} mean
source (A1: sign-matched error-set mean; A2: pooled correct-set mean)
--- does weakening the judge break a correct case?}

A cross-domain variant runs the ArmB boost on domains the judge
candidate wasn't paired with. A case counts as flipped (ArmB) or broken
(ArmA) when logit-diff changes sign.

\subsection*{B.3 Data note}

All stage outputs --- target lists, per-case results, checksums --- are
released at github.com/signcircuits-experiments/sign-circuit-scaling.
Layout: one folder per model (Gemma/, Llama/, Mistral-Small-24B/, Phi/,
Qwen/), each split into
\textless phase\textgreater\_\textless domain\textgreater\_\textless split\textgreater/
(e.g. discovery\_DET\_error/), each holding one subfolder per stage
(s01\_logit\_lens/ \ldots{} s09\_judge/). Code is in pipeline\_release/.

\clearpage
\section*{Appendix C --- The late decision: the complete logit-lens
experiment behind Figure 1 and Figures E1--E2}
\addcontentsline{toc}{section}{Appendix C --- The late decision: the complete logit-lens
experiment behind Figure 1 and Figures E1--E2}

The paper claims the sign decision happens \emph{late} --- in the last
quarter of the network. Figure 1 and the trajectory figures of Appendix
E (Figures E1--E2) carry that claim. Figure 1 plots every confirmed case
as one point at the depth where its written sign peaks: the great
majority sit past depth 0.75. Figures E1--E2 show the same thing as
group curves: flat or negative through the middle of the network, then a
steep late climb.

Figure 1 also shows a small number of points near depth 0 --- mostly
from Qwen-2.5-72B and Gemma-3-27B. If those are real early decisions,
the claim has exceptions. This appendix exists for both halves of that
picture: it gives the full recipe behind Figure 1 and Figures E1--E2,
and it works through every early point to determine which are genuine
and which are mirages of the metric.

C.1 gives the recipe and the definitions. C.2 traces one case end to
end. C.3 gives the complete case accounting (Table C1). C.4 examines
where the peaks fall --- the late majority, then each early phenomenon
by name, model by model, with its figure beside it. C.5 covers the
exploratory sixth model. Every number below is recomputable from the
released per-case files, organized per model and phase (Appendix A.6).

\textbf{C.1 The recipe}

The experiment behind both figures is one procedure, applied identically
to every case:

\begin{quote}
1. Take a confirmed case --- a problem where the model wrote a sign,
right or wrong, in its own transcript.

2. \textbf{Locate the probe token:} the last token before the sign the
model wrote (located and offset-verified per Appendix A.3.1). Everything
before it is replayed teacher-forced, exactly as the model wrote it.

3. Run one forward pass and, at every layer, decode the residual stream
at the probe position into vocabulary scores --- the \textbf{logit lens}
{[}2{]}: as if the model had to answer from that layer.

4. \textbf{Record the lead at every layer:} the written sign's logit
minus the other sign's logit. Positive means the layer state favors what
the model went on to write; negative means it favors the opposite sign.

5. \textbf{Find the peak:} the layer where the lead is largest. Relative
depth = layer / (n\_layers \ensuremath{-} 1). Depth \ensuremath{\ge} 0.75 --- the last quarter of
the network, the same boundary Figure 1 and Figures E1--E2 draw as a
dashed line --- counts as late; below 0.75, early.
\end{quote}

One forward pass per case, one peak per case, one point in Figure 1.

\textbf{The confirmation rule.} Before a case enters any statistic,
three steps decide whether its trajectory can be read at all:

\begin{quote}
1. Read the lead at the final layer.

2. If it is \ensuremath{\ge} +2 logits, the case is \textbf{confirmed}: the lens sees,
at the output, the sign the model actually wrote.

3. Otherwise the case is \textbf{excluded}: its final state disagrees
with, or barely supports, what was written, so its trajectory cannot be
interpreted. Excluded cases are not evidence against late decisions;
they carry no locatable decision. The grey band in Figure 1 marks this
\ensuremath{\pm}2 zone.
\end{quote}

\textbf{C.2 One case, end to end}

\textbf{Case NC\_4x4\_det\_201} (Qwen-2.5-72B, 4\ensuremath{\times}4 determinant,
held-out; ground truth 83, model answer 323, error family
Alternating\_Drift). The model carried out the cofactor expansion, got
every magnitude right, and wrote one wrong sign. The bracket closes on
one cofactor term; the next token is the sign of the next term. The
probe token is marked with {[} {]}:

\textbackslash det(A) = -3\ensuremath{\cdot}\textbackslash det(A\_\{11\}) -
(-4)\ensuremath{\cdot}\textbackslash det(A\_\{12\textbf{{[}\}){]}}

next token the model writes: "\ensuremath{-}" (true sign: "+")

The bold \textbf{{[}\}){]}} is the probe token: the tokenizer's final
token of the expansion line, the closing of
\textbackslash det(A\_\{12\}), and the last token before the sign the
model writes next.

Reading the trace with the definitions of C.1:

\begin{itemize}
\item
  \textbf{Probe token:} ``\})'', closing \textbackslash det(A\_\{12\}).
  The next token the model writes --- ``\ensuremath{-}'' where the true sign is ``+''
  --- is the first definitely wrong token.
\item
  \textbf{Layer 0:} the lead is already +4.82 toward ``\ensuremath{-}'' before a
  single layer has run. Layer 0 is the embedding; why it leans is the
  subject of C.4.3.
\item
  \textbf{Middle layers:} the lead goes negative --- the early lean is
  wiped out.
\item
  \textbf{Late layers:} the lead turns positive for good at 95--99\%
  depth, at the habit layers L75/L78 (\S{}4), and ends above +2 logits \ensuremath{\rightarrow}
  confirmed by the rule of C.1.
\item
  \textbf{Where it lands in Figure 1:} a solid point at depth 0 (errors
  solid, corrects open) --- one of the early points this appendix must
  explain. Its explanation, with the group statistics and the figure, is
  in C.4.3.
\end{itemize}

\textbf{C.3 The accounting}

The lens readout runs on every experiment-ready case, in both phases,
for all five models; the confirmation rule of C.1 then decides which
cases enter Figure 1 and the depth statistics. Discovery cases
identified the habit layers and the sign direction; held-out cases are a
separate batch of new problems, checked for overlap at the ID and
content level before any measurement (Appendix A.4). Discovery appears
in Table C1 for completeness; paper verdicts use held-out only.

To read one row: Gemma-3-27B held-out --- 264 cases enter, 169 are
confirmed, 95 excluded; of the 169, 49 are sign errors and 120 corrects;
the raw phase totals before the rule are 126 errors and 138 corrects.
Footnotes (a)--(f) follow the table.

\textbf{Table C1:} Complete determinant case accounting per model and
evaluation set.

\begin{longtable}[]{@{}>{\raggedright\arraybackslash}p{0.096\textwidth}>{\raggedright\arraybackslash}p{0.051\textwidth}>{\raggedright\arraybackslash}p{0.037\textwidth}>{\raggedright\arraybackslash}p{0.107\textwidth}>{\raggedright\arraybackslash}p{0.045\textwidth}>{\raggedright\arraybackslash}p{0.107\textwidth}>{\raggedright\arraybackslash}p{0.091\textwidth}>{\raggedright\arraybackslash}p{0.096\textwidth}>{\raggedright\arraybackslash}p{0.079\textwidth}>{\raggedright\arraybackslash}p{0.062\textwidth}>{\raggedright\arraybackslash}p{0.062\textwidth}@{}}
\toprule
\textbf{Model} & \textbf{Phase} & \textbf{Total} & \textbf{Confirmed
(lead \ensuremath{\ge} 2)} & \textbf{Excluded} & \textbf{Sign errors (conf.)} &
\textbf{Corrects (conf.)} & \textbf{Sign errors (raw)} &
\textbf{Corrects (raw)} & \textbf{Wrote ``+''} & \textbf{Wrote
``\ensuremath{-}''}\tabularnewline
\midrule
\endhead
Gemma-3-27B & discovery & 142 & 116 & 26 & 67 & 49 & 92 & 50 & 63 &
79\tabularnewline
Gemma-3-27B & held-out & 264 & 169 & 95 & 49 & 120 & 126 & 138 & 130 &
134\tabularnewline
Gemma-3-27B & pooled & 406 & 285 & 121 & 116 & 169 & 218 & 188 & 193 &
213\tabularnewline
Mistral-Small-24B & discovery & 173 & 107 & 66 & 36 & 71 & 95 & 78 & 41
& 132\tabularnewline
Mistral-Small-24B & held-out & 264 & 170 & 94 & 66 & 104 & 142 & 122 &
50 & 214\tabularnewline
Mistral-Small-24B & pooled & 437 & 277 & 160 & 102 & 175 & 237 & 200 &
91 & 346\tabularnewline
Phi-4-14B & discovery & 111 & 103 & 8 & 55 & 48 & 61 & 50 & 17 &
94\tabularnewline
Phi-4-14B & held-out & 299 & 282 & 17 & 143 & 139 & 150 & 149 & 36 &
263\tabularnewline
Phi-4-14B & pooled & 410 & 385 & 25 & 198 & 187 & 211 & 199 & 53 &
357\tabularnewline
Llama-3.3-70B & discovery & 96 & 91 & 5 & 12 & 79 & 16 & 80 & 27 &
69\tabularnewline
Llama-3.3-70B & held-out & 322 & 265 & 57 & 138 & 127 & 192 & 130 & 122
& 200\tabularnewline
Llama-3.3-70B & pooled & 418 & 356 & 62 & 150 & 206 & 208 & 210 & 149 &
269\tabularnewline
Qwen-2.5-72B & discovery & 106 & 75 & 31 & 43 & 32 & 59 & 47 & 41 &
65\tabularnewline
Qwen-2.5-72B & held-out & 288 & 201 & 87 & 94 & 107 & 147 & 141 & 92 &
196\tabularnewline
Qwen-2.5-72B & pooled & 394 & 276 & 118 & 137 & 139 & 206 & 188 & 133 &
261\tabularnewline
\bottomrule
\end{longtable}

\textbf{(a)} Confirmed: the sign the model wrote leads by \ensuremath{\ge} 2 logits at
the final layer. Excluded cases end with a weak lead (\textless{} 2
logits) or with the opposite sign leading; they are excluded from all
depth statistics and from Figure 1.

\textbf{(b)} Pooled = discovery + held-out sums (descriptive only).
Paper verdicts elsewhere use the held-out phase only.

\textbf{(c)} Mistral-Large-675B (5\ensuremath{\times}5 determinant, behavioral run,
sampled layer grid): 51 sign errors, no corrects harvested; the
confirmation rule was not applied --- its 51 cases are shown unfiltered
in Figure 1, as an exploratory test case only (C.5).

\textbf{(d)} Source exception (Mistral-Small discovery corrects, det and
IBP): the counts use the canonical released files; superseded working
copies of those two sets are divergent probe-pick variants (77/78 and
42/43 cases differ in probe token) and are not used. All other cells are
complete: all five models have determinant error and correct-case data
in both phases.

\textbf{(e)} Corrects' written sign is read from the written\_sign
field, verified on raw data to hold the character the model actually
wrote; three upstream *\_3stages pages display an inverted convention
for corrects (known issue there, not in this paper's data).

\textbf{(f)} The correct-case pools skew toward wrote ``\ensuremath{-}''. Which
problems a model gets right is not under experimental control, so the
wrote-``+'' / wrote-``\ensuremath{-}'' split among corrects is simply the models' own
behavior; it differs across models and does not estimate any base rate.

\textbf{C.4 Where the peaks fall}

Every confirmed case falls into one of two categories:

\begin{itemize}
\item
  \textbf{Late peak (depth \ensuremath{\ge} 0.75)} --- the majority in every model.
  This is the paper's claim; Figure 1 and Figures E1--E2 show it. C.4.1
  reads Figure E1 model by model.
\item
  \textbf{Early peak (depth \textless{} 0.75)} --- a minority, visible
  in Figure 1 as the points left of the dashed line. C.4.2 defines the
  test that sorts them and gives the census; C.4.3 (Qwen-2.5-72B) and
  C.4.4 (Gemma-3-27B) tell each model's full story; C.4.5 covers the
  other three and the residual.
\end{itemize}

\textbf{C.4.1 Late peaks: reading Figure E1}

Each panel of Figure E1 is the median trajectory of Step 4's lead ---
the \emph{written-sign} lead --- across every confirmed case of one
model, errors (red) and corrects (green) separately, with IQR shading.
All panels pool discovery and held-out; the n in each panel matches the
pooled confirmed counts of Table C1. The shared shape is the claim: the
median sits flat, or below zero, through the middle of the network, and
exits steeply only past the 0.75 line. Panel by panel (all values are
medians in logits):

\begin{itemize}
\item
  \textbf{Qwen-2.5-72B --- the extreme case.} The corrects' median is
  not just flat mid-stack, it is \emph{negative}: \ensuremath{-}4.2 at depth 0.50,
  still \ensuremath{-}6.3 at 0.75. The written sign is losing until almost the end
  --- the median first turns and stays positive only at depth 0.94
  (L74), peaking +6.9 at 0.97. Errors drift from +1.9 at mid-depth to a
  +9.6 peak at 0.97. The small non-zero lead both curves show at depth 0
  is the layer-0 embedding lean dissected in C.4.3.
\item
  \textbf{Llama-3.3-70B.} Errors hug zero through mid-depth (\ensuremath{-}0.3 at
  0.50) and end at +7.1; corrects pass +5.6 at 0.75 and surge to +26.0
  at the final layer --- the strongest late climb of the five.
\item
  \textbf{Gemma-3-27B.} Errors: +0.3 at 0.50, +6.2 at 0.75, peak +10.2
  at 0.90. Corrects show the one early bump in Figure E1: a +14.3 peak
  at depth 0.21 --- the early copying band of C.4.4 showing through the
  group median --- which collapses to \ensuremath{-}2.4 at 0.50 before the true late
  climb (+4.4 at 0.75, +8.8 at the end).
\item
  \textbf{Phi-4-14B.} Both curves sit above the noise band already at
  mid-depth (errors +4.2, corrects +3.4 at 0.50) --- the offset that
  marks Phi --- but the commitment still steepens late: errors +11.9 at
  0.75 to a +22.1 peak at 0.92; corrects +8.9 to +16.0.
\item
  \textbf{Mistral-Small-24B --- the cleanest.} Both medians stay inside
  the noise band through 0.75 (errors \ensuremath{-}0.8, corrects \ensuremath{-}0.1) and rise only
  at the very end (final +4.2 and +7.1).
\end{itemize}

One reading covers all five: whatever sign the model is about to write
becomes visible --- and stays visible --- only in the last quarter. Qwen
makes the point most sharply: on its correct cases the written sign is
actively behind until 94\% of the way through the network.

\textbf{C.4.2 Early peaks: are they early decisions?}

Figure 1 shows a handful of confirmed points left of depth 0.75,
concentrated in two models. The question is whether the model actually
decides the sign there. The peak alone cannot answer it: a curve can
spike early, collapse negative through the entire middle of the network,
and only commit at the very end --- the peak metric calls such a case
``early'' while the curves say the commitment is late. Every Qwen early
peak is exactly this (Figure C1 in C.4.3 shows three of them).

\textbf{The crossing depth.} The peak asks: at which layer is the lead
\emph{biggest}? The crossing asks a different question: from which layer
does the lead turn positive and \emph{stay} positive to the end ---
never flipping again? That layer is where the model finally locks in its
sign; divide it by (n\_layers \ensuremath{-} 1) to get the crossing depth. In Figures
C1 and C3 it is drawn as an open circle on each curve: the point where
the curve crosses zero for good. Then the same 0.75 boundary used
everywhere in the paper splits the early peaks into two kinds:

\begin{itemize}
\item
  \textbf{Artifact (crossing \ensuremath{\ge} 0.75):} the early lead collapses
  mid-stack and the real commitment is late --- the early peak was a
  mirage of the metric. Every curve in Figure C1 is of this kind.
\item
  \textbf{Genuine (crossing \textless{} 0.75):} the lead holds from
  mid-depth onward and never goes negative again --- a real early
  commitment, and a real exception to the paper's claim.
\end{itemize}

The crossing is used only in this appendix, not as the paper's main
metric: on oscillating curves the crossing is sensitive to definition,
while the peak is not.

Table C2 gives the complete census: every confirmed early-peak
determinant case, by model and class (discovery + held-out pooled; the
Qwen error column, for example, is 20 held-out + 8 discovery), the named
phenomenon behind it, and the artifact/genuine split. Two models show a
systematic early phenomenon, each with a different mechanism; the other
three do not. C.4.3--C.4.5 take them one at a time.

\textbf{Table C2:} Early-peak census --- phenomenon, counts, and fate
(confirmed determinant cases, pooled).

\begin{longtable}[]{@{}>{\raggedright\arraybackslash}p{0.141\textwidth}>{\raggedright\arraybackslash}p{0.257\textwidth}>{\raggedright\arraybackslash}p{0.108\textwidth}>{\raggedright\arraybackslash}p{0.066\textwidth}>{\raggedright\arraybackslash}p{0.058\textwidth}>{\raggedright\arraybackslash}p{0.124\textwidth}>{\raggedright\arraybackslash}p{0.066\textwidth}>{\raggedright\arraybackslash}p{0.058\textwidth}@{}}
\toprule
\textbf{Model} & \textbf{Phenomenon} & \textbf{Errors: total} &
\textbf{artifact} & \textbf{genuine} & \textbf{Corrects: total} &
\textbf{artifact} & \textbf{genuine}\tabularnewline
\midrule
\endhead
Qwen-2.5-72B & the layer-0 lookup (C.4.3) & 28 & 28 & 0 & 20 & 20 &
0\tabularnewline
Gemma-3-27B & the early copying band (C.4.4) & 33 & 17 & 16 & 109 & 34 &
75\tabularnewline
Phi-4-14B & none --- no shared site (C.4.5) & 6 & 2 & 4 & 25 & 2 &
23\tabularnewline
Llama-3.3-70B & none --- no shared site (C.4.5) & 9 & 5 & 4 & 3 & 0 &
3\tabularnewline
Mistral-Small-24B & none --- no shared site (C.4.5) & 5 & 3 & 2 & 1 & 0
& 1\tabularnewline
\bottomrule
\end{longtable}

\emph{Genuine early errors sum to 26; pooled confirmed errors sum to 703
(Table C1): 26/703 = 3.7\% (C.4.5).}

\textbf{C.4.3 Qwen-2.5-72B: early peak, late commitment --- the layer-0
lookup}

\textbf{What the curves look like.} Figure C1 shows three representative
Qwen early errors, including the case traced in C.2. Every curve has the
same shape: a spike at layer 0 (leads +4.7--4.9, before any layer has
run), a collapse below zero through the entire middle of the network,
and a sharp recovery only at the very end. The open circles mark where
each curve crosses positive for good --- 0.949, 0.975, 0.987 --- and all
20 confirmed held-out Qwen early errors cross in that same narrow window
(0.949--0.987), the corrects in 0.937--0.987. Every one fails the
crossing test of C.4.2: the lead is biggest at layer 0, but it collapses
right after, and only turns positive for good at 95--99\% depth --- the
sign is decided there, not at layer 0. The rest of this subsection
explains where the layer-0 spike comes from, rules out the alternatives,
and follows the phenomenon across domains.

\begin{center}\includegraphics[width=\linewidth]{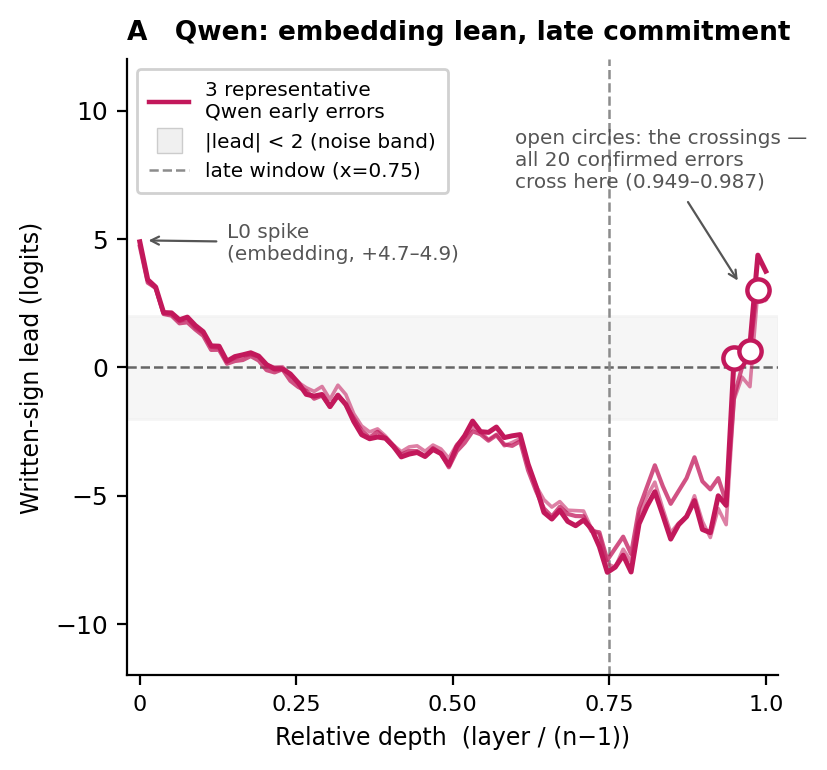}\end{center}

\textbf{Figure C1:} Three representative Qwen-2.5-72B early errors
(NC\_4x4\_det\_222, NC\_4x4\_det\_270, NC\_4x4\_det\_201 --- the case of
C.2), layer-0 leads +4.89, +4.71, +4.82. All three spike at the
embedding, fall negative through the mid layers, and cross back positive
only near depth 0.95--0.99; open circles mark each case's crossing. All
20 confirmed held-out Qwen early errors follow this pattern (crossings
0.949--0.987). Grey band: \textbar lead\textbar{} \textless{} 2 (lens
noise). Dashed line: depth 0.75.

\textbf{Mechanism.} Layer 0 is the embedding: before any layer runs, the
residual at the probe position is nothing but the embedding row of the
token sitting there. Dot that row with the fixed direction
unembed(``\ensuremath{-}'') \ensuremath{-} unembed(``+'') and you get that token's built-in lean
--- a pure table lookup, set by the tokenizer, not computed from the
problem. In Qwen-2.5-72B, 80\% of determinant probes (230 of the 288
held-out cases) sit on ``\})'', the LaTeX close of terms like
\textbackslash det(A\_\{11\}) --- exactly the probe token of the case in
C.2. Its embedding row alone predicts a +4.87 lean toward ``\ensuremath{-}''; the 230
cases probed there average +4.80, errors and corrects alike. In Qwen's
training text, ``\ensuremath{-}'' simply followed ``\})'' more often than ``+'', and
that bigram is baked into the token tables. Figure C2 shows the lookup
directly: group the 288 cases by probe token, and the lean follows the
token, not the problem.

\begin{center}\includegraphics[width=\linewidth]{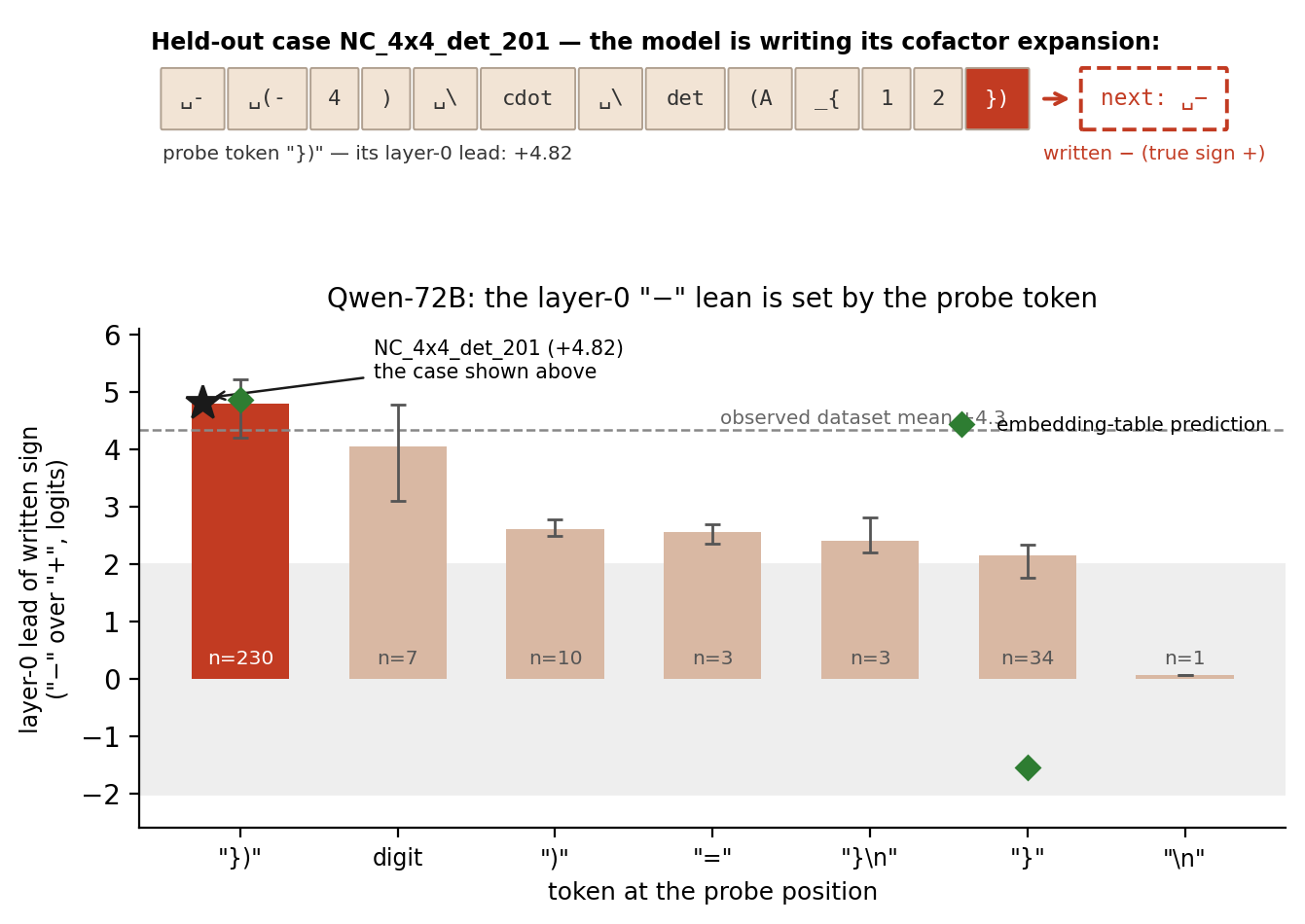}\end{center}

\textbf{Figure C2:} The layer-0 ``\ensuremath{-}'' lean follows the probe token. Mean
layer-0 lead of the written sign for all 288 held-out Qwen-2.5-72B
determinant cases, grouped by the token at the probe position; n per
group, whiskers span min--max. Green diamonds: the pure embedding-table
prediction (no layers run). Grey band \textbar lead\textbar{}
\textless{} 2: lens noise. Dashed line: dataset mean +4.3. Black star:
case NC\_4x4\_det\_201 of C.2 (+4.82) inside its ``\})'' group.

\textbf{Three controls} rule out the alternatives:

\begin{quote}
1. Across \textasciitilde25,000 sampled vocabulary tokens, only 49.6\%
lean ``\ensuremath{-}'' --- the lookup is not a global property of the tables.

2. The average lean over each question's own text is +0.99 --- far short
of the dataset mean +4.3; the context alone does not produce it.

3. Grouping the 288 cases by probe token closes the gap (Figure C2):
``\})'' predicts +4.87, observed +4.80. One secondary context effect
exists --- the ``\}'' group's embedding predicts \ensuremath{-}1.54 yet observes
+2.15, so block-0 attention over the minus-heavy context adds a few
logits --- but it never drives the headline number.
\end{quote}

\textbf{What it does to the counts.} Held-out: 48 of 147 determinant
errors peak at layer 0; 20 pass the confirmation rule (the other 28 end
with too weak a final lead), and 45 of 141 corrects peak there, 17
confirmed. Pooled with discovery, the confirmed early set is 28 errors
and 20 corrects --- every one an artifact by the crossings above (Table
C2). The identical lean in corrects is the point: it is a prior the
token carries in, not a decision. The late layers wipe it out and
re-commit at 95--99\% depth.

\textbf{Across domains.} A weight constant travels with the tokens. On
the shared 48-problem arithmetic control (Appendix D), Qwen-2.5-72B
leans ``\ensuremath{-}'' on 48/48 problems (mean +2.89) while Llama-3.3-70B on the
identical problems shows none (mean +0.19). On the 76 held-out IBP cases
the layer-0 lean is weaker (mean +2.8) --- too weak to outrank a real
late peak, which is why the lookup is invisible in the IBP data.
Verdict: none of Qwen's early peaks is a genuinely early sign decision.

\textbf{C.4.4 Gemma-3-27B: the early copying band}

\textbf{Mechanism.} Gemma's early lean is a different animal: it does
not sit at layer 0 and it is not a token-table lookup. Its layer-0
values are at chance --- positive on only 14 of the 37 lean-test cases.
Instead, Gemma pre-loads ``\ensuremath{-}'' in an early band of layers: every one of
the 37 lean-test cases leans ``\ensuremath{-}'' somewhere in layers 8--13, and 32 of
the 33 confirmed early cases peak at layers 5--13 (one at layer 45)
(Figure C3). The lean is graded by what the model just wrote: +16.6 when
a literal ``\ensuremath{-}'' sits within 5 characters of the probe, down to +2--6
after closing braces, and it grows with the count of nearby minus
characters (r = +0.42). Those characters are the model's own expansion
text --- negative entries, operands, intermediate results. This is
shallow copying of the model's own recent minus ink, erased by about
layer 15, well before the habit layers L58/L60 (\S{}4) act. (The lean test
is a 37-case subset --- indicative, not tight.)

\textbf{Two fates.} Unlike Qwen, Gemma's early band does not always get
erased. Figure C3 shows both outcomes. Artifact: cases
NC\_4x4\_det\_1138 and NC2\_4x4\_det\_0401 (orange, crossing 0.951 each)
spike in the band, fall below zero, and re-commit late --- the late
layers overwrite the early guess. Genuine: cases NC2\_4x4\_det\_0848 and
NC2\_4x4\_det\_0944 (green, crossings 0.590 and 0.574) hold a positive
lead from mid-depth onward --- the early guess survives to the end.

\begin{center}\includegraphics[width=\linewidth]{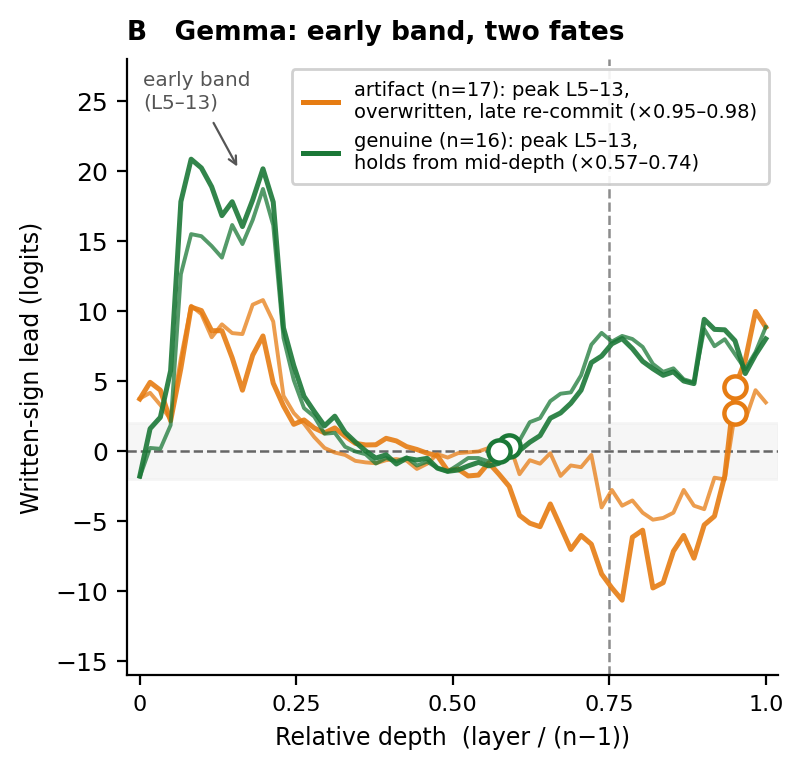}\end{center}

\textbf{Figure C3:} Gemma-3-27B: the early band and its two fates. Two
artifact cases (NC\_4x4\_det\_1138, NC2\_4x4\_det\_0401; both crossing
0.951, orange) spike at the band and are overwritten --- the lead goes
negative and only re-commits late. Two genuine cases
(NC2\_4x4\_det\_0848, NC2\_4x4\_det\_0944; crossings 0.590 and 0.574,
green) hold a positive lead from mid-depth onward. Open circles:
crossings. Grey band: \textbar lead\textbar{} \textless{} 2. Dashed
line: depth 0.75.

\textbf{Counts.} Of the 33 confirmed early errors, 17 are artifacts
(crossing 0.951--0.984) and 16 are genuine (crossing 0.000--0.738,
median 0.631). Of the 109 confirmed early corrects, 34 are artifacts and
75 are genuine (median crossing 0.672) --- the same band appears in
corrects too. The asymmetry is the reading: when the early guess is
right it usually survives (75 of 109); when it is wrong it is usually
overwritten and re-committed late (17 of 33). Gemma-3-27B has a real
early-commitment signature --- and it is the primary source of the
genuine-early residual below.

\textbf{C.4.5 The other three models, and the residual}

Phi-4-14B, Llama-3.3-70B and Mistral-Small-24B contribute 6, 9 and 5
confirmed early errors (artifact/genuine 2/4, 5/4 and 3/2) --- no
layer-0 lean, no early band, no shared site (Table C2). With so few
cases and no pattern, there is no mechanism to analyse.

\textbf{The bottom line.} Across all five models, genuinely early error
commitments are 26 of 703 confirmed errors (3.7\%) --- the limit of the
late-decision account, stated plainly. The visual evidence is Figures C1
and C3: every Qwen early peak is an artifact of the embedding lookup,
and Gemma's genuine early commits (Figure C3, green) are the primary
source of the 26.

\textbf{C.5 The exploratory sixth model (Mistral-Large-675B)}

Mistral-Large-675B was run on the 5\ensuremath{\times}5 determinant task as a
behavioral-only exploratory probe: the causal battery (ablations,
patching, dose--response steering) was not run, because at 675 billion
parameters the compute cost is prohibitive. The model appears only in
Figure 1a as grey open circles and in Table C1 footnote (c): the
confirmation rule was not applied, and its 51 cases are shown
unfiltered.

The probe covered 51 validated sign-error cases; the logit-lens readout
and direct logit attribution ran on all 51. Of these, 39 peak in the
last quarter of the network (relative depth \ensuremath{\ge} 0.75 under the convention
layer / (61 \ensuremath{-} 1), i.e. \ensuremath{\ge} L45) --- consistent with the late-decision
pattern of the five core models.

\textbf{Table C3:} Mistral-Large-675B stage accounting.

\begin{longtable}[]{@{}ll@{}}
\toprule
\textbf{Stage} & \textbf{Count}\tabularnewline
\midrule
\endhead
Sign-error cases (experiment-ready) & 51\tabularnewline
Logit-lens readout run & 51\tabularnewline
Direct logit attribution run & 51\tabularnewline
Peak in last quarter of depth (\ensuremath{\ge} 0.75, i.e. \ensuremath{\ge} L45) & 39\tabularnewline
Causal battery (ablations, patching, steering) & not run\tabularnewline
\bottomrule
\end{longtable}

\clearpage
\section*{Appendix D --- The arithmetic control}
\addcontentsline{toc}{section}{Appendix D --- The arithmetic control}

\emph{The same late-commitment test, on the plainest possible task ---
and on completions where the model writes the right answer.}

\textbf{D.1 What the control tests}

\textbf{Question.} Does late commitment also appear on plain arithmetic
--- a setting far from the algebraic structure of the determinant and
IBP tasks --- and does it appear when the model is right, not only when
it errs?

\textbf{The probe set.} 48 short subtraction problems, identical for all
five models. Each problem reads, verbatim: ``Compute 476 \ensuremath{-} 675. Show all
steps. Write your final numerical answer with its sign (use '+' if the
result is positive, '-' if negative).'' The true signs are mixed by
design: 25 problems have a negative answer, 23 a positive one. Each
model generates its own worked solution, and the logit lens then tracks
the model's lead for the sign it wrote, at the written-sign position ---
the same readout as everywhere else in the paper.

\textbf{These are correct completions, not errors.} On this task the
models are almost always right: 237 of the 240 completions (5 models \ensuremath{\times}
48 problems) carry the correct written sign. Phi, Mistral-Small-24B,
Llama, and Qwen are 48/48 correct; Gemma wrote the wrong sign in 3 of
its 48 cases (noted in D.2). There is therefore no error/correct split
to make here --- the control measures commitment depth on (almost
entirely) correct arithmetic.

\textbf{Rules.} The set is a single pooled collection; there is no
discovery/held-out split. The same confirmation rule applies
(written-sign lead \ensuremath{\ge} +2 logits at the final layer), and the same late
threshold (relative depth \ensuremath{\ge} 0.75). All 240 cases pass contamination
checks.

\textbf{D.2 Results}

\textbf{Results.} Every single case across all 240 is confirmed --- no
case is excluded. The weakest final lead in the entire set is +6 logits.
Four of five models commit at the very end of the residual stream: Phi,
Mistral-Small-24B, Llama, and Qwen each produce 48/48 late cases (100\%)
with a median rel\_depth of 1.00. The majority of those cases peak at
the final layer itself: 36 of 48 for Phi, 46 of 48 for Mistral-S, 48 of
48 for Llama, and 38 of 48 for Qwen --- not just late, but last.

Gemma is again the exception. Only 22 of 48 Gemma cases (46\%) are late,
and the median relative depth is 0.082. The distribution is not diffuse:
it is bimodal. Twenty-five cases peak at layer 5 (of 62), and 21 peak at
layer 61, with 2 cases in between. This bimodal pattern is consistent
with the early-copying band documented in Appendix C.4.4 for the determinant
domain. In that appendix, Gemma's early-error peaks are attributed to an
L8--13 shallow-copy mechanism; the arithmetic control's layer-5 cluster
may reflect a similar or related phenomenon, though the arithmetic set
of 48 cases is too small and structurally too different from the
determinant cases to test the mechanism. We report the bimodal split as
indicative, not as a claim about mechanism.

Gemma's 3 wrong-sign cases are disclosed rather than dropped: one peaks
at layer 12 and two at layer 61. Removing them changes nothing material
--- the bimodal split remains (25 cases at layer 5, 19 at layer 61).

\begin{center}\includegraphics[width=\linewidth]{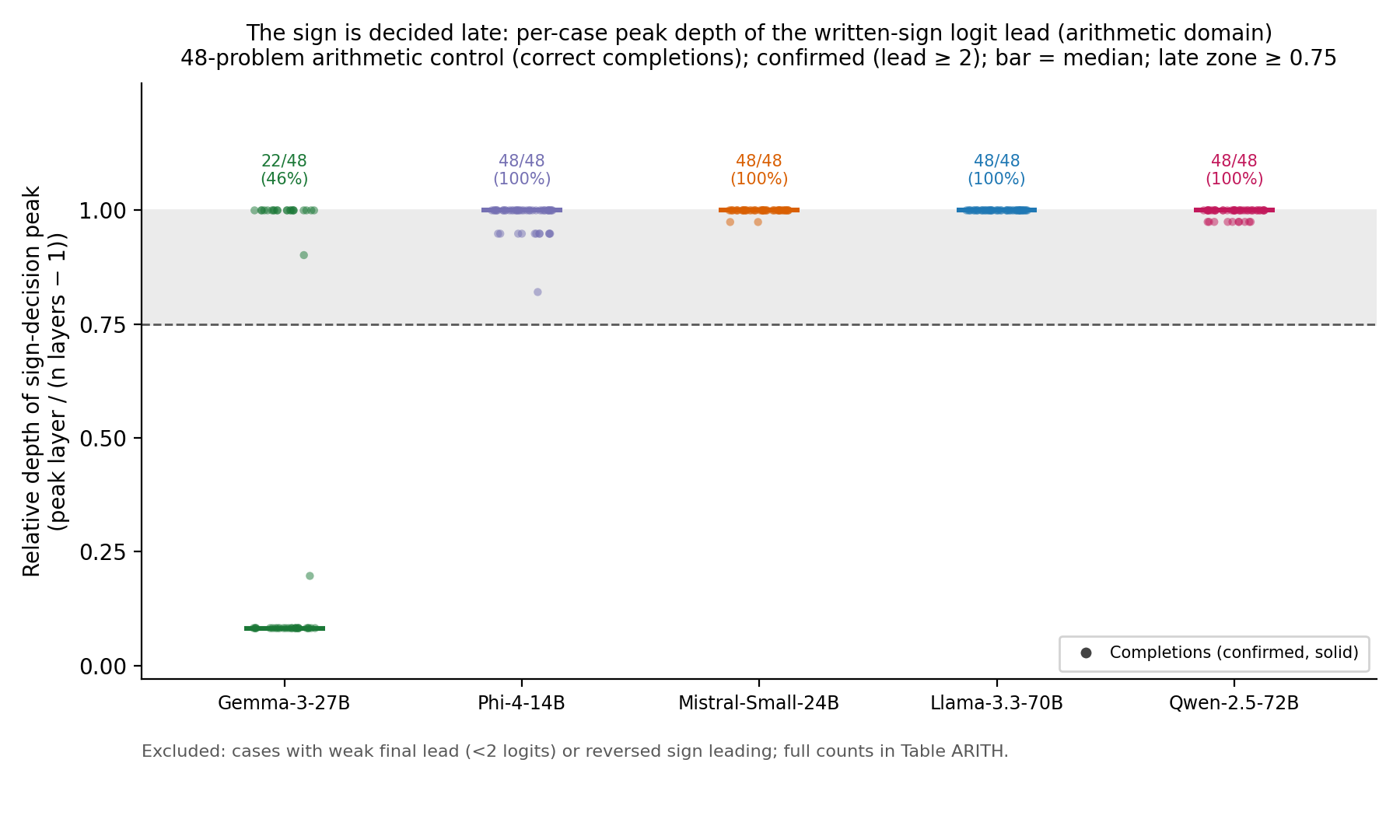}\end{center}

\textbf{Figure D1: Arithmetic control --- late-decision depth
distribution.} Each panel shows one model's 48 confirmed
arithmetic-control cases (solid dots) by peak relative depth (x-axis, 0
= first layer, 1 = last layer). 237 of the 240 completions carry the
correct written sign (Gemma's 3 exceptions are noted in the text).
Horizontal bar: per-model median. Dashed vertical line: rel\_depth =
0.75. No cases are excluded (minimum final lead across all 240 cases is
+6 logits). Phi, Mistral-S, Llama, and Qwen cluster tightly at
rel\_depth = 1.0 (medians all 1.00; final-layer peak counts: Phi 36/48,
Mistral-S 46/48, Llama 48/48, Qwen 38/48). Gemma displays the bimodal
split described in the text: 25 cases at layer 5 and 21 at layer 61,
visible here as two distinct clusters at either end of the depth axis.

Counts. Table D1 gives the per-model census --- confirmed,
excluded, and late counts, with the median relative depth --- for all
240 completions.

\begin{longtable}[]{@{}lllllll@{}}
\toprule
\textbf{Model} & \textbf{Total} & \textbf{Confirmed} & \textbf{Excluded}
& \textbf{Late} & \textbf{Late \%} & \textbf{Median
rel\_depth}\tabularnewline
\midrule
\endhead
Gemma-3-27B & 48 & 48 & 0 & 22 & 46\% & 0.082\tabularnewline
Phi-4-14B & 48 & 48 & 0 & 48 & 100\% & 1.00\tabularnewline
Mistral-Small-24B & 48 & 48 & 0 & 48 & 100\% & 1.00\tabularnewline
Llama-3.3-70B & 48 & 48 & 0 & 48 & 100\% & 1.00\tabularnewline
Qwen-2.5-72B & 48 & 48 & 0 & 48 & 100\% & 1.00\tabularnewline
\bottomrule
\end{longtable}

\textbf{Reading the table.} Take the Gemma-3-27B row: all 48 of its
arithmetic cases pass the confirmation rule (48 confirmed, 0 excluded),
but only 22 of the 48 (46\%) peak late, and the median relative depth is
0.082 --- the bimodal split described above. Every other row reads 48
confirmed, 0 excluded, 48 late (100\%), median 1.00: on plain arithmetic
those four models commit at the very end of the residual stream.

\textbf{D.3 What the control shows}

Two things. First, late commitment is not an artifact of the algebra
tasks --- and not a property of errors: even when writing the correct
sign of a plain subtraction, four of the five models still commit at the
very end of the residual stream, most often at the final layer itself.
Second, the control separates weight constants from decisions. Appendix
C.4.3 uses these same 48 problems to test whether Qwen-2.5-72B's layer-0
probe-token lookup is a fixed property of its embedding tables: all 48
Qwen cases lean ``\ensuremath{-}'' at layer 0 (mean +2.89), while Llama-3.3-70B on
the identical problems shows no such lean (mean +0.19). The lookup
travels with the weights, not with the task (Appendix C.4.3).

Data. The shared problem set is
pipeline\_release/data\_shared/arith\_control\_problems.json; the
per-case records behind this control are
\{Model\}/arith\_control/expA\_logit\_lens.json in the released
repository (stage s00 generates each model's own solutions; s01 is the
lens readout). Figure provenance: ledger-fed final figures
(table1\_master.csv, fig1\_counts.csv, fig1\_late\_decision.png) in
the figures/ directory.

\clearpage
\section*{Appendix E --- Full readout results: determinant and IBP}
\addcontentsline{toc}{section}{Appendix E --- Full readout results: determinant and IBP}

The observational record for both tasks, standalone: trajectories,
per-layer DLA, and judge-layer curves for the determinant and for IBP,
plus the IBP case accounting. Method reminders are brief; the causal
experiments stay in \S{}5 (Appendices H, I).

\textbf{E.1 Scope and method}

This appendix is the observational record behind \S{}3--\S{}4, both tasks.
Determinant: the full trajectories (Figure E1), full-depth DLA (Figure
E3), and judge curves (Figure E5) that the main text summarizes. IBP:
the complete observational results --- accounting (Table E1),
trajectories (Figure E2), DLA (Figure E4), judge curves (Figure E6) ---
and the two-domain role comparison (Figure E7, Table E2). The
replication question: does the late-commitment pattern of \S{}3 hold on a
second domain? We run the same readout stages on integration by parts
(IBP), a structurally different sign-tracking task.

What is IBP. IBP applies \ensuremath{\int}u dv = uv \ensuremath{-} \ensuremath{\int}v du. The sign error: the model
writes uv + \ensuremath{\int}v du, losing the minus. Every later step runs on the
corrupted expression. We ask where that one sign token is committed.

Method. Identical to the determinant (Appendix C.1): same logit-lens
readout, same confirmation rule (written-sign lead \ensuremath{\ge} +2 at the final
layer), same late threshold (rel\_depth \ensuremath{\ge} 0.75). Discovery and held-out
sets are pooled throughout. Mistral-Large-675B has no IBP data. Bank
construction: A.2.

\textbf{E.2 Case accounting --- IBP}

Determinant accounting is Table C1 (Appendix C.3). Table E1 is its IBP
counterpart, counted the same way: totals, confirmed, excluded, error
and correct counts, and the wrote-sign split. Bold pooled rows are the
counts the figures use; footnote (e) maps columns to figures.

Table E1: IBP case accounting per model and evaluation set. Bold rows
(pooled) feed the figures; see footnote (e).

\begin{longtable}[]{@{}lllllllllll@{}}
\toprule
\textbf{Model} & \textbf{Phase} & \textbf{Total} & \textbf{Conf.} &
\textbf{Excl.} & \textbf{Err (conf.)} & \textbf{Corr (conf.)} &
\textbf{Err} & \textbf{Corr} & \textbf{Wrote ``+''} & \textbf{Wrote
``\ensuremath{-}''}\tabularnewline
\midrule
\endhead
Gemma & disc. & 120 & 106 & 14 & 56 & 50 & 70 & 50 & 60 &
60\tabularnewline
Gemma & held-out & 223 & 192 & 31 & 45 & 147 & 73 & 150 & 44 &
179\tabularnewline
\textbf{Gemma} & \textbf{pooled} & \textbf{343} & \textbf{298} &
\textbf{45} & \textbf{101} & \textbf{197} & \textbf{143} & \textbf{200}
& \textbf{104} & \textbf{239}\tabularnewline
Mistral & disc. & 82 & 69 & 13 & 29 & 40 & 39 & 43 & 20 &
62\tabularnewline
Mistral & held-out & 226 & 181 & 45 & 31 & 150 & 70 & 156 & 76 &
150\tabularnewline
\textbf{Mistral} & \textbf{pooled} & \textbf{308} & \textbf{250} &
\textbf{58} & \textbf{60} & \textbf{190} & \textbf{109} & \textbf{199} &
\textbf{96} & \textbf{212}\tabularnewline
Phi & disc. & 90 & 86 & 4 & 36 & 50 & 40 & 50 & 13 & 77\tabularnewline
Phi & held-out & 223 & 204 & 19 & 58 & 146 & 77 & 146 & 61 &
162\tabularnewline
\textbf{Phi} & \textbf{pooled} & \textbf{313} & \textbf{290} &
\textbf{23} & \textbf{94} & \textbf{196} & \textbf{117} & \textbf{196} &
\textbf{74} & \textbf{239}\tabularnewline
Llama & disc. & 45 & 37 & 8 & 14 & 23 & 22 & 23 & 7 & 38\tabularnewline
Llama & held-out & 68 & 65 & 3 & 18 & 47 & 21 & 47 & 52 &
16\tabularnewline
\textbf{Llama} & \textbf{pooled} & \textbf{113} & \textbf{102} &
\textbf{11} & \textbf{32} & \textbf{70} & \textbf{43} & \textbf{70} &
\textbf{59} & \textbf{54}\tabularnewline
Qwen & disc. & 35 & 35 & 0 & 10 & 25 & 10 & 25 & 0 & 35\tabularnewline
Qwen & held-out & 76 & 73 & 3 & 23 & 50 & 26 & 50 & 0 &
76\tabularnewline
\textbf{Qwen} & \textbf{pooled} & \textbf{111} & \textbf{108} &
\textbf{3} & \textbf{33} & \textbf{75} & \textbf{36} & \textbf{75} &
\textbf{0} & \textbf{111}\tabularnewline
\bottomrule
\end{longtable}

(a) Confirmed: the written sign leads by \ensuremath{\ge} 2 logits at the final layer.
Excluded cases end weak or opposite; they leave all depth statistics.

(b) Pooled = discovery + held-out sums (descriptive). Paper verdicts
elsewhere use held-out only.

(c) Gemma = Gemma-3-27B; Mistral = Mistral-Small-24B; Phi = Phi-4-14B;
Llama = Llama-3.3-70B; Qwen = Qwen-2.5-72B.

(d) Late = rel\_depth \ensuremath{\ge} 0.75. Source exception (Mistral discovery
corrects): counts use the canonical released file; the superseded
working copy is a divergent probe-pick variant (42/43 cases differ) and
is not used. Same for determinant discovery corrects (77/78; Table C1
footnote (d), Appendix C).

(e) Which number feeds which figure. Figure E2: pooled Err (conf.) /
Corr (conf.) --- e.g. Gemma 101 / 197. Figure E4: pooled Err / Corr ---
e.g. Gemma 143 / 200. Figure E7 (IBP rows): the same pooled Err / Corr
cases split by written sign; each panel's n+/n\ensuremath{-} sums to its class column
(e.g. Gemma errors 79+64 = 143), not to the all-case Wrote columns.

\textbf{E.3 Trajectories --- determinant and IBP}

Determinant first. Figure E1 is the full path behind the peaks of \S{}3:
median written-sign lead at every relative depth, errors against
corrects, pooled confirmed cases. All five models share one shape ---
the lead sits inside the noise band, or below zero, through the middle
of the network and exits steeply past depth 0.75. Qwen is the sharpest:
on its corrects the median lead is \ensuremath{-}4.2 at depth 0.50, \ensuremath{-}6.3 at 0.75, and
first turns positive for good at depth 0.94 (layer 74). Llama's corrects
end at +26.0. Gemma's corrects show the early copying bump of \S{}3 (+14.3
at depth 0.21) before dipping to \ensuremath{-}2.4 at mid-depth. Phi's curves ride
slightly above the band through the middle (errors +4.2, corrects +3.4
at depth 0.50); Mistral is the cleanest.

\begin{center}\includegraphics[width=\linewidth]{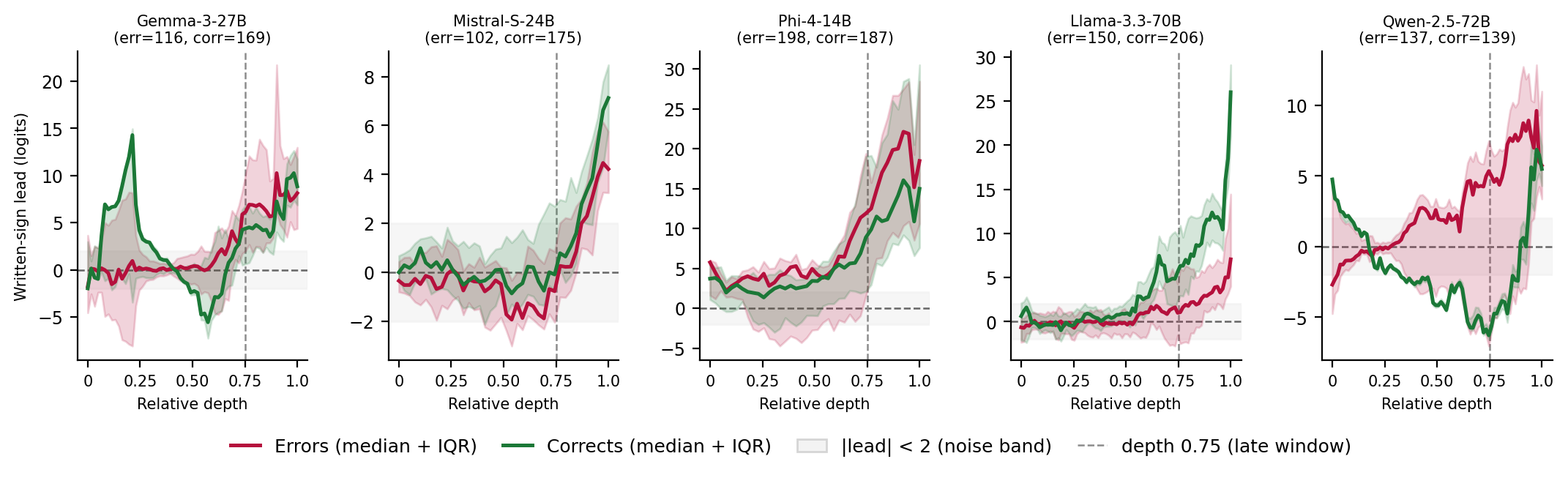}\end{center}

Figure E1: Median written-sign-lead trajectories --- determinant. Median
lead (logits) vs relative depth, pooled confirmed cases, IQR shading.
Crimson: errors; green: corrects. Grey band: \textbar lead\textbar{}
\textless{} 2. Dashed line: depth 0.75. Counts (error / correct): Gemma
116/169, Mistral 102/175, Phi 198/187, Llama 150/206, Qwen 137/139.

IBP next. Figure E2 reads like Figure E1. All five models show the same
shape: flat through mid-depth, steep exit after 0.75.

Two contrasts with the determinant. First, Gemma's corrects no longer
rise early: median peak depth 0.98 here versus 0.21 on determinants
(36\% late there, 94\% late here). The early-correct pattern is
determinant-specific. Second, Phi's above-band offset is gone: both
curves sit in the band through mid-depth (errors median +0.13 at depth
0.51).

\begin{center}\includegraphics[width=\linewidth]{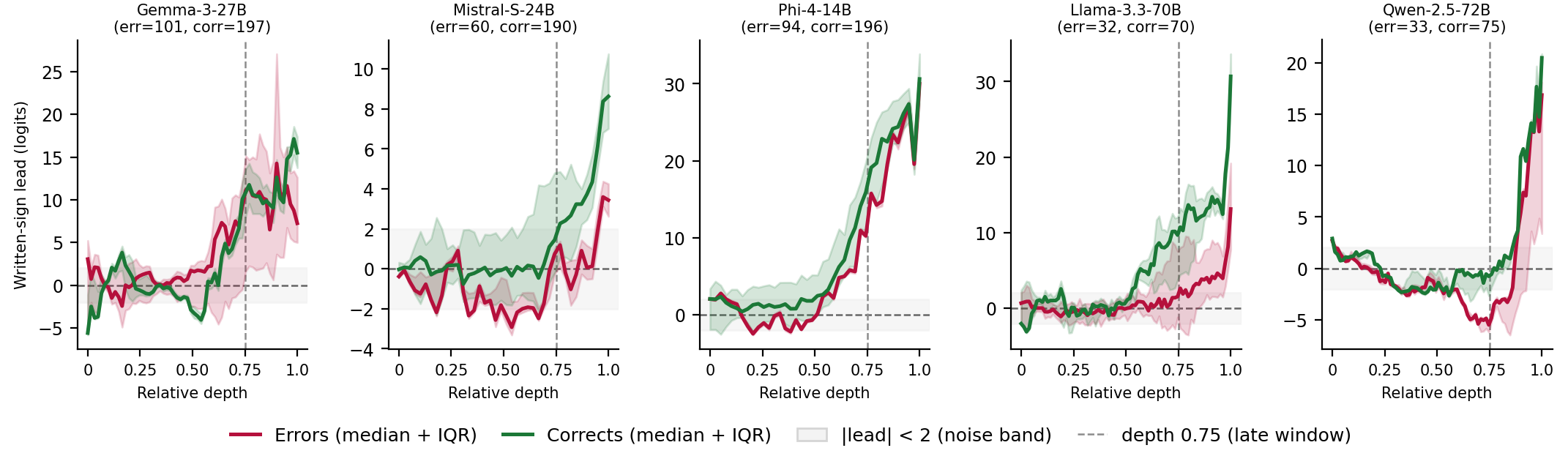}\end{center}

Figure E2: Median written-sign-lead trajectories --- IBP. Layout as
Figure E1: median lead (logits) vs relative depth, pooled confirmed
cases, IQR shading. Crimson: errors; green: corrects. Grey band:
\textbar lead\textbar{} \textless{} 2. Dashed line: depth 0.75. Counts
(error / correct): Gemma 101/197, Mistral 60/190, Phi 94/196, Llama
32/70, Qwen 33/75.

Lateness is sharper than the determinant result: Mistral errors 95\%
late, Phi 100\%, Llama 97\%, Qwen 100\%. Gemma errors 70\% late ---
lowest, still a clear majority. Lateness characterizes sign-writing
itself, in both classes.

Raw per-case records behind the trajectory and DLA figures are released
in the repository (stages s01--s02, all five models, both phases, both
classes); the judge-curve records are the released s09 files (Appendix
B). Inventory with checksums: ledger\_manifest\_full.csv.

\textbf{E.4 Habit-layer DLA}

Computed as Figure 2 (\S{}4). Determinant first: Figure E3 shows the full
depth axis --- the signal concentrates in the last layers, at the
gold-banded habit layers; outside the late window, activity is scattered
and small. Errors and corrects rise at the same layers.

\begin{center}\includegraphics[width=\linewidth]{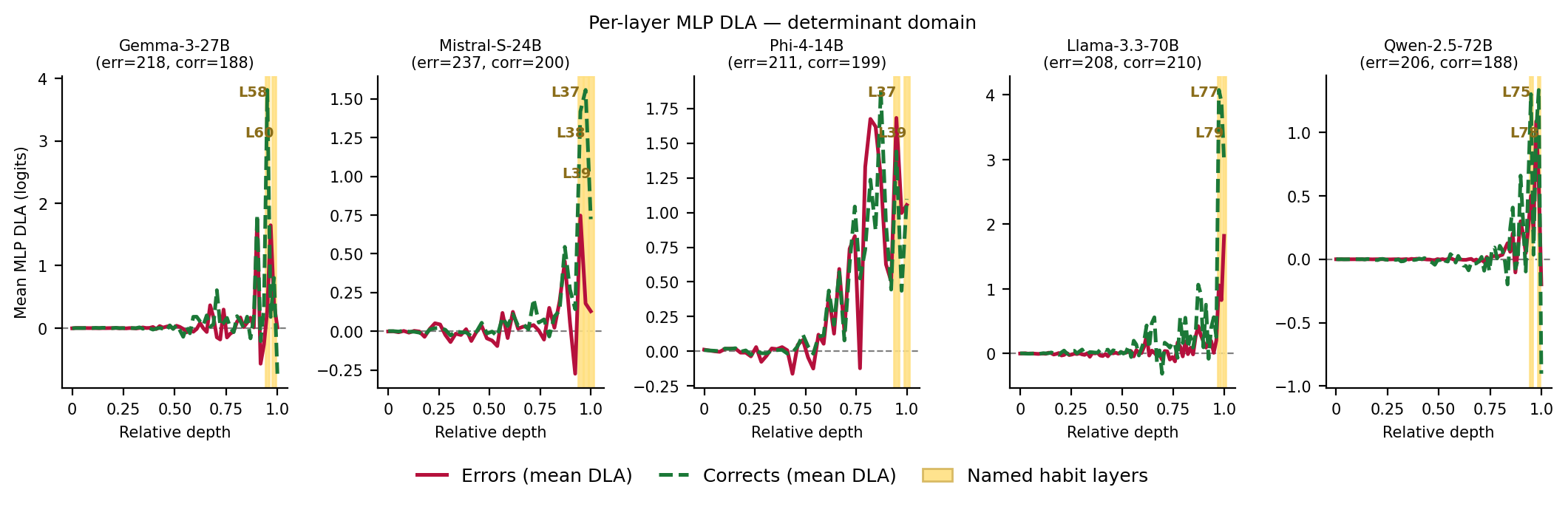}\end{center}

Figure E3: Per-layer MLP DLA, determinant. Mean direct contribution of
each layer's MLP to the written-sign lead, full depth axis. Crimson
solid: errors; green dashed: corrects. Gold bands, bold labels: habit
layers from \S{}4 --- Gemma L58/L60; Mistral L37/L38/L39; Phi L37/L39;
Llama L77/L79; Qwen L75/L78. Counts in panel headers (err / corr): all
cases in the pooled DLA files, discovery + held-out.

IBP next: Figure E4 repeats the readout --- the habit-layer structure
replicates, errors and corrects rising at the same layers.

\begin{center}\includegraphics[width=\linewidth]{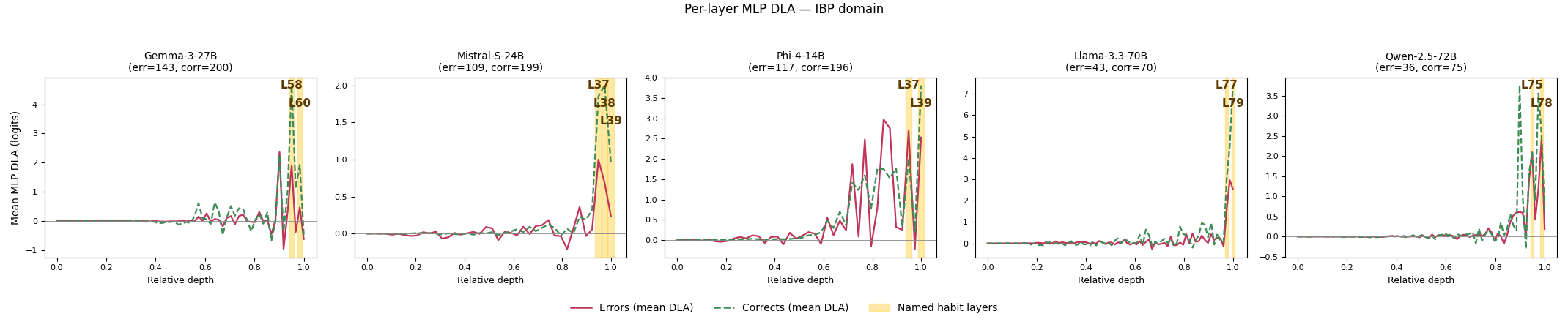}\end{center}

Figure E4: Per-layer MLP DLA, IBP. Layout as Figure E3: mean direct
contribution of each layer's MLP to the written-sign lead. Crimson
solid: errors; green dashed: corrects. Gold bands, bold labels: habit
layers from \S{}4 --- Gemma L58/L60; Mistral L37/L38/L39; Phi L37/L39;
Llama L77/L79; Qwen L75/L78. Pooled counts in Table E1.

\textbf{E.5 Judge-layer curves}

Per-layer lead of the correct sign on error cases --- the judge readout
of \S{}4. Only Qwen and Llama are shown --- the two mechanistic poles.
Determinant first (Figure E5): here the judge fights --- Qwen's curve
zigzags through its late window, stamp pushes down and partial rescues
pull back (annotated on the figure), while Llama slides monotonically at
its amplifier L77--79. Judge files exist for all five models (Qwen IBP:
held-out only); the other three are summarized as flip counts in the
released s09 judge files (Appendix B.3).

\begin{center}\includegraphics[width=\linewidth]{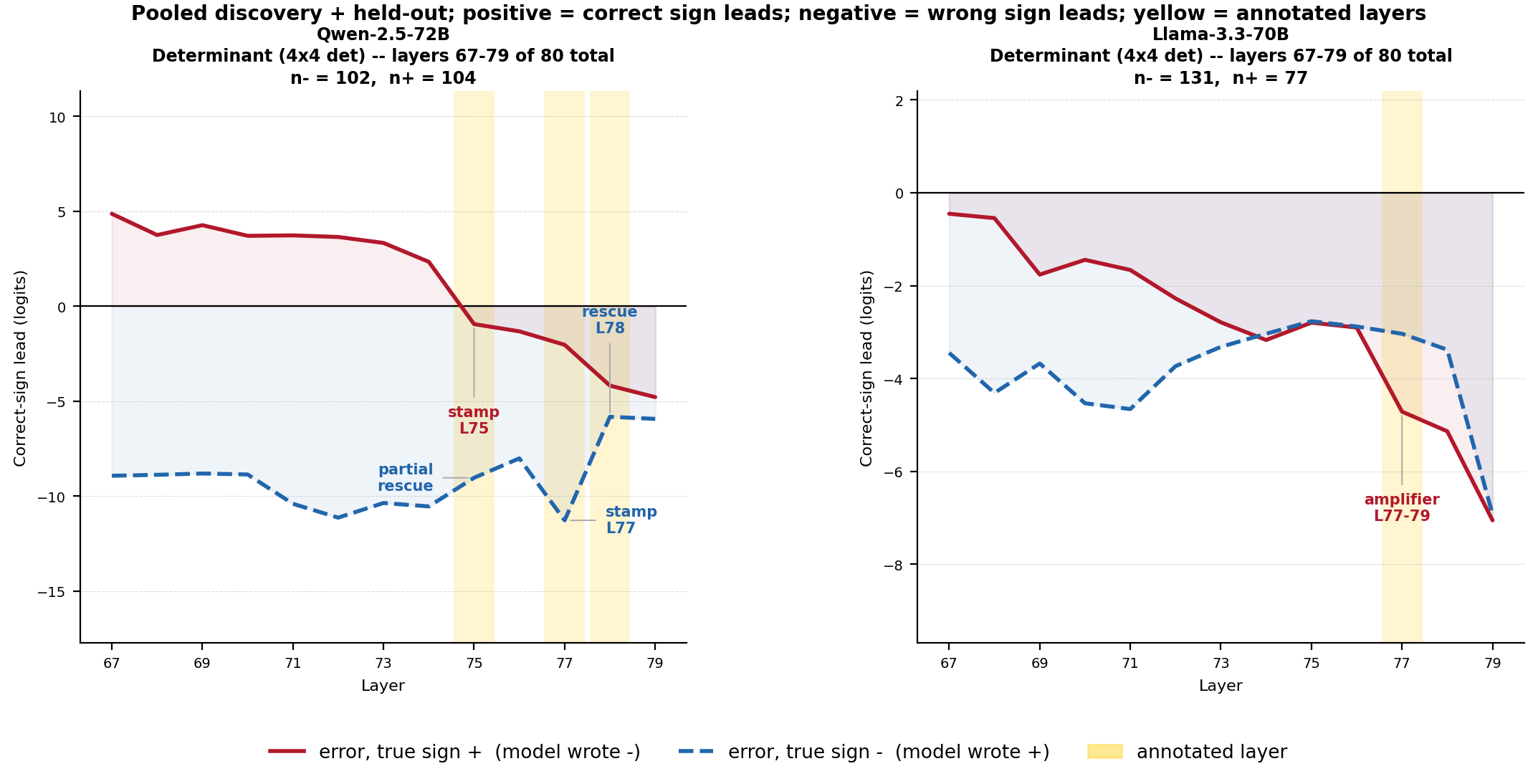}\end{center}

Figure E5: The judge --- determinant (layers 67--79 of 80). Per-layer
lead of the correct sign on pooled determinant errors. Red solid: true +
(model wrote \ensuremath{-}); blue dashed: true \ensuremath{-} (model wrote +). Qwen n\ensuremath{-}/n+ =
102/104; Llama 131/77. Yellow bands: annotated layers.

IBP next (Figure E6): the same readout on IBP errors.

\begin{center}\includegraphics[width=\linewidth]{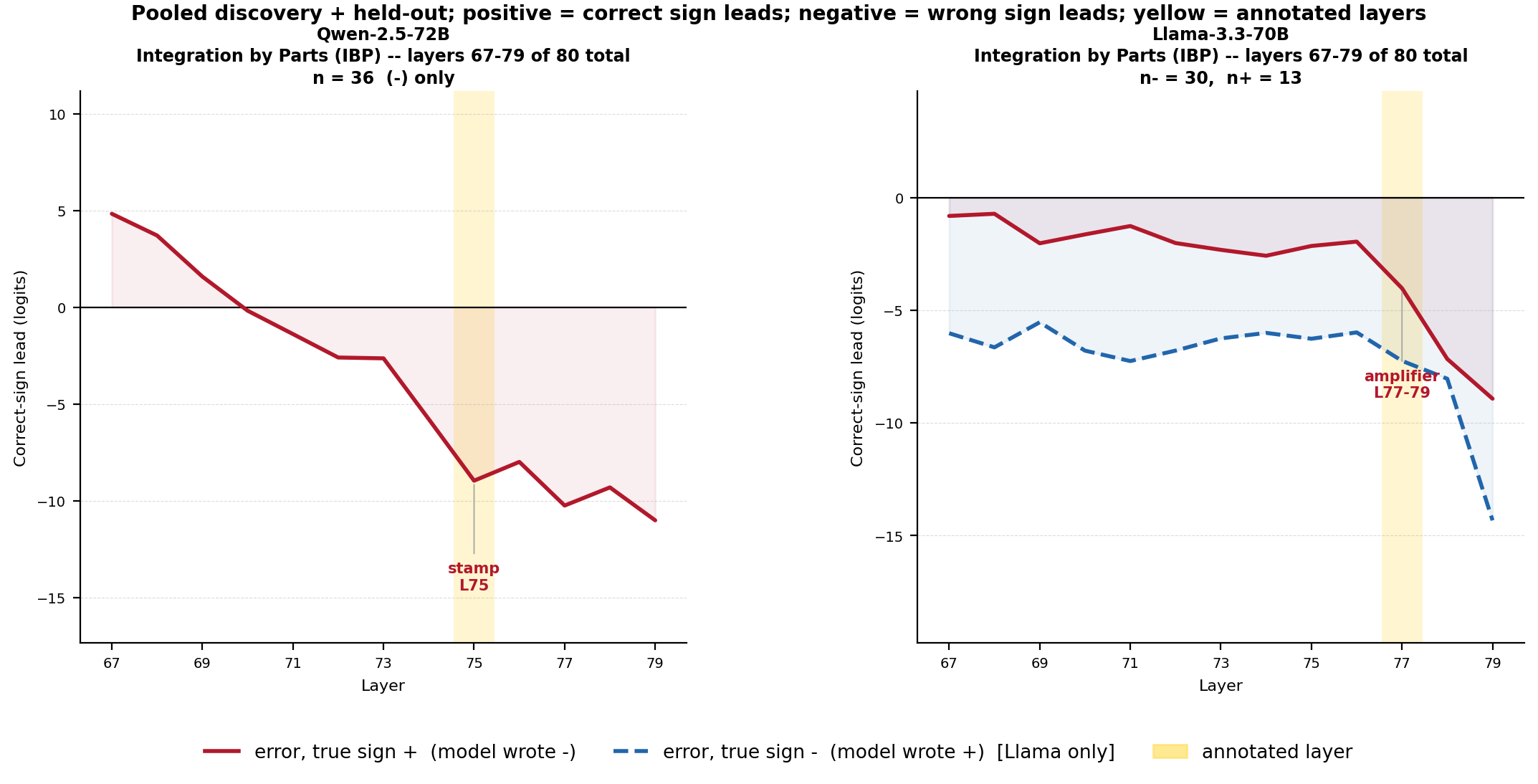}\end{center}

Figure E6: The judge that loses --- IBP (layers 67--79 of 80). Per-layer
lead of the correct sign on pooled IBP errors. Red solid: true + (model
wrote \ensuremath{-}); blue dashed: true \ensuremath{-} (model wrote +). Qwen n\ensuremath{-}/n+ = 36/0 --- no
wrong + written, so no blue curve; Llama 30/13. Yellow bands: annotated
layers.

Qwen: a monotone slide to about \ensuremath{-}11 through L75, with no rescue zigzag
--- on IBP the judge does not fight. Llama: both curves slide down
through L77--L79, amplifying an earlier decision. The timing matches the
determinant; the roles do not (Section E.6).

\textbf{E.6 Comparison --- stamps and amplifiers, errors versus
corrects, both domains}

Sign-split DLA as in Figure 2 (\S{}4), both classes and both domains. The
error rows are cases where the model wrote the wrong sign; the corrects
rows repeat the identical analysis on correct answers. Figure E7 shows
all four blocks; Table E2 gives the statistics behind each verdict.

\begin{center}\includegraphics[width=\linewidth]{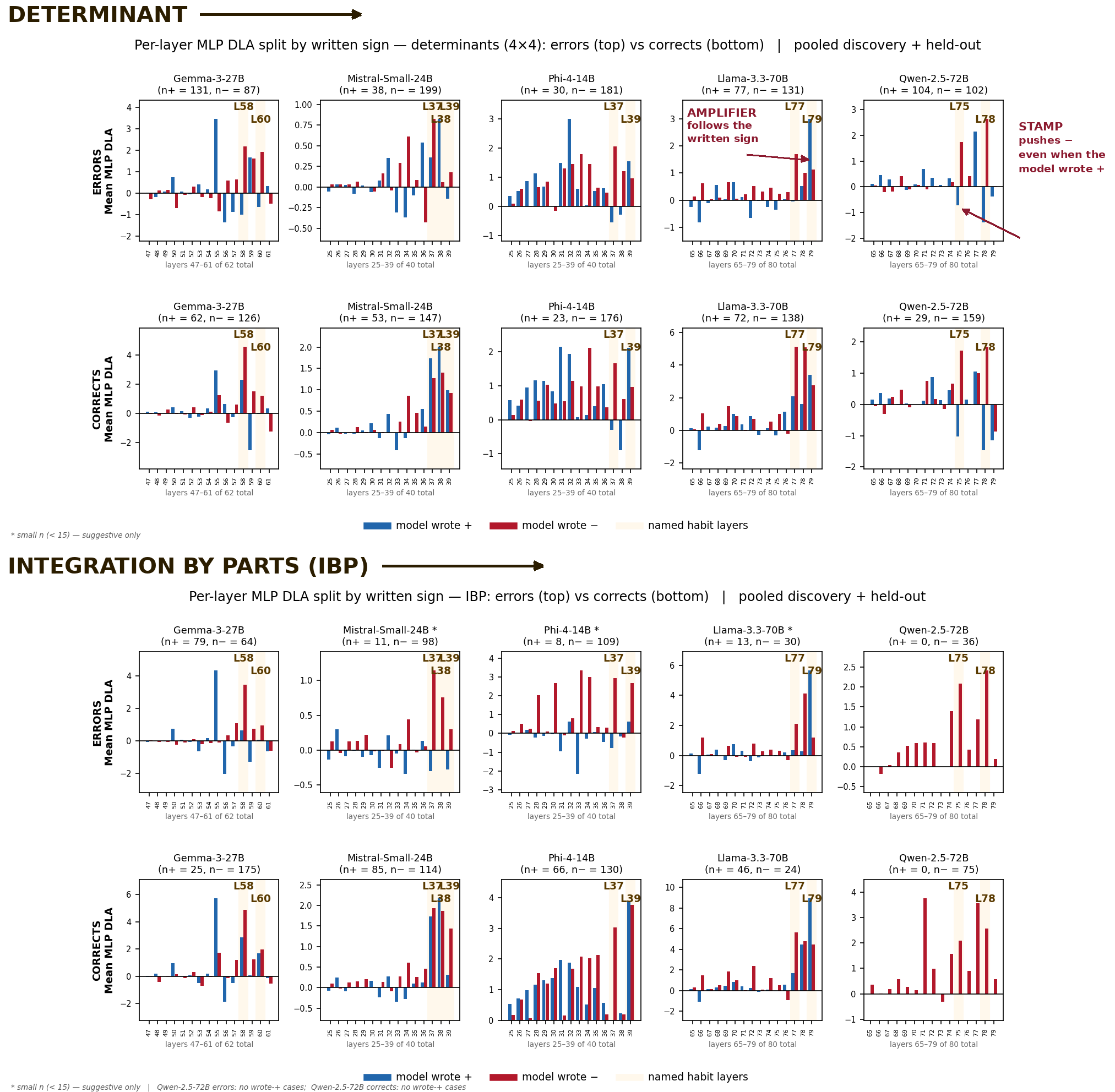}\end{center}

Figure E7: Stamper: pushes \ensuremath{-} no matter which sign the model
wrote. Amplifier: pushes whichever sign the model wrote, further in that
direction. Judge: a late layer that pushes back toward the correct sign
on error cases. Per-layer MLP DLA split by written sign --- determinant
and IBP, errors and corrects, all five models, pooled discovery +
held-out. Blue: wrote +; red: wrote \ensuremath{-}; gold bands: habit layers. All n's
verified against the master ledger. Anomalies: * small n (\textless{}
15), suggestive only (Mistral IBP errors n+ = 11; Phi IBP errors n+ = 8;
Llama IBP errors n+ = 13); Qwen IBP has no blue bars in either class
(0/36 errors, 0/75 corrects wrote +).

\textbf{Table E2: Layer roles by domain and case class --- uniform
accounting.} Two rows per model: error cases and correct cases. Each
cell: mean MLP DLA toward the written sign at the named layer, 95\%
bootstrap CI (10,000 percentile resamples), case count; every cell pools
discovery + held-out. The wrote-+ columns carry the verdict: interval
entirely below 0 = stamper; entirely above 0 = amplifier; crossing 0 =
indeterminate. The wrote-\ensuremath{-} columns are the sanity check: positive for
every model, both domains, both classes, since both roles push \ensuremath{-} when
the model wrote \ensuremath{-}. (a) Qwen wrote + in zero IBP cases of either class
(0/36 errors, 0/75 corrects), so the test is not computable; behavioral
evidence on errors: final logit lead of the correct sign \ensuremath{-}11.9 {[}\ensuremath{-}14.2,
\ensuremath{-}9.4{]}, 33/33 negative. (b) Mistral: error wrote-+ CIs cross 0 (det) or
rest on n = 11 (IBP); its correct wrote-+ CIs are positive but small in
magnitude (\ensuremath{\le} +1), hence amplifier (weak). (c) * = small n (\textless{}
15), suggestive only.

\begin{longtable}[]{@{}>{\raggedright\arraybackslash}p{0.039\textwidth}>{\raggedright\arraybackslash}p{0.039\textwidth}>{\raggedright\arraybackslash}p{0.041\textwidth}>{\raggedright\arraybackslash}p{0.103\textwidth}>{\raggedright\arraybackslash}p{0.103\textwidth}>{\raggedright\arraybackslash}p{0.134\textwidth}>{\raggedright\arraybackslash}p{0.134\textwidth}>{\raggedright\arraybackslash}p{0.138\textwidth}>{\raggedright\arraybackslash}p{0.134\textwidth}@{}}
\toprule
\textbf{Model} & \textbf{Layer} & \textbf{Cases} & \textbf{Det role} &
\textbf{IBP role} & \textbf{Det wrote +} & \textbf{Det wrote \ensuremath{-}} & \textbf{IBP wrote +} & \textbf{IBP wrote \ensuremath{-}}\tabularnewline
\midrule
\endhead
Gemma & L58 & errors & stamper & amplifier & \textbf{-0.98 {[}-1.42, -0.51{]} n = 131} & \vtop{\hbox{\strut +2.18}\hbox{\strut {[}+1.77,
+2.60{]}}\hbox{\strut n = 87}} &
\vtop{\hbox{\strut +0.66}\hbox{\strut {[}+0.08, +1.21{]}}\hbox{\strut n
= 79}} & \vtop{\hbox{\strut +3.46}\hbox{\strut {[}+2.99,
+3.93{]}}\hbox{\strut n = 64}}\tabularnewline
& & corrects & amplifier & amplifier &
\vtop{\hbox{\strut +2.30}\hbox{\strut {[}+1.63, +2.96{]}}\hbox{\strut n
= 62}} & \vtop{\hbox{\strut +4.56}\hbox{\strut {[}+4.27,
+4.83{]}}\hbox{\strut n = 126}} &
\vtop{\hbox{\strut +2.88}\hbox{\strut {[}+2.34, +3.49{]}}\hbox{\strut n
= 25}} & \vtop{\hbox{\strut +4.90}\hbox{\strut {[}+4.73,
+5.07{]}}\hbox{\strut n = 175}}\tabularnewline
Qwen & L75 & errors & stamper/ judge & --- (a) &
\vtop{\hbox{\strut -0.72}\hbox{\strut {[}-0.88, -0.55{]}}\hbox{\strut n
= 104}} & \vtop{\hbox{\strut +1.74}\hbox{\strut {[}+1.60,
+1.89{]}}\hbox{\strut n = 102}} & --- (a) &
\vtop{\hbox{\strut +2.09}\hbox{\strut {[}+1.94, +2.23{]}}\hbox{\strut n
= 36}}\tabularnewline
& & corrects & stamper & --- (a) &
\vtop{\hbox{\strut -1.02}\hbox{\strut {[}-1.25, -0.76{]}}\hbox{\strut n
= 29}} & \vtop{\hbox{\strut +1.72}\hbox{\strut {[}+1.66,
+1.79{]}}\hbox{\strut n = 159}} & --- (a) &
\vtop{\hbox{\strut +2.09}\hbox{\strut {[}+2.07, +2.11{]}}\hbox{\strut n
= 75}}\tabularnewline
Phi & L39 & errors & amplifier & amplifier &
\vtop{\hbox{\strut +1.56}\hbox{\strut {[}+0.97, +2.13{]}}\hbox{\strut n
= 30}} & \vtop{\hbox{\strut +0.97}\hbox{\strut {[}+0.68,
+1.29{]}}\hbox{\strut n = 181}} &
\vtop{\hbox{\strut +0.62}\hbox{\strut {[}+0.30, +0.91{]}}\hbox{\strut n
= 8 *}} & \vtop{\hbox{\strut +2.67}\hbox{\strut {[}+2.38,
+2.95{]}}\hbox{\strut n = 109}}\tabularnewline
& & corrects & amplifier & amplifier &
\vtop{\hbox{\strut +2.12}\hbox{\strut {[}+1.32, +3.04{]}}\hbox{\strut n
= 23}} & \vtop{\hbox{\strut +0.96}\hbox{\strut {[}+0.66,
+1.27{]}}\hbox{\strut n = 176}} &
\vtop{\hbox{\strut +3.89}\hbox{\strut {[}+3.39, +4.38{]}}\hbox{\strut n
= 66}} & \vtop{\hbox{\strut +3.77}\hbox{\strut {[}+3.37,
+4.16{]}}\hbox{\strut n = 130}}\tabularnewline
Llama & L79 & errors & amplifier & amplifier &
\vtop{\hbox{\strut +2.99}\hbox{\strut {[}+2.45, +3.52{]}}\hbox{\strut n
= 77}} & \vtop{\hbox{\strut +1.13}\hbox{\strut {[}+0.62,
+1.67{]}}\hbox{\strut n = 131}} &
\vtop{\hbox{\strut +5.66}\hbox{\strut {[}+3.57, +7.60{]}}\hbox{\strut n
= 13 *}} & \vtop{\hbox{\strut +1.19}\hbox{\strut {[}+0.06,
+2.32{]}}\hbox{\strut n = 30}}\tabularnewline
& & corrects & amplifier & amplifier &
\vtop{\hbox{\strut +3.38}\hbox{\strut {[}+2.63, +4.08{]}}\hbox{\strut n
= 72}} & \vtop{\hbox{\strut +2.76}\hbox{\strut {[}+2.42,
+3.09{]}}\hbox{\strut n = 138}} &
\vtop{\hbox{\strut +8.93}\hbox{\strut {[}+8.25, +9.58{]}}\hbox{\strut n
= 46}} & \vtop{\hbox{\strut +4.45}\hbox{\strut {[}+4.19,
+4.70{]}}\hbox{\strut n = 24}}\tabularnewline
Mistral & L39 & errors & indet. (b) & indet. (b) &
\vtop{\hbox{\strut -0.14}\hbox{\strut {[}-0.36, +0.08{]}}\hbox{\strut n
= 38}} & \vtop{\hbox{\strut +0.18}\hbox{\strut {[}+0.07,
+0.29{]}}\hbox{\strut n = 199}} &
\vtop{\hbox{\strut -0.28}\hbox{\strut {[}-0.53, -0.04{]}}\hbox{\strut n
= 11 *}} & \vtop{\hbox{\strut +0.30}\hbox{\strut {[}+0.15,
+0.47{]}}\hbox{\strut n = 98}}\tabularnewline
& & corrects & amplifier (weak) (b) & amplifier (weak) (b) &
\vtop{\hbox{\strut +0.99}\hbox{\strut {[}+0.64, +1.39{]}}\hbox{\strut n
= 53}} & \vtop{\hbox{\strut +0.93}\hbox{\strut {[}+0.74,
+1.13{]}}\hbox{\strut n = 147}} &
\vtop{\hbox{\strut +0.32}\hbox{\strut {[}+0.17, +0.47{]}}\hbox{\strut n
= 85}} & \vtop{\hbox{\strut +1.44}\hbox{\strut {[}+1.23,
+1.66{]}}\hbox{\strut n = 114}}\tabularnewline
\bottomrule
\end{longtable}

\textbf{How to read this table.} Each cell answers one question: when
the model wrote this sign, which way did this layer push, and how hard?
The number is the layer's average push, in logits, toward the sign the
model wrote. Positive: the layer supported the written sign. Negative:
the layer pushed against it, toward minus. For scale: a logit difference
of \ensuremath{\Delta} multiplies the odds between the two sign tokens by e\^{}\ensuremath{\Delta} --- a
standard softmax identity (see, e.g., Bishop, 2006) --- so one logit is
a \ensuremath{\times}2.7 shift in odds and two logits \ensuremath{\times}7.4. The bracket is a 95\%
confidence interval: we only have n cases, so the true average could sit
anywhere in that range. The verdict takes one glance at the wrote-+
cells. Bracket fully below zero: the layer pushes minus even when the
model wrote plus --- a stamper. Bracket fully above zero: the layer
follows the model's choice --- an amplifier. Bracket crossing zero: no
call. Worked example --- the shaded cell: Gemma, determinant errors,
wrote +. The bracket {[}\ensuremath{-}1.42, \ensuremath{-}0.51{]} sits fully below zero: even when
Gemma wrote +, layer L58 reliably pushed toward \ensuremath{-}. That single cell is
the stamper verdict. The wrote-\ensuremath{-} cells cannot separate the two roles,
because both roles push minus there. They serve instead as a check on
the whole method: all twenty are positive, exactly as they must be. Why
we give all of this: every 'stamper' or 'amplifier' claim in this paper
is a claim about one of these cells. The table lets the reader verify
each one directly, with the same test applied everywhere.

Roles are properties of the layers, not of error cases. Qwen's
minus-stamp, Llama's amplifier and Phi's L39 amplifier all appear on
corrects too. Mistral's errors are indeterminate; its corrects lean
amplifier, weakly. One divergence: Gemma's determinant L58 --- stamper
on errors, amplifier on corrects. Its determinant corrects commit early
(median depth 0.21), before L58's window.

Across domains, the dependence is selective. Gemma L58 changes role; Phi
L39 and Llama L79 do not. Qwen's role test is not computable on IBP: no
wrote-+ cases in either class.

What separates errors from corrects is the sign the machinery is applied
to, not the machinery. And the timing never moves: in both domains, both
classes commit in the last few percent of depth. Localization claims
must name the task; the late-commitment claim need not.

\clearpage
\section*{Appendix F --- Origin of the minus habit: base model vs.
instruct model}
\addcontentsline{toc}{section}{Appendix F --- Origin of the minus habit: base model vs.
instruct model}

\emph{The habit the paper steers is not created by chat-training. This
appendix asks where it comes from --- and tests the answer directly on
one model.}

\textbf{The question.} The main text shows that each model carries fixed
late layers that push the minus sign --- a ``minus stamp.'' Where does
such a habit come from: the final chat-training stage, or earlier, in
pretraining?

\textbf{The hypothesis (general).} Written mathematics marks negative
results explicitly (``\ensuremath{-}67'') while positive results are written bare
(``67'' --- almost never ``+67''). Any model pretrained on such text
sees the minus sign written at answer positions far more often than the
plus sign, so a ``when in doubt, write minus'' prior is a plausible
thing for pretraining to install --- in any of the five models, not one
in particular. This is stated as a hypothesis: corpus statistics have
not been measured (\S{}7 sketches the corpus probe that would test it; left
to follow-up work).

\textbf{Scope: one model.} The direct test needs the base
(pretrained-only) and instruct versions of the same model, so it was
designed for and executed on Qwen-2.5-72B only. It has not been run on
the other four models in this paper; generalisation to them is left to
follow-up work.

\textbf{Method.} (1) Take Qwen2.5-72B (the base pretrained model, before
instruction tuning) and the instruct model studied in the paper. (2)
Build 49 short arithmetic completions where a signed result is written,
and locate the written sign token; a matched instruct-generated set of
48 completions provides the third condition. (3) Teacher-forced at that
position, measure each late MLP's DLA to the sign --- exactly the s02
DLA readout (Appendix B) --- separately for completions where the
written sign is minus and where it is plus. A fixed stamp shows as a
layer pushing minus in both groups.

\textbf{Results.} The stamp is already present in the base model: L78's
mean MLP DLA is +3.44 toward minus on minus-writers and \ensuremath{-}1.23 on
plus-writers (i.e., still pushing minus in both groups). The instruct
model on the same base texts shows the same pattern, larger: +4.82 /
\ensuremath{-}1.99. On its own sampled texts the instruct model gives +5.39 / \ensuremath{-}1.51
--- instruction tuning amplifies the L78 stamp by roughly 40--60\%
(derived from the DLA values reported in this paragraph; raw values in
the verification note below) but does not create it. L75 behaves
differently: in the base model it is not a fixed minus layer (+1.50 on
plus-writers, i.e., helping plus in that group); its stamp behaviour is
consistent with this emerging under instruction tuning.

\textbf{Reading.} Qwen's core minus habit (L78) predates chat-training
--- consistent with a pretraining origin such as the sign-marking
asymmetry in written mathematics --- and instruction tuning strengthens
it and recruits L75. This is one model, one probe set (49--48 arithmetic
completions), and no corpus measurement: supporting evidence for the
origin hypothesis, not proof.

\textbf{Data.} Three raw result files in the repository (path:
reference/\allowbreak qwen\_original/\allowbreak reference\_qwen/\allowbreak results/):
Qwen\_base/\allowbreak arith\_simple/\allowbreak v2\_mean\_ablation/\allowbreak expB\_dla.json (base model,
n=49); Qwen/\allowbreak arith\_simple\_base\_text/\allowbreak v2\_mean\_ablation/\allowbreak expB\_dla.json
(instruct model on same base texts, n=49);
Qwen/\allowbreak arith\_simple/\allowbreak v2\_mean\_ablation/\allowbreak expB\_dla.json (instruct model on
its own sampled texts, n=48). The code is the same DLA stage used
everywhere else (s02\_dla); nothing changes between conditions except
which model weights are loaded (base vs instruct) and which text is
replayed. Per-case DLA at L75/L78 and controls L30/L50 are logged for
every case; the means above are computed directly from those per-case
values. Related readouts: Appendix B (the s02 DLA stage) and Appendix H
(dose-response on det\_4x4 held-out).

All six numbers in the Results paragraph (3.44 / \ensuremath{-}1.23 / 4.82 / \ensuremath{-}1.99 /
5.39 / \ensuremath{-}1.51) have been verified against the raw JSON files listed
above; raw values are 3.4357 / \ensuremath{-}1.2331 / 4.8219 / \ensuremath{-}1.9900 / 5.3864 /
\ensuremath{-}1.5140. The +1.50 base-model L75 figure on plus-writers is verified at
1.5028.

\clearpage
\section*{Appendix G --- The Ablation Ladder: Attention Heads, Single MLP
Layers, and Joint MLP
Removal}

\subsection*{G.1 Contract}

This appendix answers one question three ways: is the wrong sign carried
by any single component of the circuit, or only by the identified
late-MLP set acting together? It climbs a three-rung ladder --- ablate
one attention head, then one MLP layer, then all identified MLP layers
jointly --- and stands behind two claims in Section 5: that \emph{no
single attention head accounts for the effect}, and that the localised
layers are \emph{causally necessary but not sufficient, with no single
layer alone accounting for the effect}.

By the end you can verify three things. (i) In no cell --- discovery or
held-out --- does a single attention head move the written sign's lead
by more than \ensuremath{-}0.46 logits, and single-head strict flips are essentially
zero. (ii) No single MLP layer exceeds \ensuremath{-}0.63 outside Llama --- whose
L77--L79 trio is the one place a single layer is individually strong,
reaching \ensuremath{-}3.46 on IBP. (iii) Removing the identified MLP set jointly
cuts the lead by \ensuremath{-}0.9 to \ensuremath{-}7.1 logits in three of five models, exceeding
40 matched random 5-layer sets, and by \ensuremath{-}1.3 to \ensuremath{-}8.6 in the cases whose
baseline lead is at least +2 logits.

The ordering \emph{head single layer \textless{} joint MLP} holds
everywhere, with exactly two documented exceptions carried openly: Gemma
IBP heads (G.6, closed by a joint-head follow-up) and Phi, where every
rung is flat and the ordering claim is vacuous (G.7).

\subsection*{G.2 Shared protocol}

One measurement backs every number in this appendix. Each case is
replayed teacher-forced to the answer-sign position; the metric is the
mean change in the written sign's final logit lead (ablated minus
baseline), so negative means the written sign weakened.

The three rungs differ only in what is removed. Rung 1 zeroes one
attention head's output at the sign position (a mean-ablation variant,
expC\ensuremath{'}, covers Qwen and Llama). Rung 2 removes the component of one MLP
layer's output along the frozen sign direction. Rung 3 applies that same
removal at all identified layers simultaneously, and compares against 40
matched random 5-layer sets --- the null that a same-sized ablation
anywhere would do the same.

Candidates were identified on the discovery split and frozen; the
held-out split is the confirmation, so held-out errors are the headline
cells. Determinant and IBP are never mixed: every table stacks the
determinant block on top and the IBP block below.

Two further conventions. Where a table reports a \emph{baseline \ensuremath{\ge} +2}
column, it restricts to cases whose baseline lead is at least +2 logits
--- a case leading by +0.3 cannot mechanically show a \ensuremath{-}3 drop, so this
subset shows the effect where there is room to see it. Corrects (the
same removal applied to correctly-signed answers) are reported alongside
errors at rung 3; rungs 1--2 were run on errors only.

\textbf{Worked example (rung 3): Llama IBP.} Take Llama-3.3-70B on IBP
errors (n = 21 held-out cases). The identified five layers are L64, L66,
L77, L78, L79. The baseline wrong-sign lead is +14.61 logits. After
jointly ablating the sign-direction component at all five layers, the
lead falls to +7.52 --- a drop of \ensuremath{-}7.09. Forty random 5-layer sets
average \ensuremath{-}0.29 and stay within {[}\ensuremath{-}0.96, +0.69{]}: the targeted effect is
more than 7\ensuremath{\times} their worst case. Yet no single layer gets there alone ---
the best, L79, gives \ensuremath{-}3.46 (Table G3) --- and the best attention head
was never even worth a held-out run (Table G2). The habit is spread
redundantly across the identified MLP set, and only removing the set
collapses the lead.

\subsection*{G.3 Rung 1 --- single attention
heads}

Before any head was touched, stage 2 already measured what each head
says directly to the output: its direct logit attribution (DLA) --- the
head's contribution, read through the unembedding, to the written sign's
lead. Table G1 shows the strongest head in each discovery cell next to
the strongest MLP layer in the same cell, in the same units. In four of
ten cells (Llama and Qwen, both tasks) the best head sits 2--7\ensuremath{\times} below
the best MLP layer before any intervention is made; in Gemma's two cells
the best head's direct path is essentially zero, in Mistral the two are
comparable, and in Phi the ordering reverses --- the caveat the next
paragraph takes up.

Table G1: Stage-2 direct logit attribution, discovery errors ---
the strongest attention head vs. the strongest MLP layer in the same
cell (mean logits contributed to the written sign's lead; positive
pushes the wrong sign). Determinant block on top, IBP below.

\begin{longtable}[]{@{}llll@{}}
\toprule
Model & n & Top head (DLA) & Top MLP layer (DLA)\tabularnewline
\midrule
\endhead
Determinant & & &\tabularnewline
Gemma & 92 & L60H4 \ensuremath{-}0.00 & L55 +2.89\tabularnewline
Llama & 16 & L78H32 \ensuremath{-}0.53 & L79 +3.06\tabularnewline
Mistral & 95 & L38H7 +0.43 & L37 +0.53\tabularnewline
Phi & 61 & L36H33 +2.89 & L37 +1.72\tabularnewline
Qwen & 59 & L77H44 +0.36 & L77 +1.12\tabularnewline
IBP & & &\tabularnewline
Gemma & 70 & L60H4 \ensuremath{-}0.01 & L55 +2.95\tabularnewline
Llama & 22 & L78H32 +0.66 & L78 +2.01\tabularnewline
Mistral & 39 & L38H7 +0.87 & L37 +1.01\tabularnewline
Phi & 40 & L36H33 +3.99 & L33 +3.35\tabularnewline
Qwen & 10 & L77H40 +1.01 & L78 +2.44\tabularnewline
\bottomrule
\end{longtable}

\noindent Per-head DLAs sum exactly to each layer's attention DLA (verified per
case). Stage-2 n's precede the late-circuit filter, so they can differ
slightly from the ablation n's below.

Table G1 also shows why DLA alone cannot settle rung 1: it misleads in
both directions. In Phi, L36H33 has the largest DLA in the entire grid
(+2.89 det, +3.99 IBP --- above every MLP layer), yet ablating it does
almost nothing (\ensuremath{-}0.46, and the effect vanishes on re-run) --- DLA
overstates. In Gemma, the late heads have \emph{zero} direct-path DLA,
yet ablating them on IBP moves the lead (G.6) --- DLA understates,
because those heads act only indirectly, through the downstream MLPs.
Gemma's pusher/corrector labels therefore come from a degenerate ranking
of near-zero values; for Gemma it is the causal test, not the ranking,
that carries the weight. Both failure modes are exactly why every head
was tested by ablation.

\textbf{Worked example: Qwen H44.} Qwen-2.5-72B's H44 at layer 77 is the
top-ranked pusher --- the head the DLA ranking says should contribute
most to writing the wrong sign (+0.36, Table G1). Baseline wrong-sign
lead: +5.19 logits (mean, n = 147 held-out determinant errors). Mean
change after ablating H44's output at the answer-sign position: \ensuremath{-}0.17
logits, with zero of 147 cases flipping. The joint MLP ablation on the
same cases (Table G4) gives \ensuremath{-}2.52. The best single attention head
delivers under 7\% of the joint MLP effect and flips nothing.

To read one row of Table G2: \emph{n} is the error-pool size;
\emph{heads} is the number of candidate heads in the grid (pushers,
correctors, and a mid-rank control per layer, from the stage-2 ranking);
\emph{strongest} and \emph{most positive} are the extreme mean effects
across the grid. Method "zero" is s05b zero-ablation; "mean" is the
expC\ensuremath{'} mean-ablation battery. Rows marked \emph{disc} are discovery-split
grids --- the only split where that grid exists.

Table G2: Per-head ablation on errors --- the extreme heads of
every grid. Determinant block on top, IBP below.

\begin{longtable}[]{@{}lllllll@{}}
\toprule
Model (split) & Method & n & Heads & Strongest head & Most positive &
Strict flips\tabularnewline
\midrule
\endhead
Determinant & & & & & &\tabularnewline
Gemma (held-out) & zero & 126 & 5 & L60H5 \ensuremath{-}0.16 & L60H4 +0.13 &
0\tabularnewline
Llama (held-out) & mean & 192 & 12 & H47 \ensuremath{-}0.03 & H41 +0.09 &
0\tabularnewline
Mistral (held-out) & zero & 142 & 8 & L38H29 \ensuremath{-}0.15 & L38H30 +0.06 &
0\tabularnewline
Phi (disc) & zero & 61 & 8 & L36H33 \ensuremath{-}0.46 & L35H37 +0.03 &
0\tabularnewline
Qwen (held-out) & mean & 147 & 12 & H44 \ensuremath{-}0.17 & L75H62 +0.04 &
0\tabularnewline
IBP & & & & & &\tabularnewline
Gemma (held-out) & zero & 73 & 5 & L60H5 \ensuremath{-}0.45 & L60H4 +0.51 &
0\tabularnewline
Llama & --- & --- & --- & not run & --- & ---\tabularnewline
Mistral (held-out) & zero & 70 & 8 & L38H7 \ensuremath{-}0.05 & L38H30 +0.07 &
0\tabularnewline
Phi (disc) & zero & 40 & 8 & L36H33 \ensuremath{-}0.23 & L36H35 +0.08 &
0\tabularnewline
Qwen (disc) & mean & 10 & 12 & H44 \ensuremath{-}0.12 & --- & 0\tabularnewline
\bottomrule
\end{longtable}

\noindent Unlabelled heads are at each model's top habit layer (Llama/Qwen: L77
unless prefixed L75). Phi has no held-out head grid; its discovery grid
was re-run once (v3) and the L36H33 effect vanished (+0.01 det, \ensuremath{-}0.04
IBP). Gemma IBP is the one grid where a head effect approaches
single-layer size --- treated in G.6.

Two stress tests push rung 1 past single heads. For Qwen, zeroing
\emph{all} of layer 77's attention at once --- every head simultaneously
--- fixes zero cases and produces zero strict flips on both splits (n =
44 discovery, 147 held-out). And the mean-ablation battery (expC\ensuremath{'})
agrees with the zero-ablation grids wherever both exist: every head
effect within \ensuremath{\pm}0.17 for Qwen and within \ensuremath{\pm}0.09 for Llama.

No single attention head --- and for Qwen, not even the whole attention
layer --- carries the wrong sign.

\subsection*{G.4 Rung 2 --- single MLP
layers}

To read one row: \emph{top identified layer} is the strongest single
layer from the frozen identified set when its sign-direction component
is removed alone; \emph{runner-up} is the next strongest;
\emph{controls} gives the range across control and check layers with no
hypothesised role. All cells are held-out errors --- held-out
single-layer coverage is complete for all five models.

Table G3: Single-layer sign-direction ablation, held-out errors.
Determinant block on top, IBP below.

\begin{longtable}[]{@{}lllll@{}}
\toprule
Model & n & Top identified layer & Runner-up & Controls
(range)\tabularnewline
\midrule
\endhead
Determinant & & & &\tabularnewline
Gemma & 126 & L55 \ensuremath{-}0.52 & L59 \ensuremath{-}0.38 & +0.04 to +0.15\tabularnewline
Llama & 192 & L79 \ensuremath{-}1.79 & L78 \ensuremath{-}1.39 & +0.05 to +0.30\tabularnewline
Mistral & 142 & L37 \ensuremath{-}0.15 & L38 \ensuremath{-}0.08 & \ensuremath{-}0.20 to +0.18\tabularnewline
Phi & 150 & L37 \ensuremath{-}0.21 & L39 \ensuremath{-}0.17 & \ensuremath{-}0.10 to +0.12\tabularnewline
Qwen & 147 & L78 \ensuremath{-}0.46 & L75 \ensuremath{-}0.42 & \ensuremath{-}0.03 to +0.02\tabularnewline
IBP & & & &\tabularnewline
Gemma & 73 & L58 \ensuremath{-}0.63 & L55 \ensuremath{-}0.49 & \ensuremath{-}0.15 to \ensuremath{-}0.14\tabularnewline
Llama & 21 & L79 \ensuremath{-}3.46 & L78 \ensuremath{-}2.62 & +0.22 to +0.45\tabularnewline
Mistral & 70 & L37 \ensuremath{-}0.18 & L34 \ensuremath{-}0.12 & \ensuremath{-}0.10 to +0.21\tabularnewline
Phi & 77 & L37 \ensuremath{-}0.24 & L39 +0.04 & \ensuremath{-}0.11 to \ensuremath{-}0.04\tabularnewline
Qwen & 26 & L77 \ensuremath{-}0.46 & L75 \ensuremath{-}0.14 & \ensuremath{-}0.03 to +0.02\tabularnewline
\bottomrule
\end{longtable}

\noindent Mistral's control range includes a check layer (L32, \ensuremath{-}0.20) that
slightly exceeds its best identified layer --- all Mistral single-layer
effects sit near zero, which is exactly the redundant-spread picture:
its joint effect on baseline \ensuremath{\ge} +2 cases is \ensuremath{-}2.73 (Table G4). Llama is
the one model with individually strong layers (L79/L78/L77); even there
the joint effect roughly doubles the best single layer.

Outside Llama, no single layer in any model reaches \ensuremath{-}0.7 on held-out
errors. The sign is written redundantly, and the redundancy is the point
of rung 3.

\subsection*{G.5 Rung 3 --- the identified MLP set
jointly}

To read one row: \emph{\ensuremath{\Delta} (n)} is the mean lead change over all cases
when the full identified set is ablated at once; \emph{\ensuremath{\Delta}, baseline \ensuremath{\ge} +2
(n)} restricts to cases with baseline lead of at least +2 logits;
\emph{beats null} counts cases whose drop exceeds the 5th percentile of
their own 40 random-set draws; \emph{null mean {[}range{]}} summarises
the 40 matched random 5-layer sets; \emph{flips} counts held-out cases
whose written-sign lead crosses zero (baseline \textgreater{} 0,
post-ablation \textless{} 0), with the mean flip count over the same
cases under the 40 random sets in parentheses. Both splits are shown so
nothing is hidden --- discovery is where the layers were found, held-out
is the frozen confirmation.

\textbf{Table G4: Joint ablation of the identified MLP set, errors.
Determinant block on top, IBP below.}

\begin{longtable}[]{@{}>{\raggedright\arraybackslash}p{0.076\textwidth}>{\raggedright\arraybackslash}p{0.068\textwidth}>{\raggedright\arraybackslash}p{0.137\textwidth}>{\raggedright\arraybackslash}p{0.076\textwidth}>{\raggedright\arraybackslash}p{0.158\textwidth}>{\raggedright\arraybackslash}p{0.103\textwidth}>{\raggedright\arraybackslash}p{0.137\textwidth}>{\raggedright\arraybackslash}p{0.123\textwidth}@{}}
\toprule
Model & Disc \ensuremath{\Delta} (n) & Disc null mean {[}range{]} & Held \ensuremath{\Delta} (n) & Held \ensuremath{\Delta},
baseline \ensuremath{\ge} +2 (n) & Held beats null & Held null mean {[}range{]} & Held
flips (rand.)\tabularnewline
\midrule
\endhead
Determinant & & & & & & &\tabularnewline
Gemma & \ensuremath{-}3.62 (92) & \ensuremath{-}0.24 {[}\ensuremath{-}1.45, +0.76{]} & \ensuremath{-}1.24 (126) & \ensuremath{-}3.15 (49)
& 33/126 & +0.49 {[}\ensuremath{-}0.36, +1.31{]} & 22 (5.5)\tabularnewline
Llama & \ensuremath{-}4.87 (16) & \ensuremath{-}0.62 {[}\ensuremath{-}2.36, +0.43{]} & \ensuremath{-}4.25 (192) & \ensuremath{-}5.61
(138) & 91/192 & \ensuremath{-}0.67 {[}\ensuremath{-}1.42, +0.16{]} & 41 (11.5)\tabularnewline
Mistral & \ensuremath{-}0.69 (93) & \ensuremath{-}0.36 {[}\ensuremath{-}0.81, +0.11{]} & \ensuremath{-}0.32 (142) & \ensuremath{-}2.73
(66) & 44/142 & \ensuremath{-}0.23 {[}\ensuremath{-}0.57, +0.16{]} & 11 (4.0)\tabularnewline
Phi & +0.43 (61) & \ensuremath{-}0.90 {[}\ensuremath{-}2.12, \ensuremath{-}0.12{]} & \ensuremath{-}0.19 (150) & \ensuremath{-}0.27 (143)
& 7/150 & \ensuremath{-}2.70 {[}\ensuremath{-}4.47, \ensuremath{-}1.27{]} & 2 (3.4)\tabularnewline
Qwen & \ensuremath{-}2.08 (59) & \ensuremath{-}0.46 {[}\ensuremath{-}1.55, +0.13{]} & \ensuremath{-}2.52 (147) & \ensuremath{-}3.53 (94)
& 82/147 & \ensuremath{-}0.42 {[}\ensuremath{-}1.47, +0.17{]} & 24 (2.1)\tabularnewline
IBP & & & & & & &\tabularnewline
Gemma & \ensuremath{-}2.86 (70) & \ensuremath{-}0.26 {[}\ensuremath{-}0.96, +0.29{]} & \ensuremath{-}0.91 (73) & \ensuremath{-}1.73 (45)
& 25/73 & +0.03 {[}\ensuremath{-}0.28, +0.38{]} & 5 (0.6)\tabularnewline
Llama & \ensuremath{-}4.01 (22) & \ensuremath{-}0.07 {[}\ensuremath{-}1.05, +0.66{]} & \ensuremath{-}7.09 (21) & \ensuremath{-}8.62 (18)
& 15/21 & \ensuremath{-}0.29 {[}\ensuremath{-}0.96, +0.69{]} & 1 (0.1)\tabularnewline
Mistral & +0.29 (39) & +0.29 {[}+0.04, +0.48{]} & \ensuremath{-}0.34 (70) & \ensuremath{-}1.41
(31) & 27/70 & \ensuremath{-}0.07 {[}\ensuremath{-}0.24, +0.21{]} & 2 (1.3)\tabularnewline
Phi & +1.61 (40) & +1.00 {[}+0.05, +1.68{]} & \ensuremath{-}0.04 (77) & \ensuremath{-}0.25 (58) &
0/77 & \ensuremath{-}0.10 {[}\ensuremath{-}1.04, +1.01{]} & 4 (5.0)\tabularnewline
Qwen & \ensuremath{-}0.76 (10) & \ensuremath{-}0.12 {[}\ensuremath{-}0.64, +0.46{]} & \ensuremath{-}0.98 (26) & \ensuremath{-}1.29 (23) &
2/26 & \ensuremath{-}0.11 {[}\ensuremath{-}0.70, +0.66{]} & 0 (0.0)\tabularnewline
\bottomrule
\end{longtable}

\noindent Qwen IBP held-out (\ensuremath{-}0.98, n = 26) is descriptive only: wide CI, and its
discovery arm (n = 10) did not beat the null. Phi never separates from
random at any rung --- see G.7, including the one place the random sets
outperform the targeted set.

The flip column keeps the mean from hiding the case-level story. Mistral
is the clearest example: its held-out determinant mean is only \ensuremath{-}0.32,
yet the joint removal flips 11 cases outright against 4.0 expected under
the random sets --- the flips sit in cases with modest baselines (+1.9
to +2.9), where a small targeted push is enough to cross zero. Every
responsive model flips above its own random rate (Qwen det 24 vs 2.1,
Gemma det 22 vs 5.5); Phi is the exception in the other direction,
flipping at or below chance in both tasks (G.7).

The same removal applied to \emph{correct} answers tests necessity for
the correct sign --- the direction carries "this answer is minus," so
removing it should also damage answers whose minus is right.

Table G5: Joint ablation of the identified MLP set, corrects.
Determinant block on top, IBP below. A dash marks a cell that was never
run (G.8 gives the reason for each).

\begin{longtable}[]{@{}>{\raggedright\arraybackslash}p{0.076\textwidth}>{\raggedright\arraybackslash}p{0.068\textwidth}>{\raggedright\arraybackslash}p{0.137\textwidth}>{\raggedright\arraybackslash}p{0.076\textwidth}>{\raggedright\arraybackslash}p{0.158\textwidth}>{\raggedright\arraybackslash}p{0.103\textwidth}>{\raggedright\arraybackslash}p{0.137\textwidth}>{\raggedright\arraybackslash}p{0.123\textwidth}@{}}
\toprule
Model & Disc \ensuremath{\Delta} (n) & Disc null mean {[}range{]} & Held \ensuremath{\Delta} (n) & Held \ensuremath{\Delta},
baseline \ensuremath{\ge} +2 (n) & Held beats null & Held null mean {[}range{]} & Held
flips (rand.)\tabularnewline
\midrule
\endhead
Determinant & & & & & & &\tabularnewline
Gemma & \ensuremath{-}4.82 (50) & \ensuremath{-}0.11 {[}\ensuremath{-}0.60, +0.42{]} & \ensuremath{-}2.63 (138) & \ensuremath{-}2.64
(120) & 57/138 & \ensuremath{-}1.03 {[}\ensuremath{-}1.78, +0.29{]} & 15 (7.0)\tabularnewline
Llama & \ensuremath{-}6.91 (80) & \ensuremath{-}0.72 {[}\ensuremath{-}2.29, +0.10{]} & --- & --- & --- & --- &
---\tabularnewline
Mistral & \ensuremath{-}3.40 (78) & \ensuremath{-}1.08 {[}\ensuremath{-}1.61, \ensuremath{-}0.43{]} & \ensuremath{-}3.02 (122) & \ensuremath{-}3.97
(104) & 63/122 & \ensuremath{-}1.21 {[}\ensuremath{-}1.65, \ensuremath{-}0.48{]} & 16 (2.7)\tabularnewline
Phi & \ensuremath{-}0.74 (50) & \ensuremath{-}1.61 {[}\ensuremath{-}3.44, \ensuremath{-}0.32{]} & +0.99 (149) & +0.93 (139)
& 3/149 & \ensuremath{-}0.72 {[}\ensuremath{-}2.02, \ensuremath{-}0.08{]} & 1 (2.6)\tabularnewline
Qwen & --- & --- & --- & --- & --- & --- & ---\tabularnewline
IBP & & & & & & &\tabularnewline
Gemma & \ensuremath{-}5.59 (50) & \ensuremath{-}0.37 {[}\ensuremath{-}0.50, \ensuremath{-}0.17{]} & +0.21 (150) & +0.19
(147) & 29/150 & \ensuremath{-}0.03 {[}\ensuremath{-}0.34, +0.21{]} & 0 (0.0)\tabularnewline
Llama & +0.48 (23) & +0.20 {[}\ensuremath{-}0.02, +0.45{]} & +0.23 (47) & +0.23 (47)
& 7/47 & \ensuremath{-}0.11 {[}\ensuremath{-}0.67, +0.51{]} & 1 (0.0)\tabularnewline
Mistral & \ensuremath{-}4.62 (43) & \ensuremath{-}0.78 {[}\ensuremath{-}1.20, \ensuremath{-}0.30{]} & \ensuremath{-}4.32 (156) & \ensuremath{-}4.49
(150) & 137/156 & \ensuremath{-}0.73 {[}\ensuremath{-}1.10, \ensuremath{-}0.44{]} & 2 (0.6)\tabularnewline
Phi & \ensuremath{-}4.88 (50) & \ensuremath{-}1.17 {[}\ensuremath{-}2.86, +0.33{]} & \ensuremath{-}2.60 (146) & \ensuremath{-}2.60 (146)
& 5/146 & \ensuremath{-}3.00 {[}\ensuremath{-}4.38, \ensuremath{-}1.77{]} & 0 (0.0)\tabularnewline
Qwen & --- & --- & --- & --- & --- & --- & ---\tabularnewline
\bottomrule
\end{longtable}

\noindent Where the correct answer's sign is minus-heavy (Gemma and Mistral det,
Mistral and Phi IBP discovery), the joint ablation damages corrects at
the same scale as errors --- the direction is load-bearing for correct
minus signs too, anticipating the C2-on-correct steering result quoted
in Appendix H. Llama IBP corrects (mostly plus-signed) are untouched
(+0.23), the expected sign-specificity. Here a flip is damage --- a
correct written sign pushed to the wrong side --- and the same pattern
repeats: Mistral flips 16 correct determinant answers against 2.7
expected under random sets (Gemma 15 vs 7.0), while Phi again sits at
chance (1 vs 2.6).

\begin{center}\includegraphics[width=\linewidth]{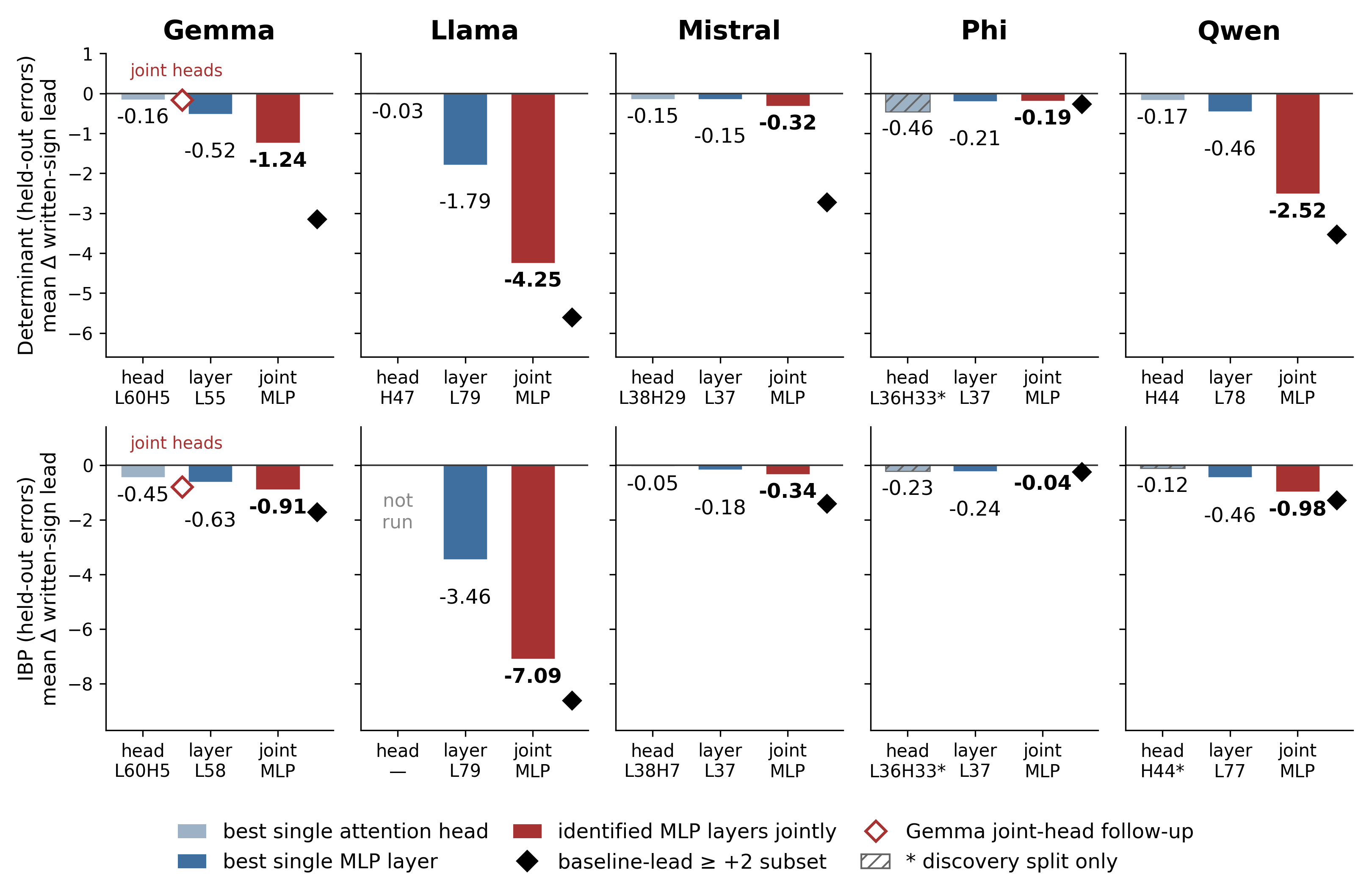}\end{center}

\emph{Figure G1: The ablation ladder on held-out errors --- determinant
on top, IBP below, one panel per model. Three bars per panel: the
strongest single attention head (grey; hatched where only a discovery
grid exists), the strongest single identified MLP layer (blue), and the
full identified set ablated jointly (red). Black diamonds mark the joint
effect in the baseline-lead \ensuremath{\ge} +2 subset; the open red diamond is Gemma's
joint-head follow-up (G.6). In every responsive model the bars step down
the ladder; Phi's stay flat at every rung (G.7).}

\subsection*{G.6 Exception 1 --- Gemma IBP heads, and the follow-up that
closed
it}

Gemma IBP is the only cell in Table G2 where single heads reach
single-layer size: on the discovery grid (Table G2 reports the held-out
values, \ensuremath{-}0.45 and +0.51), pusher L60H5 gives \ensuremath{-}1.01 and corrector L60H4
gives +1.36 --- comparable to Gemma's best single layers (Table G3).
Taken alone this would leave open that Gemma's IBP sign rides on
attention.

So the joint-head arms were run exactly here (and, as a control, on
Gemma det and Mistral det, where heads were already flat). Arms: H1
zeroes all pushers jointly, H2 pushers plus the corrector, H3 correctors
only, H4 mid-rank control heads.

Table G6: Joint-head ablation --- every grid that exists. Mean \ensuremath{\Delta}
lead (strict flips).

\begin{longtable}[]{@{}llllll@{}}
\toprule
Model / task (split) & n & H1 pushers & H2 +corrector & H3 correctors &
H4 control\tabularnewline
\midrule
\endhead
Gemma det (disc) & 92 & \ensuremath{-}0.64 (1) & \ensuremath{-}0.24 (2) & +0.75 (1) & \ensuremath{-}0.00
(0)\tabularnewline
Gemma det (held-out) & 126 & \ensuremath{-}0.16 (3) & \ensuremath{-}0.03 (2) & +0.20 (2) & +0.01
(0)\tabularnewline
Gemma IBP (disc) & 70 & \ensuremath{-}1.64 (3) & \ensuremath{-}0.54 (3) & +1.85 (1) & \ensuremath{-}0.01
(0)\tabularnewline
Gemma IBP (held-out) & 73 & \ensuremath{-}0.80 (5) & \ensuremath{-}0.41 (4) & +0.83 (2) & \ensuremath{-}0.02
(0)\tabularnewline
Mistral det (disc) & 93 & \ensuremath{-}0.29 (1) & \ensuremath{-}0.22 (0) & +0.11 (0) & +0.01
(1)\tabularnewline
\bottomrule
\end{longtable}

The follow-up closes the exception rather than hiding it. Even with
\emph{every} pusher head removed at once, Gemma IBP gives \ensuremath{-}1.64 on
discovery --- real, but 57\% of Gemma's own joint-MLP effect on the same
split (\ensuremath{-}2.86) --- and replicates at \ensuremath{-}0.80 held-out with 5 of 73 flips.
On held-out both effects attenuate together (joint-MLP \ensuremath{-}0.91 over all
cases), so the cleaner held-out comparison is the subset with room to
move: joint-MLP \ensuremath{-}1.73 on baseline \ensuremath{\ge} +2 cases versus joint-heads \ensuremath{-}0.80.
The H4 control heads sit at zero throughout, and no configuration of
heads ever exceeds the MLP set.

Recall from Table G1 that these same heads have zero direct-path DLA.
Gemma's L60 attention heads relay some minus-ness on IBP --- through the
downstream MLPs --- but they do not carry it. The ladder's ordering
survives: all heads jointly \ensuremath{\le} all MLP layers jointly, on every split.

\subsection*{G.7 Exception 2 --- Phi, where the ordering claim is
vacuous}

Phi-4 is flat at every rung: best single head \ensuremath{-}0.46 (discovery, and gone
on re-run), best single layer \ensuremath{-}0.21, joint set \ensuremath{-}0.19 held-out with only
7 of 150 cases beating their random null --- and its strict flips run at
or below the random-set rate (det 2 vs 3.4 expected, IBP 4 vs 5.0; Table
H4).

One caveat must be stated at full strength: on Phi det held-out, the 40
\emph{random} 5-layer sets average \ensuremath{-}2.70 {[}\ensuremath{-}4.47, \ensuremath{-}1.27{]} --- the
targeted set does \emph{not} beat random in Phi; the direction is the
same on the discovery split (targeted +0.43, null \ensuremath{-}0.90). For Phi we
therefore claim no single-component ordering at all; the claim is
vacuous, not merely weak.

Because Phi resisted, it received an escalation family no other model
needed. A cumulative staircase ablated the top-k DLA-ranked layers on
discovery errors (n = 61 det, 40 IBP): determinant creeps from +0.01 at
k = 1 to only \ensuremath{-}1.34 at k = 15, while IBP moves the \emph{wrong way},
+2.23 at k = 15. Window arms (all MLPs in L24--39: \ensuremath{-}1.28 det but +2.02
IBP; L30--39: \ensuremath{-}0.67 / +1.79) do no better than a random window control
(\ensuremath{-}0.44 / +0.48).

What finally moves Phi is the combined removal quoted in Section 5 of
the main text --- MLP sign-direction subtraction plus
amplifier-attention zeroing (s08 arm C4): \ensuremath{-}6.70 det and \ensuremath{-}7.11 IBP
held-out at \ensuremath{\alpha} = 1, though with zero strict flips, because Phi's baseline
margins are enormous (+19.5 and +22.0 logits on average) and absorb even
large pushes --- the oversized-margin story told in Section 6. Phi is
reported throughout as the null it is.

The combined removal is not Phi-specific. Table G7 reports the same arm
(s08 arm C4: sign-direction subtraction at the habit-layer MLPs plus
amplifier-attention zeroing) at \ensuremath{\alpha} = 1 on held-out errors for all five
models and both tasks --- the grid quoted in Section 5 of the main text.
Entries are the mean change in the wrong sign's final lead, in logits; n
is the held-out error count. Higher doses of the subtraction component
alone (arm C2, without the attention zeroing) are the steering
experiment of Appendix H.

Table G7: Combined removal (s08 arm C4) at \ensuremath{\alpha} = 1, held-out
errors, all five models. Mean \ensuremath{\Delta} in the wrong sign's final lead (logits);
n in parentheses. Source metric: mean\_delta\_ld{[}C4\textbar a1.0{]},
master ledger.

\begin{longtable}[]{@{}lll@{}}
\toprule
Model & det \ensuremath{\Delta} lead (n) & IBP \ensuremath{\Delta} lead (n)\tabularnewline
\midrule
\endhead
Gemma-3-27B & \ensuremath{-}0.12 (126) & \ensuremath{-}0.52 (72)\tabularnewline
Mistral-Small-24B & \ensuremath{-}0.19 (132) & \ensuremath{-}0.96 (67)\tabularnewline
Phi-4-14B & \ensuremath{-}6.70 (146) & \ensuremath{-}7.11 (76)\tabularnewline
Llama-3.3-70B & \ensuremath{-}3.09 (189) & \ensuremath{-}8.16 (20)\tabularnewline
Qwen-2.5-72B & \ensuremath{-}2.84 (99) & \ensuremath{-}4.94 (25)\tabularnewline
\bottomrule
\end{longtable}

\subsection*{G.8 What was not run, and
why}

Every absent cell in this appendix falls into one of three buckets,
stated so that no gap is silent.

\emph{Closed by logic.} Joint-head ablation for Phi, Llama, and Qwen was
not run because their single-head grids were flat everywhere (Table G2)
--- with no individual head effect to aggregate, the joint-head question
answers itself; it was run only where it was live (Gemma, plus Mistral
det as a control, G.6). Per-head grids on corrects were not run anywhere
for the same reason: heads do nothing on errors, where the effect lives.
Llama's IBP head grid was never built --- its det heads were at most
\ensuremath{-}0.03, and its IBP pool (n = 21) was reserved for the layer-level arms.

\emph{Halted on a null.} Mistral IBP joint ablation returned +0.29 on
discovery and Phi IBP +1.61 --- both wrong-signed nulls. The pre-planned
held-out replications were still run and are reported in Table G4 (\ensuremath{-}0.34
and \ensuremath{-}0.04, both \textasciitilde null); what was halted is everything
downstream --- no escalation family was pursued for Mistral, and Phi's
escalation (G.7) was motivated by its determinant cell, not IBP. The
nulls stand in the tables rather than being rescued.

\emph{Coverage gap.} Qwen joint-MLP-on-corrects was never run in any
split or domain, and Llama det corrects has a discovery arm only (\ensuremath{-}6.91,
n = 80, plus a held-out single-layer probe: L78 \ensuremath{-}2.45, n = 130). Both
gaps are patched by a stronger test of the same claim: the C2-on-correct
steering battery (Appendix H) applies the same direction-removal
generatively. Quoting its \ensuremath{\alpha} = 1 mean \ensuremath{\Delta} with breaks counted at \ensuremath{\alpha} = 3:
Qwen det discovery \ensuremath{-}0.83 (n = 47), held-out \ensuremath{-}1.04 (n = 141, 49 broken),
Qwen IBP held-out \ensuremath{-}2.59 (n = 50), and Llama det held-out \ensuremath{-}5.65 (n = 130,
74 broken). The single true hole is Qwen IBP discovery C2-on-correct,
which has no substitute and is simply absent.

\subsection*{G.9 Provenance}

Every number in this appendix was recomputed for this draft from the
per-case JSONs of the pipeline stages (s02 DLA, s05 joint and
single-layer, s05b per-head zero, s05c joint-head, s08 combined arms)
and the expC\ensuremath{'} mean-ablation battery, then cross-checked cell by cell
against the master results ledger
(\texttt{paper/appendix/ledger\_results.csv}); the figure and all tables
are generated by one script in the repository, with nothing hand-typed.

\clearpage
\section*{Appendix H --- The Sign Direction: Construction and Causal
Tests}
\addcontentsline{toc}{section}{Appendix H --- The Sign Direction: Construction and Causal
Tests}

\emph{Results of two pipeline stages (Appendix B):
s06\_direction\_subtraction --- removing the direction at one layer at a
time (H.3) --- and s08\_c2\_steering --- the C2 condition, both habit
layers at once, pooled discovery + held-out, reported per task (H.4).
All numbers recomputed from the raw per-case stage outputs and
cross-checked against the master results ledger
(paper/appendix/ledger\_results.csv); see H.5 for the pooling formula,
metric names, and source files.}

\textbf{H.1 What this appendix shows --- and from which stages}

Every intervention in this appendix acts on one object: the sign
direction --- a single vector in the residual stream along which ``\ensuremath{-}''
and ``+'' differ. H.2 defines that vector exactly. The rest of the
appendix reports its two causal tests, in pipeline order (the s00--s09
pipeline, Appendix B):

\begin{itemize}
\item
  \textbf{Stage s06\_direction\_subtraction (H.3).} Remove the direction
  at one habit layer at a time and measure the dose--response, with
  control layers and a random direction as nulls --- pooled over
  discovery + held-out (Phi: discovery only; H.3.1). This is the first
  causal test: writer layers move the written sign, controls stay flat.
\item
  \textbf{Stage s08\_c2\_steering (H.4).} The C2 condition of \S{}5:
  subtract the direction at both habit layers at once and ask what it
  does to four cohorts --- minus-written errors, plus-written errors,
  correct-minus and correct-plus --- pooled over discovery + held-out,
  per task. This is the specificity battery.
\end{itemize}

The construction itself (H.2) uses only model weights and discovery data
and was frozen before either stage ran. Every number's source file, and
the ledger cross-check, is in H.5.

\textbf{H.2 The sign direction: construction}

\textbf{How it is computed, step by step.} (1) Take the unembedding
matrix W\ensuremath{U} --- the map from the residual stream to token scores. (2) Take
the row for the space-prefixed token `` \ensuremath{-}'' and subtract the row for ``
+'' (the forms the models actually write mid-solution):

\emph{d = W\ensuremath{U}(`` \ensuremath{-}'') \ensuremath{-} W\ensuremath{U}(`` +'')}

Moving the hidden state along d raises minus scores exactly as much as
it lowers plus scores. (3) Multiply d elementwise by the final
layer-norm's scale weights, so the direction lives in the same
coordinates as the residual stream the layers write. (4) Normalize to
unit length: \ensuremath{\hat{d}} = d / \ensuremath{\|}d\ensuremath{\|} --- the sign direction. (5) To remove the
direction at a layer, subtract \ensuremath{\alpha} times the projection of that layer's
output h onto \ensuremath{\hat{d}}, at the sign position during the teacher-forced run:

\emph{h \ensuremath{\leftarrow} h \ensuremath{-} \ensuremath{\alpha} (h \ensuremath{\cdot} \ensuremath{\hat{d}}) \ensuremath{\hat{d}}}\quad (\ensuremath{\alpha} = dose)

At \ensuremath{\alpha} = 1 the layer's \ensuremath{\hat{d}}-component is removed exactly; higher doses push
the same amount past zero. Both stages below use this one mechanic; the
paper's shorthand for it is subtracting the minus direction
(minus-removal).

Construction uses only model weights and discovery data --- never
held-out cases. The same recipe applies to all five models; only the
targeted habit layers differ per model (the frozen sets of \S{}4):
Gemma-3-27B L58 + L60 (controls L23, L39); Mistral-Small-24B L37/L38/L39
(controls L15, L25); Phi-4-14B L37 + L39 (controls L15, L25);
Llama-3.3-70B L77 + L79 (controls L30, L50); Qwen-2.5-72B L75 + L78
(controls L30, L50).

\textbf{H.3 Stage s06\_direction\_subtraction --- single-layer
dose--response}

\textbf{H.3.1 Design and eligibility}

The stage's question: is the sign direction a real causal handle ---
does removing it shift which sign gets written --- or merely a
correlate? The design: on determinant errors, remove the direction (the
projection subtraction of H.2, step 5) at one layer at a time --- each
writer (habit) layer, each control layer, and a magnitude-matched random
direction --- and measure the change in the written sign's logit lead. A
case enters only if its logit-lens peak is above layer 5 --- an earlier
peak means there is no late circuit to test (Appendix C); the same
frozen rule governs the s08 error runs (H.4.3).

Like the s08 battery, this stage pools discovery + held-out (the H.4.2
formula: counts add, means are case-weighted). The two phases ran the
identical experiment --- same script, same layers, same doses \ensuremath{\alpha} =
0.5/1/2 --- for four of the five models, and their case sets are
disjoint. The exception is Phi-4-14B, reported on discovery data only:
at the standard doses its held-out run barely moves --- at most 4 of 146
cases flip (writer L39 at \ensuremath{\alpha} = 3; writer L37: zero at every dose \ensuremath{\alpha} \ensuremath{\le} 3)
--- because Phi's single-layer projections are small against its
oversized margins. The dose grid was therefore deliberately escalated to
\ensuremath{\alpha} = 5/8/12, and those are the informative Phi cells; the two grids share
no dose, so Phi's phases cannot be summed.

\begin{longtable}[]{@{}>{\raggedright\arraybackslash}p{0.103\textwidth}>{\raggedright\arraybackslash}p{0.054\textwidth}>{\raggedright\arraybackslash}p{0.271\textwidth}>{\raggedright\arraybackslash}p{0.187\textwidth}>{\raggedright\arraybackslash}p{0.151\textwidth}>{\raggedright\arraybackslash}p{0.144\textwidth}@{}}
\toprule
\textbf{Model} & \textbf{Phases} & \textbf{n discovery} & \textbf{n
held-out} & \textbf{n used} & \textbf{Doses \ensuremath{\alpha}}\tabularnewline
\midrule
\endhead
Llama-3.3-70B & pooled & 15 & 189 & 204 & 0.5, 1, 2\tabularnewline
Qwen-2.5-72B & pooled & 44 & 99 & 143 & 0.5, 1, 2\tabularnewline
Gemma-3-27B & pooled & 89 & 126 & 215 & 0.5, 1, 2\tabularnewline
Phi-4-14B & discovery only & 57 & --- & 57 & 5, 8, 12\tabularnewline
Mistral-Small-24B & pooled & 91 & 132 & 223 & 0.5, 1, 2\tabularnewline
\bottomrule
\end{longtable}

\textbf{Table H.1 --- Cases entering the single-layer dose--response.}
Every n is the late-circuit subset (logit-lens peak \textgreater{} layer
5) fixed in the released files; the two phases' case sets are disjoint,
and pooled rows follow the s08 formula (H.4.2). Phi is discovery-only at
deliberately escalated doses (see text). Qwen's 99 held-out cases are 74
plus-writers + 25 minus-writers.

\textbf{H.3.2 Results}

\textbf{One worked example --- Llama-3.3-70B at \ensuremath{\alpha} = 2.} At writer layer
L79, removing the direction produces a mean change in written-sign lead
of \ensuremath{-}1.83 logits and flips 32 of 204 cases to the correct sign. At
control layer L50 the same dose yields \ensuremath{-}0.02 --- flat. The random
direction yields +0.000 with zero flips --- null. That three-way
contrast (writer strong, control flat, random null) is the core
validation, and it holds monotonically across \ensuremath{\alpha} = 0.5, 1, 2 (Table H.2).

\begin{longtable}[]{@{}>{\raggedright\arraybackslash}p{0.103\textwidth}>{\raggedright\arraybackslash}p{0.054\textwidth}>{\raggedright\arraybackslash}p{0.271\textwidth}>{\raggedright\arraybackslash}p{0.187\textwidth}>{\raggedright\arraybackslash}p{0.151\textwidth}>{\raggedright\arraybackslash}p{0.144\textwidth}@{}}
\toprule
\textbf{Model} & \textbf{Layer} & \textbf{\ensuremath{\Delta} at \ensuremath{\alpha} = 0.5} & \textbf{\ensuremath{\Delta} at \ensuremath{\alpha}
= 1} & \textbf{\ensuremath{\Delta} at \ensuremath{\alpha} = 2} & \textbf{Flips (\ensuremath{\alpha} = 0.5 / 1 /
2)}\tabularnewline
\midrule
\endhead
Llama-3.3-70B (n = 204) & L79 (writer) & \ensuremath{-}0.46 & \ensuremath{-}0.91 & \ensuremath{-}1.83 & 5 / 13
/ 32\tabularnewline
& L77 (writer) & \ensuremath{-}0.21 & \ensuremath{-}0.44 & \ensuremath{-}0.90 & 2 / 2 / 6\tabularnewline
& L30 (control) & +0.02 & +0.00 & +0.05 & 1 / 1 / 1\tabularnewline
& L50 (control) & \ensuremath{-}0.01 & \ensuremath{-}0.01 & \ensuremath{-}0.02 & 1 / 2 / 2\tabularnewline
& random dir. & --- & +0.000 & --- & 0\tabularnewline
Qwen-2.5-72B (n = 143) & L75 (writer) & \ensuremath{-}0.04 & \ensuremath{-}0.07 & \ensuremath{-}0.13 & 0 / 0 /
0\tabularnewline
& L78 (writer) & \ensuremath{-}0.03 & \ensuremath{-}0.07 & \ensuremath{-}0.11 & 1 / 1 / 1\tabularnewline
& L30 (control) & +0.01 & +0.02 & +0.01 & 0 / 0 / 0\tabularnewline
& L50 (control) & +0.01 & +0.01 & \ensuremath{-}0.02 & 0 / 0 / 0\tabularnewline
& random dir. & --- & +0.000 & --- & 0\tabularnewline
\bottomrule
\end{longtable}

\textbf{Table H.2 --- Single-layer removal, Llama-3.3-70B and
Qwen-2.5-72B (pooled discovery + held-out).} \ensuremath{\Delta} = mean change in the
written sign's logit lead (negative = toward the correct sign); n as in
Table H.1. The held-out portions of these curves are scored in Appendix
J (prediction P4).

\begin{longtable}[]{@{}>{\raggedright\arraybackslash}p{0.102\textwidth}>{\raggedright\arraybackslash}p{0.054\textwidth}>{\raggedright\arraybackslash}p{0.271\textwidth}>{\raggedright\arraybackslash}p{0.187\textwidth}>{\raggedright\arraybackslash}p{0.151\textwidth}>{\raggedright\arraybackslash}p{0.144\textwidth}@{}}
\toprule
\textbf{Model} & \textbf{Layer} & \textbf{\ensuremath{\Delta} at low \ensuremath{\alpha}} & \textbf{\ensuremath{\Delta} at mid
\ensuremath{\alpha}} & \textbf{\ensuremath{\Delta} at high \ensuremath{\alpha}} & \textbf{Flips (low / mid /
high)}\tabularnewline
\midrule
\endhead
Gemma-3-27B (n = 215; \ensuremath{\alpha} = 0.5/1/2) & L58 (writer) & \ensuremath{-}0.031 & \ensuremath{-}0.059 &
\ensuremath{-}0.145 & 0 / 0 / 3\tabularnewline
& L60 (writer) & \ensuremath{-}0.143 & \ensuremath{-}0.281 & \ensuremath{-}0.560 & 2 / 6 / 18\tabularnewline
& L23 (control) & +0.002 & \ensuremath{-}0.001 & \ensuremath{-}0.001 & 2 / 1 / 3\tabularnewline
& L39 (control) & \ensuremath{-}0.013 & \ensuremath{-}0.013 & +0.015 & 1 / 0 / 2\tabularnewline
Phi-4-14B (n = 57, discovery; \ensuremath{\alpha} = 5/8/12) & L37 (writer) & \ensuremath{-}5.604 &
\ensuremath{-}9.501 & \ensuremath{-}15.008 & 0 / 3 / 12\tabularnewline
& L39 (writer) & \ensuremath{-}4.543 & \ensuremath{-}7.350 & \ensuremath{-}10.991 & 0 / 8 / 10\tabularnewline
& L15 (control) & \ensuremath{-}0.025 & \ensuremath{-}0.034 & \ensuremath{-}0.067 & 0 / 0 / 0\tabularnewline
& L25 (control) & \ensuremath{-}0.326 & \ensuremath{-}0.500 & \ensuremath{-}0.785 & 0 / 0 / 1\tabularnewline
Mistral-Small-24B (n = 223; \ensuremath{\alpha} = 0.5/1/2) & L37 (writer) & \ensuremath{-}0.083 &
\ensuremath{-}0.174 & \ensuremath{-}0.316 & 1 / 1 / 1\tabularnewline
& L38 (writer) & \ensuremath{-}0.059 & \ensuremath{-}0.110 & \ensuremath{-}0.242 & 0 / 1 / 2\tabularnewline
& L39 (writer) & \ensuremath{-}0.040 & \ensuremath{-}0.085 & \ensuremath{-}0.193 & 0 / 1 / 6\tabularnewline
& L15 (control) & +0.009 & +0.001 & +0.006 & 0 / 1 / 1\tabularnewline
& L25 (control) & \ensuremath{-}0.001 & \ensuremath{-}0.006 & \ensuremath{-}0.014 & 1 / 0 / 1\tabularnewline
\bottomrule
\end{longtable}

\textbf{Table H.3 --- Single-layer removal, Gemma-3-27B, Phi-4-14B and
Mistral-Small-24B.} Gemma and Mistral pooled (discovery + held-out); Phi
discovery only, at the deliberately escalated doses (Table H.1). Same
layout as Table H.2; n and doses as in Table H.1. The random direction
flips zero cases in every model (Phi's random control is stored as a
per-case edit site). Phi's large \ensuremath{\Delta} values at high doses mirror its
oversized baseline margins.

\textbf{H.3.3 Takeaways --- and how they line up with stage s08 (H.4)}

Model by model: who flips most from a single layer, and whether the
two-layer battery of H.4 says the same thing or the contrary.

\begin{itemize}
\item
  \textbf{Llama-3.3-70B --- most single-layer flips; H.4 agrees.} 32/204
  flips from L79 alone at \ensuremath{\alpha} = 2 (Table H.2) --- the largest one-layer
  effect of any model. The battery agrees: 42/121 det and 21/26 ibp
  minus-written flips (H.4.4--H.4.5).
\item
  \textbf{Phi-4-14B --- flips only at extreme doses; H.4 agrees (the
  null).} 0 flips at \ensuremath{\alpha} = 5, 12/57 only at \ensuremath{\alpha} = 12 (Table H.3). The
  battery at \ensuremath{\alpha} = 3 stays null --- 6/173 det, 4/108 ibp --- despite the
  largest mean pushes: its oversized baseline margins absorb them
  (H.4.7).
\item
  \textbf{Gemma-3-27B --- modest alone, strong in the pair; H.4 adds the
  second layer.} 18/215 flips from its best single layer (L60, \ensuremath{\alpha} = 2;
  Table H.3), but 33/84 det and 33/61 ibp once both habit layers are
  steered together (H.4.4--H.4.5) --- the pair does what neither layer
  does alone.
\item
  \textbf{Qwen-2.5-72B --- the contrary case.} At most 1 flip from any
  single layer at any dose (Table H.2), yet the battery flips 11/39 det
  and 13/35 ibp (H.4.4--H.4.5): its stamp is distributed across layers,
  so one-layer removal understates the causal handle. This redundancy is
  the preregistered P4 exception; the joint ablation is in Appendix G.
\item
  \textbf{Mistral-Small-24B --- same signature as Qwen, milder.} 6/223
  flips from its best single layer (L39, \ensuremath{\alpha} = 2; Table H.3), yet the
  largest battery counts of any model --- 62/185 det, 69/91 ibp --- with
  the direction reaching only two of its three habit layers (H.4.1).
\end{itemize}

Controls stay flat everywhere: at the shared doses (\ensuremath{\alpha} \ensuremath{\le} 2) every
control-layer mean sits within \ensuremath{\pm}0.05 logits, control flips never exceed
3 of 215, and even Phi's \ensuremath{\alpha} = 12 controls reach at most \ensuremath{-}0.79 against
\ensuremath{-}15.0 at its writers; the random direction flips zero cases in every
model (Tables H.2--H.3) --- the deviations above are about writer
layers, not the method.

\textbf{H.4 Stage s08\_c2\_steering --- the C2 steering battery}

\textbf{H.4.1 Protocol and steered sites}

A reminder of the C2 experiment (\S{}5). The sign direction was extracted
on discovery data and then frozen. The C2 condition --- minus-removal
steering --- removes the direction from the activations (the projection
subtraction of H.2, step 5) at the frozen habit-layer sites, at the
error-site sign token (the matched sign position on correct cases), at
steering strength \ensuremath{\alpha} \ensuremath{\in} \{1, 2, 3\}; all evaluation is teacher-forced, and
the outcome measure is the sign lead: the written sign's logit minus the
alternative's. A flip (on errors) or a break (on corrects) is a strict
zero-crossing of that lead.

The steered sites, checked against the frozen habit sets of \S{}4. The C2
protocol injects the direction at exactly two sites per model --- a
fixed design choice, identical for all five models. For Gemma-3-27B (L58
+ L60), Phi-4-14B (L37 + L39) and Qwen-2.5-72B (L75 + L78, MLP outputs),
the steered pair is the frozen habit set of \S{}4, exactly, in every
released target file. Mistral-Small-24B's frozen habit set is the one
with three members (L37/L38/L39), so the two protocol slots cover its
deepest pair, L38 + L39 --- frozen in every run's metadata before
scoring. If anything this understates Mistral: the direction reaches two
of its three habit layers, and Mistral still posts the largest
minus-written flip counts of any model (62/185 det, 69/91 ibp; Tables
H.4--H.5). Llama-3.3-70B steers its per-task frozen pairs: L77 + L79 on
determinants (the \S{}4 set) and L77 + L78 on integration by parts, per the
frozen, human-reviewed IBP target file, in which the minus-writer DLA
credit concentrates at L78 (mean +3.35 logits on minus-written discovery
cases, robust to leave-one-out).

This appendix asks whether the edit is specific to the sign computation,
separately on each task. The test has three parts:

\begin{itemize}
\item
  \textbf{It flips minus-written errors.} On errors where the model
  wrote ``\ensuremath{-}'', C2 at \ensuremath{\alpha} = 3 flips (pooled) Gemma 33/84, Qwen 11/39, Llama
  42/121 and Mistral 62/185 on determinants, and Gemma 33/61, Qwen
  13/35, Llama 21/26 and Mistral 69/91 on integration by parts (Phi:
  6/173 and 4/108 --- see H.4.7). Almost every flip is corrective.
\item
  \textbf{It breaks correct-minus answers.} On problems solved correctly
  with a true ``\ensuremath{-}'', the same edit pushes many toward ``+'' (Llama
  131/138 det and 24/24 ibp; Gemma 74/126 det) --- the double edge of
  \S{}5. The edit targets the sign computation itself, not ``errors''.
\item
  \textbf{What it does to ``+'' separates the models --- and the tasks.}
  On plus-written cases the stamp models (Gemma, Qwen) are untouched or
  helped, the amplifier models (Llama, and Mistral on det) flip them
  too, and nearly all of this off-target traffic lives on the
  determinant task (H.4.7).
\end{itemize}

\textbf{H.4.2 Pooling: discovery + held-out}

\textbf{Definition.} For each model, task and cohort type, the pooled
count is the sum of the discovery-phase and held-out-phase counts for
the identical cohort, and the pooled mean is the case-weighted mean over
the union of cases. The two tasks are never mixed: every figure and
table reports determinants and integration by parts separately, stacked.

\textbf{Why pooling across the phase split is sound.} The sign direction
and the habit-layer targets were extracted on discovery data and frozen
before any held-out case was scored; the held-out phase is the
out-of-sample confirmation, and its per-task \ensuremath{\alpha} = 3 endpoints stand on
their own in Figure 5 of the main text. Both phases then run the
identical C2 protocol on identically defined cohorts, so pooling changes
no definition --- it only adds cases. For a mechanistic-interpretability
battery whose per-cell n is necessarily small, the pooled view gives the
per-task estimates more statistical weight; Tables H.4--H.5 keep the
per-phase rows alongside the pooled rows, so nothing is hidden by the
sum.

\textbf{H.4.3 Coverage: where every case comes from}

Cohort sizes trace to Appendix A. The determinant correct sets match
Table A2 exactly, phase by phase (e.g. Gemma 50 discovery + 138 held-out
= 188); the held-out cells of every cohort match Table 1 exactly; the
IBP discovery/held-out splits match the chain in \S{}2 (Gemma 143 = 70 +
73, Mistral 109 = 39 + 70, Phi 117 = 40 + 77, Llama 43 = 22 + 21, Qwen
36 = 10 + 26).

The error steering runs cover exactly the late-circuit cases: the
preregistered s08 filter keeps cases whose logit-lens peak layer is
deeper than 5. Verified case by case against the experiment-ready files,
every plus-written error in every cell is late-circuit --- plus-written
coverage is complete everywhere --- and every excluded case is an
early-peaking minus-written error: Gemma 6, Qwen 64 (48 of them in det
held-out alone), Llama 7 (3 det held-out, 4 IBP), Phi 9, Mistral 21 (19
early-peak, 2 dropped by deterministic tokenization checks). All
exclusions are mechanical filters fixed before scoring; none is
outcome-based. The held-out minus-written denominators here equal Figure
5's denominators exactly.

Two cells have no run and are shown as gaps, not zeros. The frozen
steering protocol specifies error steering on held-out minus-written
errors; the discovery-phase error steering reported in the other cells
was run in addition to that protocol. Llama det is the one error cell
without the additional discovery run, so its held-out row is the
protocol-complete result, not a shortfall. The second gap is the Qwen
ibp discovery correct cohort (25 cases exist; Qwen's ibp correct rows
are held-out only). Separately, Qwen ibp has zero plus-written cases in
both phases --- a fact about the model's errors, not about coverage.

\textbf{H.4.4 Results --- 4\ensuremath{\times}4 determinant}

\begin{longtable}[]{@{}>{\raggedright\arraybackslash}p{0.102\textwidth}>{\raggedright\arraybackslash}p{0.054\textwidth}>{\raggedright\arraybackslash}p{0.271\textwidth}>{\raggedright\arraybackslash}p{0.187\textwidth}>{\raggedright\arraybackslash}p{0.151\textwidth}>{\raggedright\arraybackslash}p{0.144\textwidth}@{}}
\toprule
\textbf{Model} & \textbf{Phase} & \textbf{Error --- minus-written:
flipped (mean \ensuremath{\Delta} lead)} & \textbf{Error --- plus-written: flipped} &
\textbf{Correct --- minus: broken} & \textbf{Correct --- plus:
broken}\tabularnewline
\midrule
\endhead
Gemma-3-27B & discovery & 16/43 (\ensuremath{-}6.94) & 1/46 & 4/33 &
0/17\tabularnewline
& held-out & 17/41 (\ensuremath{-}5.37) & 1/85 & 70/93 & 6/45\tabularnewline
& \textbf{pooled} & \textbf{33/84 (\ensuremath{-}6.17)} & \textbf{2/131} &
\textbf{74/126} & \textbf{6/62}\tabularnewline
Qwen-2.5-72B & discovery & 5/14 (\ensuremath{-}7.30) & 0/30 & 17/36 &
0/11\tabularnewline
& held-out & 6/25 (\ensuremath{-}8.14) & 0/74 & 49/123 & 0/18\tabularnewline
& \textbf{pooled} & \textbf{11/39 (\ensuremath{-}7.84)} & \textbf{0/104} &
\textbf{66/159} & \textbf{0/29}\tabularnewline
Llama-3.3-70B & discovery & --- & --- & 57/62 & 0/18\tabularnewline
& held-out & 42/121 (\ensuremath{-}4.52) & 29/68 & 74/76 & 0/54\tabularnewline
& \textbf{pooled} & \textbf{42/121 (\ensuremath{-}4.52)} & \textbf{29/68} &
\textbf{131/138} & \textbf{0/72}\tabularnewline
Phi-4-14B & discovery & 1/51 (\ensuremath{-}8.41) & 0/6 & 0/39 & 0/11\tabularnewline
& held-out & 5/122 (\ensuremath{-}8.96) & 1/24 & 2/137 & 1/12\tabularnewline
& \textbf{pooled} & \textbf{6/173 (\ensuremath{-}8.80)} & \textbf{1/30} &
\textbf{2/176} & \textbf{1/23}\tabularnewline
Mistral-Small-24B & discovery & 23/71 (\ensuremath{-}0.17) & 14/20 & 39/57 &
16/21\tabularnewline
& held-out & 39/114 (\ensuremath{-}1.00) & 7/18 & 53/90 & 26/32\tabularnewline
& \textbf{pooled} & \textbf{62/185 (\ensuremath{-}0.68)} & \textbf{21/38} &
\textbf{92/147} & \textbf{42/53}\tabularnewline
\bottomrule
\end{longtable}

\textbf{Table H.4 --- 4\ensuremath{\times}4 determinant, C2 at \ensuremath{\alpha} = 3.} Flip/break counts
as k/n; in parentheses, the mean shift of the written sign's lead
(logits) on minus-written errors. Pooled rows (bold) are sums of the two
phase rows; pooled means are case-weighted. ``---'' = no run in that
phase: the preregistered plan steered held-out errors only;
discovery-phase steering was an extra run added in the other cells
(H.4.3).

\textbf{What stands out.} Of the pooled minus-written flips, corrective
flips are Gemma 33/33, Qwen 11/11, Phi 6/6, Llama 33/42 and Mistral
52/62 --- the \S{}5 distinction between flipping and repairing matters only
for the amplifiers. The plus-written column is the det story: Llama
flips 29/68 plus-written errors to the correct sign (mean \ensuremath{-}8.50 logits)
and Mistral 21/38 (mean \ensuremath{-}5.46, and already 14/20 on discovery data),
while Gemma and Qwen flip 2/131 and 0/104 and their plus-written leads
grow (+1.62 and +2.44 logits). This also resolves \S{}5's all-error flip
counts: Llama det's 71 = 42 minus-written + 29 plus-written.

\textbf{H.4.5 Results --- integration by parts}

\begin{longtable}[]{@{}>{\raggedright\arraybackslash}p{0.102\textwidth}>{\raggedright\arraybackslash}p{0.054\textwidth}>{\raggedright\arraybackslash}p{0.271\textwidth}>{\raggedright\arraybackslash}p{0.187\textwidth}>{\raggedright\arraybackslash}p{0.151\textwidth}>{\raggedright\arraybackslash}p{0.144\textwidth}@{}}
\toprule
\textbf{Model} & \textbf{Phase} & \textbf{Error --- minus-written:
flipped (mean \ensuremath{\Delta} lead)} & \textbf{Error --- plus-written: flipped} &
\textbf{Correct --- minus: broken} & \textbf{Correct --- plus:
broken}\tabularnewline
\midrule
\endhead
Gemma-3-27B & discovery & 18/33 (\ensuremath{-}8.19) & 0/35 & 1/25 &
0/25\tabularnewline
& held-out & 15/28 (\ensuremath{-}3.96) & 0/44 & 63/150 & 0 cases\tabularnewline
& \textbf{pooled} & \textbf{33/61 (\ensuremath{-}6.25)} & \textbf{0/79} &
\textbf{64/175} & \textbf{0/25}\tabularnewline
Qwen-2.5-72B & discovery & 3/10 (\ensuremath{-}7.95) & 0 cases & --- &
---\tabularnewline
& held-out & 10/25 (\ensuremath{-}7.20) & 0 cases & 0/50 & 0 cases\tabularnewline
& \textbf{pooled} & \textbf{13/35 (\ensuremath{-}7.41)} & \textbf{0 cases} &
\textbf{0/50} & \textbf{0 cases}\tabularnewline
Llama-3.3-70B & discovery & 8/12 (\ensuremath{-}14.52) & 0/7 & 23/23 & 0
cases\tabularnewline
& held-out & 13/14 (\ensuremath{-}20.72) & 0/6 & 1/1 & 0/46\tabularnewline
& \textbf{pooled} & \textbf{21/26 (\ensuremath{-}17.86)} & \textbf{0/13} &
\textbf{24/24} & \textbf{0/46}\tabularnewline
Phi-4-14B & discovery & 1/40 (\ensuremath{-}14.58) & 0 cases & 0/37 &
0/13\tabularnewline
& held-out & 3/68 (\ensuremath{-}15.11) & 0/8 & 0/93 & 0/53\tabularnewline
& \textbf{pooled} & \textbf{4/108 (\ensuremath{-}14.91)} & \textbf{0/8} &
\textbf{0/130} & \textbf{0/66}\tabularnewline
Mistral-Small-24B & discovery & 30/35 (\ensuremath{-}8.03) & 0 cases & 15/23 &
17/20\tabularnewline
& held-out & 39/56 (\ensuremath{-}6.05) & 2/11 & 71/91 & 62/65\tabularnewline
& \textbf{pooled} & \textbf{69/91 (\ensuremath{-}6.81)} & \textbf{2/11} &
\textbf{86/114} & \textbf{79/85}\tabularnewline
\bottomrule
\end{longtable}

\textbf{Table H.5 --- integration by parts, C2 at \ensuremath{\alpha} = 3.} Same layout as
Table H.4. ``0 cases'' marks empty cohorts (the model produced no such
cases in that phase); ``---'' marks the one missing Qwen discovery
correct run. Corrective flips among the pooled minus-written flips:
Gemma 33, Qwen 13, Llama 21, Phi 4, Mistral 67.

\textbf{What stands out.} The same edit bites harder on target: Llama
flips 21/26 (81\%) at a mean push of \ensuremath{-}17.9 logits and Mistral 69/91
(76\%), against 35\% and 34\% on determinants. And it spills less off
target: across all five models the ibp plus-written cells flip 2 cases
in total (both Mistral), against 53 on determinants. The glaring
exception is Mistral's correct-plus cohort, 79/85 broken --- on
corrects, Mistral's ibp direction is anything but clean (H.4.7).

\textbf{H.4.6 The pooled figures}

\begin{center}\includegraphics[width=\linewidth]{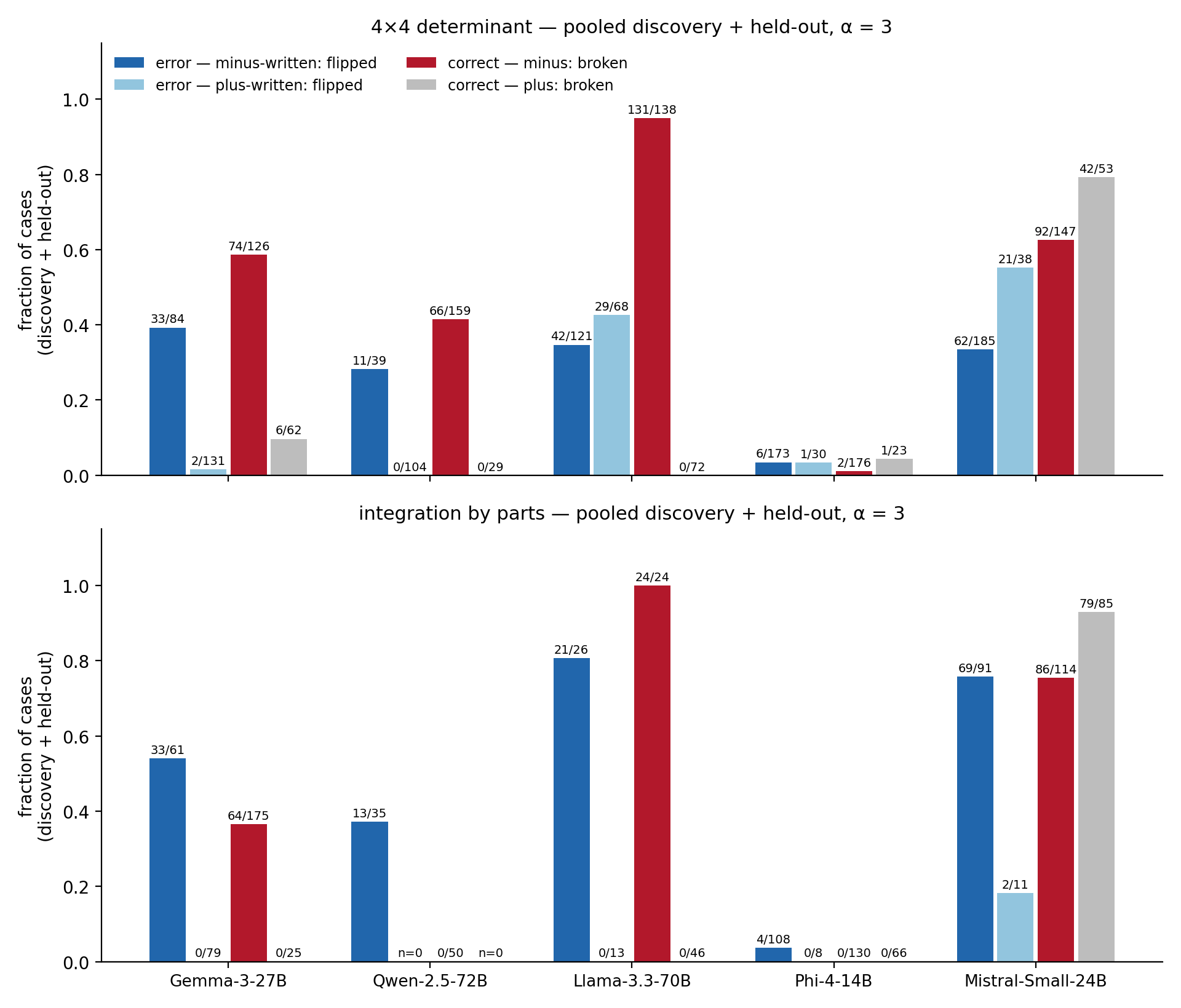}\end{center}

\textbf{Figure H.1 --- Specificity fingerprint, pooled discovery +
held-out, \ensuremath{\alpha} = 3.} Top: 4\ensuremath{\times}4 determinant; bottom: integration by parts.
Four bars per model, tagged as in Tables H.4--H.5: error ---
minus-written (flipped), error --- plus-written (flipped), correct ---
minus (broken), correct --- plus (broken); k/n above each bar. Llama's
det bars pool the held-out phase only (H.4.3); Qwen ibp has no
plus-written cases in either phase.

\begin{center}\includegraphics[width=\linewidth]{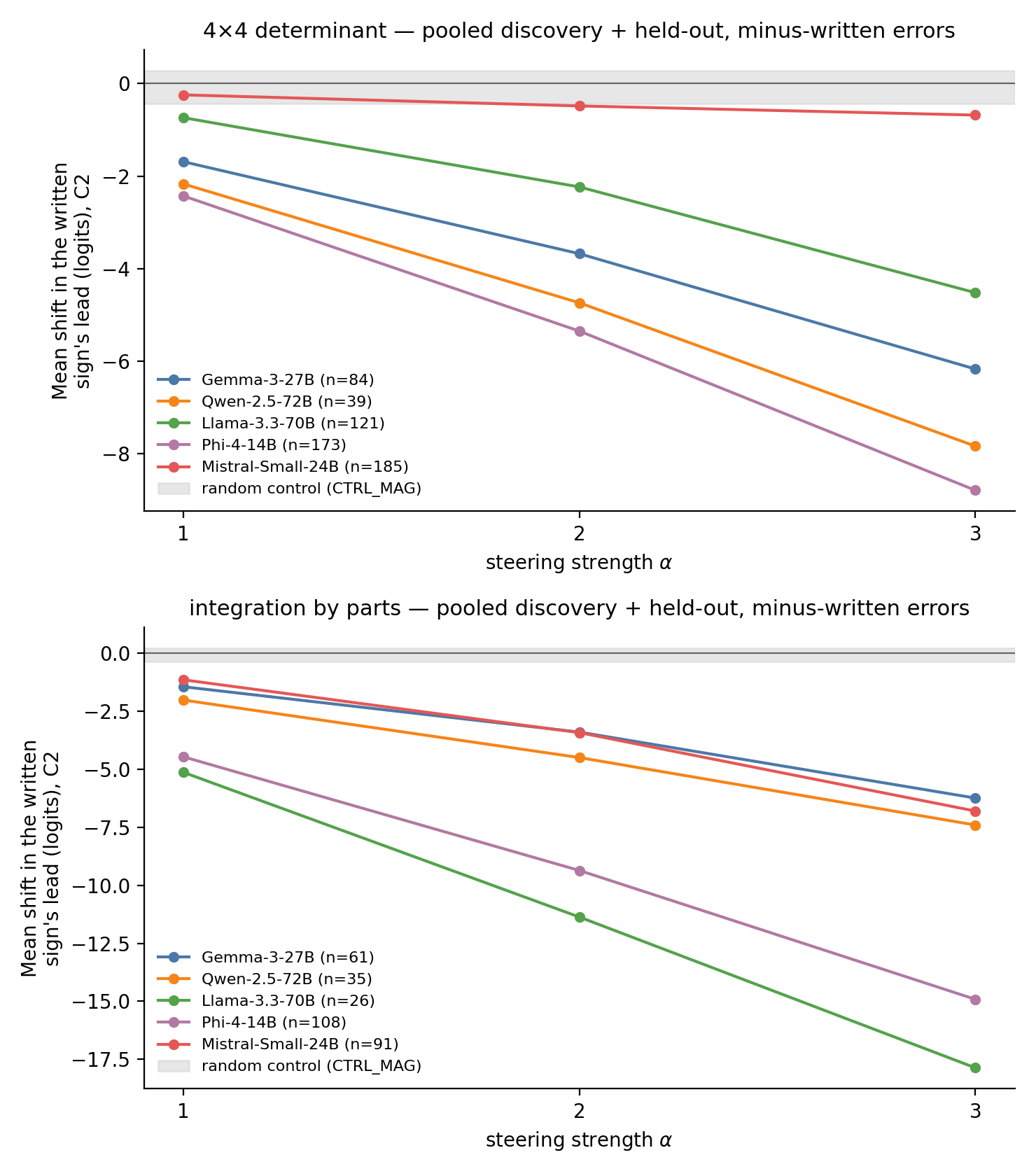}\end{center}

\textbf{Figure H.2 --- Dose--response, pooled discovery + held-out,
minus-written errors.} Top: 4\ensuremath{\times}4 determinant; bottom: integration by
parts. Mean shift in the written sign's lead under C2 versus steering
strength \ensuremath{\alpha}, one line per model (legend gives pooled n); the grey band
spans the CTRL\_MAG control means. Note the y-scales: the ibp panel
reaches \ensuremath{-}18 logits.

\textbf{H.4.7 What it means}

\textbf{Stamp versus amplifier, now causal.} Figure 2 of the main text
reads the habit layers' DLA split by written sign --- a correlational
fingerprint. The plus-written steering column is its causal test, and
the two agree model by model. Qwen and Gemma behave as fixed minus
stamps: removing the stamp cannot hurt a written ``+'', and indeed their
plus-written det leads grow (+2.44, +1.62 logits; 0/104 and 2/131
flips). Llama behaves as an amplifier of whatever sign is written:
subtracting the direction suppresses the written ``+'' too (29/68 det
flips, mean \ensuremath{-}8.50). Mistral joins the amplifier camp on determinants ---
21/38 plus-written flips, already 14/20 on discovery data --- so a
reading of Mistral as ``intermediate'', based on the held-out phase
alone (7/18), understates it. Phi's plus cells are small and mixed (1/30
det at mean \ensuremath{-}2.47), consistent with its L37-stamp / L39-amplifier split
in Figure 2. Plus cells with n \ensuremath{\le} 13 (Llama ibp, Mistral ibp, Phi ibp)
are suggestive only.

\textbf{Specificity is task-dependent.} Integration by parts is the more
sensitive task on target: at \ensuremath{\alpha} = 3 the mean push on minus-written errors
is \ensuremath{-}17.9 versus \ensuremath{-}4.5 logits for Llama, \ensuremath{-}14.9 versus \ensuremath{-}8.8 for Phi, and
\ensuremath{-}6.8 versus \ensuremath{-}0.7 for Mistral, with flip rates to match (81\% vs 35\%,
76\% vs 34\%). Yet it is the cleaner task off target: the plus-written
mean pushes on ibp sit near the control band (Gemma \ensuremath{-}1.89, Llama \ensuremath{-}2.21,
Phi +0.84, Mistral +0.86 logits) while on det they are large for the
amplifiers and positive for the stamps. The per-task directions
evidently overlap the plus-written computation to very different degrees
--- ibp's direction is nearly orthogonal to it --- so specificity is a
property of the task's direction, not of the model alone. Three caveats:
Qwen's empty ibp plus cells are silence, not evidence of cleanliness;
Gemma's ibp correct-plus cohort is 25 discovery cases; and Mistral's
79/85 broken ibp correct-plus answers show its ibp direction is
emphatically not clean on corrects --- the one large ibp off-target
effect.

\textbf{The double edge, and why Phi is the null.} Correct-minus
breakage confirms the double edge wherever the push clears the margin:
Llama's correct-minus cohorts are essentially wiped out (131/138 det,
24/24 ibp), Gemma breaks 74/126 on det. Phi barely moves in any cohort
(6/173, 4/108 flipped; 2/176, 0/130 broken) even though its on-target
push is among the largest (\ensuremath{-}14.9 logits on ibp): its oversized baseline
margins --- median sign lead near 25 logits --- absorb the push, the \S{}6
story. Breakage is also cohort-dependent: Gemma breaks 4/33 discovery
correct-minus but 70/93 held-out. The discovery correct sets are the
deepsign role-matched selections used for layer-finding, not random
draws, so per-phase rates differ; the pooled row reports the union of
the two honestly.

\textbf{H.5 Provenance and verification}

\textbf{Pooling formula.} Pooled count = discovery count + held-out
count for the identical (model, task, cohort) cell. Pooled mean =
(n\_d\ensuremath{\cdot}m\_d + n\_h\ensuremath{\cdot}m\_h) / (n\_d + n\_h), i.e. the case-weighted mean
over the union of cases.

\textbf{Source files.} The steering battery (H.4) comes from
\{Model\}/\{discovery\textbar heldout\}\_*/s08\_c2\_steering/ --- errors
from *error*combined\_ablation.json, controls from
*error*ctrl\_mag.json, corrects from *c2\_on\_correct.json. The
single-layer dose--response (H.3) comes from the
s06\_direction\_subtraction/ JSONs of both phases (Phi: discovery only;
Table H.1); its eligibility lists (peak \textgreater{} 5) are
recomputable from the s01\_logit\_lens records.

\textbf{Sign convention on corrects.} written\_sign on correct cases
carries the sign present in the text at the matched site (deepsign
role-matched inputs), verified per-case against the released input
files.

\textbf{Controls.} Across every CTRL\_MAG evaluation in the error runs
of both phases --- 4,215 case \ensuremath{\times} dose evaluations --- exactly one case
flips (Llama det held-out, \ensuremath{\alpha} = 3); across all 5,100 control evaluations
on correct cohorts, zero break. The grey band in Figure H.2 spans the
per-model CTRL\_MAG means, all within \ensuremath{\pm}0.28 logits of zero.

\textbf{Verification.} Every count and mean in this appendix and both
figures was recomputed from the raw per-case JSONs and cross-checked
against paper/appendix/ledger\_results.csv (stage s08\_c2\_steering):
pooled counts equal the sums of the per-phase ledger rows
(n\_flipped\_minus{[}C2\textbar a\ensuremath{\alpha}{]}; n\_flipped \ensuremath{-} n\_flipped\_minus
for plus-written; n\_broken{[}C2\textbar a3.0{]}; n\_wrote\_minus\_all),
and pooled means equal the case-weighted averages of the per-phase
mean\_delta\_ld\_minus{[}C2\textbar a\ensuremath{\alpha}{]} rows. All match. Every \ensuremath{\Delta}, flip
count and n in Tables H.2--H.3 likewise matches the stage-s06 ledger
rows (mean\_delta\_ld{[}layer\textbar \ensuremath{\alpha}{]},
n\_flipped{[}layer\textbar \ensuremath{\alpha}{]}, n\_cases\_all): the per-phase ledger
rows match the raw JSONs, the pooled counts equal their sums, and the
pooled means their case-weighted averages. The construction and removal
mechanic stated in H.2 match the release code: the same
unembedding-difference \ensuremath{\hat{d}} and the same projection-subtraction hook appear
in the experiment scripts of both stages. Both figures regenerate
end-to-end from paper/appendix/fig\_G\_specificity.py.

\clearpage
\section*{Appendix I --- Preregistration and the Held-Out Scorecard
(P1--P6)}
\addcontentsline{toc}{section}{Appendix I --- Preregistration and the Held-Out Scorecard
(P1--P6)}

\emph{Before touching any held-out data, we wrote down --- and locked
with git commits --- exactly what we would measure and what number would
count as a pass. This appendix shows what was locked, and what the data
then did. Figure 6 (\S{}6) is the one-glance summary; this is the evidence
behind each of its cells.}

\textbf{I.1 Why this appendix exists, and how to read it}

A causal claim is easy to overfit: measure enough things and something
will look like steering. Our protection is preregistration. For each
model we committed a short document to git --- a written contract fixing
the data split, the layers to test, the direction recipe, six
predictions, and a numeric pass bar for each --- before computing any
held-out number for that model. The held-out data then get exactly one
chance to meet each bar, and every outcome is reported, including the
misses.

The appendix reads in this order: I.2 records what was frozen and when
(Table I.1); I.3 states the six predictions in plain words; I.4 explains
how a verdict is decided; I.5--I.9 give one scorecard table per model
--- the frozen bar, the measured value, the verdict; I.10 adds it up.

Four terms cover everything in the tables. \textbf{det} is the
arithmetic task (4\ensuremath{\times}4 determinants where the model writes one wrong sign
mid-computation); \textbf{IBP} is the calculus task (integration by
parts, where the wrong sign appears in the recursion). \textbf{\ensuremath{\Delta} lead}
is the change in the logit lead of the written sign over the opposite
sign --- negative means the intervention moved the model away from what
it wrote. \textbf{\ensuremath{\alpha}} is the dose (strength) of an intervention, and a
\textbf{flip} means the written sign actually changed.

\textbf{I.2 What was frozen, and when}

Six preregistration documents were frozen --- one per model, plus a
separate Llama IBP document. Each lives in the repository in its own
folder (preregistration\_llama/, preregistration\_qwen/,
preregistration\_gemma/, preregistration\_phi/,
preregistration\_mistral/), and each folder was locked with a single git
commit before any held-out number existed for that model. Every freeze
commit contains the same three things:

\begin{itemize}
\item
  the preregistration document itself --- the discovery/held-out split,
  the layer sets and the sign-direction recipe (both found on discovery
  data), the six predictions, and a numeric pass bar for each
  confirmatory one;
\item
  the held-out data files themselves --- the error and correct-answer
  batteries, copied in untouched, so the exact cases to be tested were
  locked before anything was measured on them;
\item
  the frozen target files --- the layer choices carried over from
  discovery data, committed unmodified and never re-derived.
\end{itemize}

Each document's own scope section is the full record of what its battery
covers; where the models differ, the frozen document --- not this
appendix --- is the authority.

The documents were frozen on different dates (August 31 to September 15)
because the models were run in sequence. Each document was frozen before
any held-out number existed for its own model; later documents used
lessons from earlier models to set sensible bars. That is the
replication design --- frozen tests of the same claim, repeated across
models --- not a leak of the data being predicted. No document was
edited after its freeze, with one disclosed exception (Gemma, below),
which only withdrew a pass bar and changed no number.

\begin{longtable}[]{@{}>{\raggedright\arraybackslash}p{0.063\textwidth}>{\raggedright\arraybackslash}p{0.192\textwidth}>{\raggedright\arraybackslash}p{0.700\textwidth}@{}}
\toprule
\textbf{Model} & \textbf{Freeze date (commit)} & \textbf{What else to
know}\tabularnewline
\midrule
\endhead
Llama-3.3-70B & det: 2026-08-31 (03e5549); IBP: 2026-09-01 (fb98035) &
The separate IBP document (12 bars) was adjudicated on its own: 11 of 12
met, CONFIRMED (c418503). Table I.5 and Figure 6 score the det document;
its P6 steering arm is labeled exploratory \dag{}.\tabularnewline
Qwen-2.5-72B & 2026-08-31 (64580f2) & Held-out results committed
2026-09-03 (412976b) and 2026-09-05 (3bb48f8).\tabularnewline
Gemma-3-27B & 2026-09-13 (df73a1a; amended 8d56bb3) & Same-day
amendment, before any measurement: one pass bar was withdrawn (the
correct-answer battery became descriptive), so P5 carries no
verdict.\tabularnewline
Phi-4-14B & 2026-09-15 (67d0d73) & The correct-answer battery is
descriptive by design (no pass bar).\tabularnewline
Mistral-Small-24B & 2026-09-14 (04c2cd2) & The file header still reads
``DRAFT''; the Sep 14 commit is the freeze. Held-out run 2026-09-19
(7e7ea87).\tabularnewline
\bottomrule
\end{longtable}

\textbf{Table I.1 --- The freeze record.} Every freeze commit contains
the three items listed above --- the document, the untouched held-out
data, and the frozen targets --- and was pushed before its model's
held-out run. For what each battery covers, the preregistration document
in each model's folder is the full record.

\textbf{I.3 The six predictions}

The same six predictions appear in every document, each with a frozen
numeric bar. In plain words:

\begin{itemize}
\item
  \textbf{P1 --- errors are decided late.} The layer where the wrong
  sign takes the lead sits in the last stretch of the network, past a
  frozen depth cut-off.
\item
  \textbf{P2 --- habit layers write the sign.} The frozen habit layers
  push the written sign's logit up (positive DLA --- direct logit
  attribution).
\item
  \textbf{P3 --- one direction separates ``\ensuremath{-}'' from ``+''.} Projected
  onto the frozen sign direction, minus-written and plus-written cases
  fall on the sides the plan registered.
\item
  \textbf{P4 --- removing the direction moves errors, dose by dose} (for
  Phi the plan predicts the opposite: a null or backfire).
\item
  \textbf{P5 --- the sign signal also moves correct answers} in the
  registered way.
\item
  \textbf{P6 --- steering flips error cases} while a magnitude-matched
  control direction does nothing.
\end{itemize}

Five models \ensuremath{\times} six predictions = thirty cells. Figure 6 scores
twenty-nine: Gemma's P5 has no pass bar (the pre-measurement amendment
made it descriptive), so it appears on the figure as ``---'' and is
excluded from the denominator.

\textbf{I.4 How verdicts are scored}

A prediction is CONFIRMED if the frozen bar was met, NEAR-MISS if it
missed narrowly (by a margin defined in the preregistration itself),
MISS if the bar was clearly not met, and NOT SCORED if no bar existed.
One rule governs everything: the frozen plan is never revised after
seeing held-out data. Where a frozen site turned out not to be the best
site (Gemma, I.5), the test still runs at the frozen site and the result
stands as measured.

Four footnote symbols, shared with Figure 6: \textbf{\dag{}} Llama's P6
steering arm was added in Phase 2 (no Llama steering file existed at
freeze time); it is labeled exploratory and scored under the frozen
disclosure policy. \textbf{\ddag{}} Phi's P1 IBP arm (92.2\%) narrowly missed
its 95\% bar; the det arm passed clearly. The preregistration itself
defines the cell verdict by the pooled rule --- that rule was frozen in
advance, not chosen after seeing the number, and the near-miss is
disclosed. \textbf{\S{}} Mistral's P3 uses the minus\textgreater plus
projection convention --- the polarity of its direction, fixed from
discovery data in the frozen document. \textbf{\P{}} One frozen rule sets
every P6 flip denominator: the cases whose baseline lead is positive
(for Qwen, its registered target --- minus-written cases). The same rule
simply yields different counts per model: Gemma 66, Mistral 80, Llama
163, Qwen 25, Phi 145.

\textbf{I.5 Gemma-3-27B (62 layers; frozen habit sites L55, L58)}

\emph{The honest-penalty case: the frozen sites were not the best sites,
and we scored them anyway.}

Held-out error set: 126 det, 73 IBP. The preregistration froze L55 and
L58 as the causal test sites; held-out data later placed the strongest
late writers at L58/L60. We did not swap layers after the fact: Table
I.2 scores the frozen sites as written, and Gemma therefore takes four
near-misses. That penalty is the protocol working --- the main text,
which is free to use the held-out-identified sites, reports the strong
effects.

\begin{longtable}[]{@{}>{\raggedright\arraybackslash}p{0.039\textwidth}>{\raggedright\arraybackslash}p{0.498\textwidth}>{\raggedright\arraybackslash}p{0.296\textwidth}>{\raggedright\arraybackslash}p{0.107\textwidth}@{}}
\toprule
\textbf{Pred.} & \textbf{Preregistered bar (held-out)} &
\textbf{Measured value} & \textbf{Verdict}\tabularnewline
\midrule
\endhead
P1 & Decided late: lens peak \ensuremath{\ge} L55 in \ensuremath{\ge} 50\% of cases, per domain & det
54/126 = 42.9\%; IBP 29/73 = 39.7\% & NEAR-MISS\tabularnewline
P2 & Mean MLP DLA \textgreater{} 0 at L55 and L58, both domains & L55
+0.913 (pass); L58 \ensuremath{-}0.353 (fail) & NEAR-MISS\tabularnewline
P3 & Minus-written mean projection \textless{} plus-written at the
capture layers & ordering inverted at L58/L60; only L56 in the
registered direction & NEAR-MISS\tabularnewline
P4 & Joint ablation dose--response: pooled mean \ensuremath{\Delta} lead \ensuremath{\le} \ensuremath{-}1.5 and \ensuremath{\ge} 30\%
of cases beat the null & mean \ensuremath{-}1.235; 33/126 = 26.2\% beat the null &
NEAR-MISS\tabularnewline
P5 & No pass bar (descriptive by pre-measurement amendment) & reported
descriptively only & NOT SCORED\tabularnewline
P6 & Steering flips errors, control flat: mean \ensuremath{\Delta} lead negative and
monotone; strict flips at \ensuremath{\alpha} = 3 \ensuremath{\ge} 5\%; \textbar control\textbar{} \ensuremath{\le} 0.5
& 18/66 = 27.3\% flips at \ensuremath{\alpha} = 3; mean \ensuremath{-}3.33; control \ensuremath{\approx} 0 \P{} &
CONFIRMED\tabularnewline
\bottomrule
\end{longtable}

\textbf{Table I.2 --- Gemma-3-27B held-out scorecard.} 1 CONFIRMED, 4
NEAR-MISS, 1 NOT SCORED.

\textbf{I.6 Mistral-Small-24B (40 layers; habit layers L37, L38, L39)}

\emph{Full confirmation on both tasks, with one narrow IBP miss on
timing depth.}

Held-out error set: 142 det, 70 IBP. The steering stage takes exactly
two layers per model --- a rule applied identically to every model.
Mistral's frozen pair was L38 + L39, so L37, despite carrying the
largest single-site DLA, was not steered.

\begin{longtable}[]{@{}>{\raggedright\arraybackslash}p{0.041\textwidth}>{\raggedright\arraybackslash}p{0.428\textwidth}>{\raggedright\arraybackslash}p{0.366\textwidth}>{\raggedright\arraybackslash}p{0.107\textwidth}@{}}
\toprule
\textbf{Pred.} & \textbf{Preregistered bar (held-out)} &
\textbf{Measured value} & \textbf{Verdict}\tabularnewline
\midrule
\endhead
P1 & Decided late: lens peak \ensuremath{\ge} L30 in \ensuremath{\ge} 40\% det and \ensuremath{\ge} 60\% IBP & det
74/142 = 52.1\% (pass); IBP 39/70 = 55.7\% vs the 60\% bar & NEAR-MISS
(IBP)\tabularnewline
P2 & DLA \textgreater{} 0 at L37 and \textless{} 0 at L36 (det);
\textgreater{} 0 at L37 and L38 (IBP) & det L37 +0.893, L36 \ensuremath{-}0.236; IBP
L37 +0.993, L38 +0.611 & CONFIRMED\tabularnewline
P3 \S{} & Minus-written mean projection \textgreater{} plus-written at L37
(minus\textgreater plus convention \S{}) & minus +0.673, plus \ensuremath{-}0.467 &
CONFIRMED\tabularnewline
P4 & Joint ablation: pooled mean \ensuremath{\le} \ensuremath{-}0.30; \ensuremath{\ge}+2 stratum \ensuremath{\le} \ensuremath{-}1.5; \ensuremath{\ge} 20\%
beat the null & mean \ensuremath{-}0.317; \ensuremath{\ge}+2 stratum (n = 66) \ensuremath{-}2.734; 44/142 =
31.0\% beat the null & CONFIRMED\tabularnewline
P5 & Corrects broken at \ensuremath{\alpha} = 3 in \ensuremath{\ge} 40\% of baseline\textgreater0 cases;
mean \ensuremath{\Delta} lead negative, monotone & 79/107 = 73.8\% broken; mean \ensuremath{-}11.18 &
CONFIRMED\tabularnewline
P6 & Error flips at \ensuremath{\alpha} = 3 \ensuremath{\ge} 20\% (det) / \ensuremath{\ge} 40\% (IBP);
\textbar control\textbar{} \ensuremath{\le} 0.1 & det 46/80 = 57.5\% flips; mean \ensuremath{-}5.54;
control \ensuremath{\approx} 0 \P{} & CONFIRMED\tabularnewline
\bottomrule
\end{longtable}

\textbf{Table I.3 --- Mistral-Small-24B held-out scorecard.} 5
CONFIRMED, 1 NEAR-MISS.

\textbf{I.7 Phi-4-14B (40 layers; habit layers L37, L39)}

\emph{The model registered NOT to steer --- and, as predicted, it did
not.}

Held-out error set: 150 det, 77 IBP. Phi's document makes the inverse
bet: P4 predicts a null or backfire, and P5--P6 predict near-zero flips.
Confirming these predictions is what separates ``localization'' from
``steerability'' --- the dissociation at the center of the paper.

\begin{longtable}[]{@{}>{\raggedright\arraybackslash}p{0.040\textwidth}>{\raggedright\arraybackslash}p{0.483\textwidth}>{\raggedright\arraybackslash}p{0.310\textwidth}>{\raggedright\arraybackslash}p{0.107\textwidth}@{}}
\toprule
\textbf{Pred.} & \textbf{Preregistered bar (held-out)} &
\textbf{Measured value} & \textbf{Verdict}\tabularnewline
\midrule
\endhead
P1 \ddag{} & Decided late: lens peak \ensuremath{\ge} L30 in \ensuremath{\ge} 80\% det and \ensuremath{\ge} 95\% IBP & det
142/150 = 94.7\% (pass); IBP 71/77 = 92.2\% \ddag{} & CONFIRMED
\ddag{}\tabularnewline
P2 & DLA \textgreater{} 0 at L37 and L39, both above controls L15/L25 &
L37 +1.665, L39 +1.156; controls +0.038, +0.105 &
CONFIRMED\tabularnewline
P3 & Single-site ablation at L37/L39 more negative than at the controls
& L37 \ensuremath{-}0.208, L39 \ensuremath{-}0.175 vs L15 +0.117, L25 \ensuremath{-}0.103 &
CONFIRMED\tabularnewline
P4 & Joint ablation null (the unsteerable prediction): mean \ensuremath{\ge} \ensuremath{-}1.0;
\textless{} 50\% beat the null & mean \ensuremath{-}0.192; 7/150 = 4.7\% beat the
null & CONFIRMED\tabularnewline
P5 & Direction subtraction near-zero flips: \ensuremath{\le} 10\% at \ensuremath{\alpha} = 3; random
control 0 flips & L37 0/146 = 0.0\%; L39 4/146 = 2.7\% &
CONFIRMED\tabularnewline
P6 & Steering flips \ensuremath{\le} 10\%; \textbar control\textbar{} \ensuremath{\le} 0.2 with 0
strict flips & 6/145 = 4.1\% flips at \ensuremath{\alpha} = 3; mean \ensuremath{-}8.039; control \ensuremath{\approx} 0 &
CONFIRMED\tabularnewline
\bottomrule
\end{longtable}

\textbf{Table I.4 --- Phi-4-14B held-out scorecard.} 6 CONFIRMED (P1 IBP
near-miss disclosed \ddag{}). Figure 5 reports the same steering run
restricted to minus-written cases (5/122 det, 3/68 IBP) --- hence the
different denominators.

\textbf{I.8 Llama-3.3-70B (80 layers; habit layers L77, L79)}

\emph{The largest model and the cleanest pass: five fully confirmatory
arms, all met.}

Held-out error set: 192 det. Table I.5 scores Llama's det document; the
IBP arms were frozen and adjudicated under their own document (Table
I.1). Five of the six arms here are fully confirmatory and all five
passed; the sixth (P6 steering) was added in Phase 2, is labeled
exploratory \dag{}, and is scored under the frozen disclosure policy rather
than hidden.

\begin{longtable}[]{@{}>{\raggedright\arraybackslash}p{0.038\textwidth}>{\raggedright\arraybackslash}p{0.312\textwidth}>{\raggedright\arraybackslash}p{0.483\textwidth}>{\raggedright\arraybackslash}p{0.107\textwidth}@{}}
\toprule
\textbf{Pred.} & \textbf{Preregistered bar (held-out)} &
\textbf{Measured value} & \textbf{Verdict}\tabularnewline
\midrule
\endhead
P1 & Decided late: lens peak \ensuremath{\ge} L60 in \ensuremath{\ge} 60\% of det cases & 152/192 =
79.2\% & CONFIRMED\tabularnewline
P2 & Pooled mean MLP DLA \textgreater{} 0 at L77 and L79 & L77 +1.067,
L79 +1.717 & CONFIRMED\tabularnewline
P3 & Pairwise separation \ensuremath{\ge} 70\% at L79 & 80.1\% of case pairs
(6,753/8,432); plus-written projection \textless{} 0 in 61/68 = 89.7\%;
means +0.019 (minus) vs \ensuremath{-}0.791 (plus) & CONFIRMED\tabularnewline
P4 & Direction subtraction: mean \ensuremath{\Delta} lead negative at L77 and L79 at every
\ensuremath{\alpha}, monotone & L77 \ensuremath{-}0.437 / \ensuremath{-}0.891 and L79 \ensuremath{-}0.860 / \ensuremath{-}1.723 at \ensuremath{\alpha} = 1/2;
controls L30/L50 \ensuremath{\approx} 0 & CONFIRMED\tabularnewline
P5 & Reverse patching on corrects: mean \ensuremath{\Delta} lead negative at both layers;
broken \ensuremath{\le} 5\% & L77 \ensuremath{-}1.530, L79 \ensuremath{-}3.093; broken 0/130 &
CONFIRMED\tabularnewline
P6 \dag{} & Exploratory \dag{}: mean \ensuremath{\Delta} lead negative, monotone; control \ensuremath{\approx} 0 &
flips 14/32/71 at \ensuremath{\alpha} = 1/2/3 (71/163 = 43.6\%); mean \ensuremath{-}7.15; control \ensuremath{\approx} 0 \P{}
& CONFIRMED \dag{}\tabularnewline
\bottomrule
\end{longtable}

\textbf{Table I.5 --- Llama-3.3-70B held-out scorecard.} 6 CONFIRMED (P6
exploratory \dag{}).

\textbf{I.9 Qwen-2.5-72B (80 layers; habit layers L75, L78)}

\emph{One honest miss: the separation the plan predicted exists, but one
layer higher than frozen.}

Held-out error set: 147 det. One denominator note: the P1 bar counts all
147 held-out errors, exactly as the preregistration specifies; \S{}3 quotes
a higher late-peak share because it uses only lens-confirmed cases --- a
different, disclosed quantity, not a revision of this one.

\begin{longtable}[]{@{}>{\raggedright\arraybackslash}p{0.038\textwidth}>{\raggedright\arraybackslash}p{0.328\textwidth}>{\raggedright\arraybackslash}p{0.466\textwidth}>{\raggedright\arraybackslash}p{0.107\textwidth}@{}}
\toprule
\textbf{Pred.} & \textbf{Preregistered bar (held-out)} &
\textbf{Measured value} & \textbf{Verdict}\tabularnewline
\midrule
\endhead
P1 & Decided late: lens peak \ensuremath{\ge} L60 in \ensuremath{\ge} 60\% of det cases (all held-out
errors) & 90/147 = 61.2\% & CONFIRMED\tabularnewline
P2 & Mean MLP DLA \textgreater{} 0 at L75 and L78 & L75 +0.472, L78
+0.568 & CONFIRMED\tabularnewline
P3 & Minus-written mean projection \textless{} plus-written at L75 and
L78 & ordering reversed at L75/L78; the separation holds only at L79
(77.9\% of case pairs), not at the preregistered layers &
MISS\tabularnewline
P4 & Direction subtraction: L78 mean \ensuremath{\Delta} lead negative and monotone at
every \ensuremath{\alpha} & minus-written: \ensuremath{-}0.773 / \ensuremath{-}1.605 at \ensuremath{\alpha} = 0.5/1; monotone; random
control \ensuremath{\approx} 0 & CONFIRMED\tabularnewline
P5 & Judge gap at L79: minus-correct \textgreater{} 0 and plus-correct
\textless{} 0 in the majority; broken \ensuremath{\le} 3\% & minus-correct 123/123
positive (mean +1.606); plus-correct 18/18 negative (mean \ensuremath{-}1.135);
broken 1/123 & CONFIRMED\tabularnewline
P6 & Steering: mean \ensuremath{\Delta} lead negative at every \ensuremath{\alpha}, monotone; control \ensuremath{\approx} 0 &
minus-written baseline\textgreater0 (n = 25): 6/25 = 24.0\% flips at \ensuremath{\alpha} =
3; mean \ensuremath{-}8.143; control \ensuremath{\approx} 0 \P{} & CONFIRMED\tabularnewline
\bottomrule
\end{longtable}

\textbf{Table I.6 --- Qwen-2.5-72B held-out scorecard.} 5 CONFIRMED, 1
MISS. The P3 miss is scored as frozen and kept: the separation exists,
but at L79 rather than the preregistered L75/L78 --- a finding about
redundancy, reported as a miss because that is what the frozen plan says
it is.

\textbf{I.10 What it adds up to}

Across the twenty-nine scored cells: Gemma 1 CONFIRMED / 4 NEAR-MISS / 1
NOT SCORED; Mistral 5 CONFIRMED / 1 NEAR-MISS; Phi 6 CONFIRMED; Llama 6
CONFIRMED (P6 exploratory); Qwen 5 CONFIRMED / 1 MISS. That is 23 of 29
frozen predictions confirmed --- the count on Figure 6. The six cells
that fall short are all reported: four of them come from one model
(Gemma) being scored at its frozen, suboptimal sites, and the one
outright miss (Qwen P3) is kept on the board rather than rescored.

Data. Every measured value in Tables I.2--I.6 comes from the model's
held-out stage folders in the released repository (the s01--s09 stages
of Appendix B) and is cross-checked against the master results ledger;
the preregistration documents, with their commit history, are in the
repository's preregistration\_\textless model\textgreater/ folders.

\clearpage
\section*{Appendix J --- The Reverse Arrow: Patching and Injection (P5)}
\addcontentsline{toc}{section}{Appendix J --- The Reverse Arrow: Patching and Injection (P5)}

\emph{Every causal result in the main text removes the sign signal from
errors. This appendix presents the experiments that run the arrow the
other way --- copying or injecting the error signal into correct answers
--- and the forward transplant that shows why a correct state cannot be
copied in to repair. Verdicts were frozen per model in Appendix I; this
appendix gives the protocols, the full numbers, and the controls.}

\textbf{J.1 What this appendix shows}

The model is about to write a sign. Every experiment here freezes it at
that moment --- one position, the token just before the sign appears. At
that position each layer produces its message: the activation packet it
hands forward. Patching (stage s07, Appendix B) swaps that message for a
different one --- one component, one token, nothing else --- and lets
the run continue to see which sign is written now. The swapped-in
message is always a mean: the healthy message, averaged from correct
runs, or the error message, averaged from late-circuit error runs. Three
questions organize the appendix. Can the healthy message repair an
error? No --- and the null is preregistered: a stamp layer says the same
thing on correct and error runs, so the healthy and the error message
coincide there and swapping changes nothing (J.4). Can the error message
corrupt a correct answer? Yes, in each model's registered way ---
including the model where the lead moves and oversized margins absorb
it, and the model immune in both directions (J.3). Does any of it
survive controls? The same swap at mid-network layers is null, and
sign-matched donors close the composition confound (J.3--J.4).
Prediction P5 --- \emph{the sign signal also moves correct answers in
the registered way} --- is scored from these arms in Appendix I.

\textbf{J.2 Method}

All arms are teacher-forced at the probe token --- the position just
before the sign. The patched object is one component's output (an MLP,
or L77's attention out-projection in the Llama forward arm) at that
position only, replaced by a donor-set mean; error-mean donors pool
late-circuit error cases (lens peak \textgreater{} layer 5) at the same
position. Readout: the written sign's logit lead at the probe token; \ensuremath{\Delta}
vs the unpatched run. Broken is strict --- correct at baseline, wrong
sign ahead after the patch; fixed is the mirror. The injection arms that
are not literal patches use the same readout: Mistral and Phi apply the
\S{}5 subtraction edit to correct answers; Qwen's judge arm patches
error-mean vs self-mean at the check layer and scores the per-case gap.
Controls throughout: the identical swap at mid-network layers (L30/L50
or per-model equivalents) and, for steering-style arms, CTRL\_MAG.

\textbf{J.3 The reverse arrow, model by model}

Table J.1 gives one row per model: the registered instrument, the frozen
pass bar and held-out result (both verbatim from the Appendix I
scorecard), and the verdict. The subsections add the full per-layer
numbers and controls.

\begin{longtable}[]{@{}>{\raggedright\arraybackslash}p{0.037\textwidth}>{\raggedright\arraybackslash}p{0.191\textwidth}>{\raggedright\arraybackslash}p{0.265\textwidth}>{\raggedright\arraybackslash}p{0.323\textwidth}>{\raggedright\arraybackslash}p{0.107\textwidth}@{}}
\toprule
\textbf{Model} & \textbf{Registered instrument} & \textbf{Frozen pass
bar (verbatim)} & \textbf{Held-out result (frozen)} &
\textbf{Verdict}\tabularnewline
\midrule
\endhead
Mistral & C2 subtraction on corrects at {[}L38, L39{]}, \ensuremath{\alpha} grid &
Corrects broken at \ensuremath{\alpha} = 3 in \ensuremath{\ge} 40\% of baseline\textgreater0 cases; mean
\ensuremath{\Delta} lead negative, monotone & 79/107 = 73.8\% broken; mean \ensuremath{-}11.18 &
CONFIRMED\tabularnewline
Llama & s07 error-mean patch into corrects at habit layers & Reverse
patching on corrects: mean \ensuremath{\Delta} lead negative at both layers; broken \ensuremath{\le} 5\%
& L77 \ensuremath{-}1.530, L79 \ensuremath{-}3.093; broken 0/130 & CONFIRMED\tabularnewline
Qwen & Judge arm: error-mean vs self-mean patch at check layer L79 &
Judge gap at L79: minus-correct \textgreater{} 0 and plus-correct
\textless{} 0 in the majority; broken \ensuremath{\le} 3\% & minus-correct 123/123
positive (mean +1.606); plus-correct 18/18 negative (mean \ensuremath{-}1.135);
broken 1/123 & CONFIRMED\tabularnewline
Phi & Direction subtraction, single layers L37 / L39 (inverse bet) &
Direction subtraction near-zero flips: \ensuremath{\le} 10\% at \ensuremath{\alpha} = 3; random control 0
flips & L37 0/146 = 0.0\%; L39 4/146 = 2.7\% & CONFIRMED\tabularnewline
Gemma & --- & No pass bar (descriptive by pre-measurement amendment) &
reported descriptively only & NOT SCORED\tabularnewline
\bottomrule
\end{longtable}

\textbf{Table J.1:} The reverse arrow, scored. Pass bars and results are
quoted verbatim from the frozen scorecard (Appendix I); protocols and
full per-layer numbers in J.3.1--J.3.5.

\textbf{J.3.1 Mistral --- corrects break wholesale.} The same two-layer
edit that fixes its errors, pointed at corrects (held-out determinant, n
= 122): 79 broken at \ensuremath{\alpha} = 3 --- 73.8\% of the 107 cases with positive
baseline lead --- with mean lead shift \ensuremath{-}11.18 on that subset (\ensuremath{-}8.85 over
all 122); CTRL\_MAG breaks 0/122 (mean +0.07). The IBP correct battery,
outside the frozen bar, is even starker: 133/156 broken at \ensuremath{\alpha} = 3, mean
\ensuremath{-}14.09. The model that steers most is also the least protected --- no
margin, no rescue.

\textbf{J.3.2 Llama --- the signal arrives, the margins absorb it.} The
literal transplant (s07 reverse patch): the error-set mean message
copied into correct runs at the habit layers. Determinant (n = 130): L77
\ensuremath{-}1.53, L79 \ensuremath{-}3.09, broken 0/130. IBP (n = 47): L77 \ensuremath{-}3.58, L78 \ensuremath{-}7.58,
broken 0/47. Control L30: \ensuremath{-}0.25 (det) and \ensuremath{-}0.10 (IBP), 0 broken. The
lead moves at every habit site and not one answer breaks --- the same
oversized-margin behavior its error steering shows.

\textbf{J.3.3 Qwen --- the judge reads the injected signal; the stamps
ignore it.} Registered arm (held-out determinant corrects, n = 141 = 123
minus + 18 plus): error-mean vs self-mean patch at check layer L79; the
per-case gap is positive for all 123 minus-corrects (mean +1.61),
negative for all 18 plus-corrects (\ensuremath{-}1.14), with 1/123 broken --- the
judge sees error content without flipping the answer. The unregistered
companion is just as telling: the same error-mean message patched at the
stamp sites themselves does nothing (L75 \ensuremath{-}0.29, L78 +0.04, 0/141 broken;
control L30 +0.01, 0 broken) --- a stamp's message is the same on error
and correct runs, so swapping it in is a no-op. Fixedness, demonstrated
from the reverse side.

\textbf{J.3.4 Phi --- immune in both directions.} The registered P5 arm
is the inverse bet, scored on the held-out determinant error battery (n
= 146): direction subtraction fixes almost nothing --- L37 0/146
flipped, L39 4/146 (2.7\%) at \ensuremath{\alpha} = 3, while the lead moves by \ensuremath{-}3.08 and
\ensuremath{-}3.55. The reverse arrow completes the picture: the same subtraction
pointed at corrects breaks at most 1/140 (determinant, either layer;
means \ensuremath{-}3.04 / \ensuremath{-}3.43 at \ensuremath{\alpha} = 3) and 0/146 on IBP at both layers (L39 mean
\ensuremath{-}10.17). In every arm the lead moves by 3--10 logits and the written
sign almost never follows. Phi's two-way immunity is what rules out a
miscalibrated dose or direction: a wrong lever fails one way, not both.

\textbf{J.3.5 Gemma --- descriptive.} The pass bar was withdrawn before
measurement (Appendix I); the correct-cohort numbers are reported in
Table H.4 and carried here without a verdict.

\textbf{J.4 The forward arrow: why a correct state cannot be copied in}

The forward transplant patches the correct-set mean message into error
runs. At stamp sites the preregistered prediction is \emph{null} --- the
message there carries nothing error-specific --- and it holds. Where the
pooled donor mean did move the lead (Llama L79: \ensuremath{-}1.81, 27/189 flipped),
the sign-matched control unmasks it as donor composition: with donors
matched to the written sign, L79 goes to +1.00 with 8/189 fixed, and L77
to +2.27 with 0/189 fixed. Qwen's forward numbers are small everywhere
(held-out n = 99: L75 \ensuremath{-}0.67 with 3 flips, L78 \ensuremath{-}0.69 with 1, L79 \ensuremath{-}0.44
with 2; controls L30 \ensuremath{-}0.01 with 0, L50 \ensuremath{-}0.08 with 1). Across both models
with a patching arm, the clean forward transplant repairs essentially
nothing.

\textbf{J.5 What it means}

One triple summarizes every arm: \emph{removing the direction fixes;
transplanting a correct state does not; injecting the error state
breaks} --- the signal can be written in, not copied out. Repair
requires deleting one axis --- not overwriting the whole message, which
at the decisive sites contains nothing error-specific to overwrite. Each
model's reverse-arrow behavior independently re-derives its \S{}6 mechanism
verdict --- Mistral's ungated push, Llama's margins, Qwen's stamp/judge
split, Phi's mechanism-level immunity --- and the direction-specific,
control-clean pattern doubles as specificity evidence for the whole
battery.

\textbf{J.6 Provenance and verification}

\textbf{Source files.} Reverse patch (J.3.2, J.3.3 companion):
\{Model\}/heldout\_*/s07\_patching/*\_correct\_reverse\_patch.json.
Judge arm (J.3.3): Qwen heldout s09\_judge
det\_4x4\_correct\_judge\_candidates\_armA. Subtraction on corrects
(J.3.1, J.3.4): s08\_c2\_steering *\_c2\_on\_correct.json (Mistral) and
s06\_direction\_subtraction
*\_correct\_deepsign\_expG\_prime\_bias*.json (Phi); Phi's registered
error-battery arm from s06
det\_4x4\_error\_expG\_prime\_bias\_subtraction. Forward arm (J.4):
s07\_patching *\_error\_correct\_mean\_patch.json and
*\_error\_signmatched\_patch.json.

\textbf{Criterion.} Broken and fixed are strict per-case flips of the
written sign's logit lead at the probe token (baseline correct \ensuremath{\rightarrow} edit
wrong, and the mirror); no partial credit for lead movement.

\textbf{Verification.} Every count and mean in this appendix was
recomputed from the raw per-case JSONs and cross-checked against
paper/appendix/ledger\_results.csv (stages s06, s07, s08, s09); the
Mistral 107-case denominator (baseline lead \textgreater{} 0) and the
79/107 = 73.8\% broken rate were re-derived directly from the raw
held-out file. Frozen bars and verdicts are quoted verbatim from
Appendix I and were not rescored.

\clearpage

\section*{Appendix K --- Phi-4: why the best-localized circuit resists calibrated steering}
\addcontentsline{toc}{section}{Appendix K --- Phi-4: why the best-localized circuit resists calibrated steering}
\label{app:phi_deepdive}

Everything in this appendix is exploratory and uses discovery data only; no
preregistered verdict (Appendix~I) depends on any number here.

Phi poses the sharpest version of the paper's question. It replicates every
descriptive claim --- the latest peaks of the five models, L37 stamp-like and L39
amplifier-like, MLP credit dominating attention --- yet the standard steering
battery at $\alpha \le 3$ flips almost nothing, even though the wrong sign's lead
clearly moves under the edit. The lever transmits force; the outcome does not
change. Is that an artifact of dose, direction, or sites --- or a property of the
model? The experiments below locate the answer in two measured properties:
\emph{margins} (K.1) and \emph{redundancy} (K.2).

\textbf{K.1\quad Over-determined margins.} Figure~K1a shows the baseline
wrong-sign lead at the error site for every discovery batch in the study. Phi is
in a different regime from every other model: determinant median 17.8 logits
(max 43.4), and IBP clustered in a tight 28--32 band --- 29 of 40 cases sit at
$\ge$28, a structural floor. The other batches sit at medians of roughly 5.5--7.5
logits (Gemma, Llama, Qwen; Qwen's small IBP batch, $n=10$, is higher at 17.3
but still below Phi's floor) and 0.5--3 logits for Mistral-Small. Figure~K1b
shows what the intervention does to those margins: at the strongest
standard-dose arm the edited lead tracks $\approx$0.58$\times$ baseline (det) and
$\approx$0.51$\times$ (IBP), with corr(baseline, $\Delta$lead) $=-0.90$ and
$-0.98$. The edit is a \emph{proportional} cut, not an absolute shift --- and a
$\sim$50\% cut never crosses zero for a baseline above a few logits. Accordingly,
the only two cases that flip at standard dose are the two smallest margins in the
pool (baselines 3.5 and 3.75, circled in the figure).

\begin{center}\includegraphics[width=\linewidth]{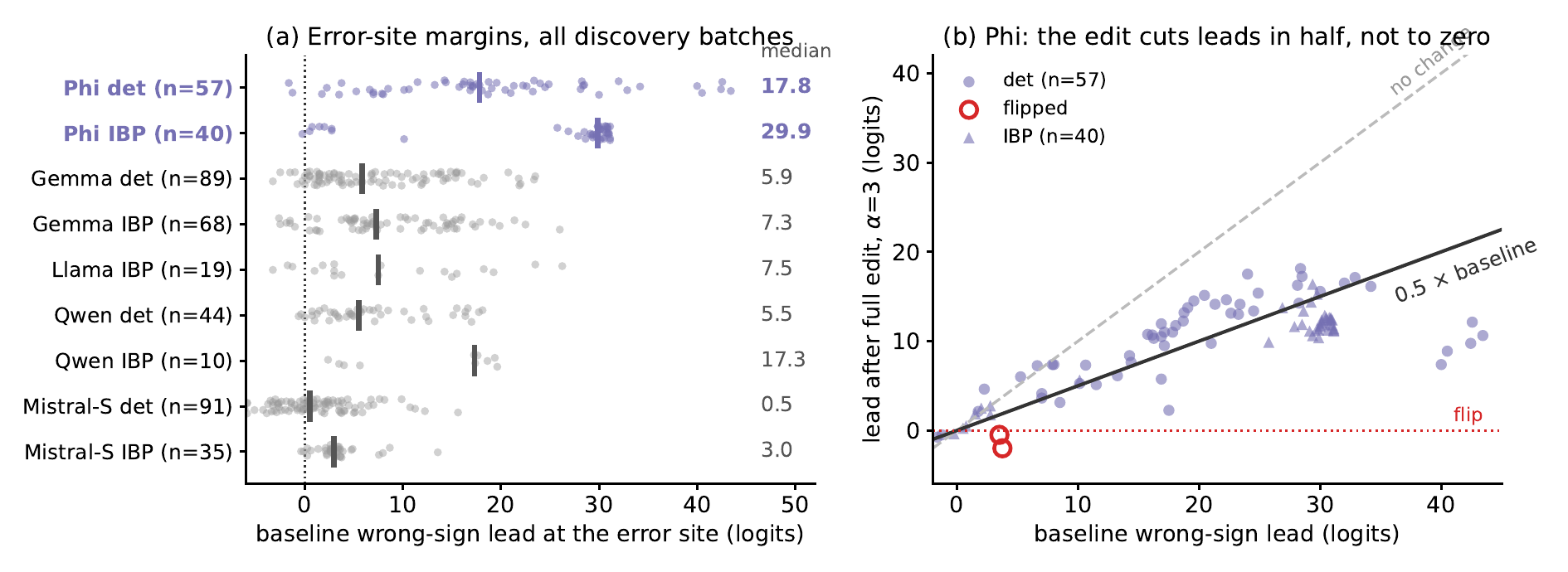}\end{center}

\textbf{Figure K1:} Margins, not location, separate Phi from the steerable
models. \textbf{(a)} Per-case baseline wrong-sign lead at the error site for
all discovery batches (dots: cases; bar and right column: median). Phi's
determinant margins are $2$--$3\times$ any other model's, and its IBP margins
form a 28--32 logit floor. \textbf{(b)} Phi, lead after the full edit at
$\alpha=3$ versus baseline lead. Cases follow the 0.5$\times$-baseline line
(solid), not the flip region below zero (dotted); the only flips (red circles)
are the two smallest baselines, 3.5 and 3.75.

\textbf{K.2\quad Distributed, redundant writing.} The margins are maintained by
an ensemble, not a bottleneck. Figure~K2a shows the per-layer direct logit
attribution to the written sign: fifteen MLP layers (det; fourteen in IBP)
across L24--39 contribute positive credit, and the two habit layers targeted by
the steering battery, L37 and L39, carry only 19\% (det) and 26\% (IBP) of the
total positive MLP credit; attention carries a further $\sim$22--29\%. Four
interventions probe the ensemble directly. Wide-window mean ablation (MLPs
L30--39, then L24--39) moves the det lead by only $-0.67$ and $-1.28$ --- at or
below the 5th percentile of a matched random 10-layer-window null ($-2.18$).
The cumulative staircase in Figure~K2b ablates the top-$k$ DLA layers for
$k=1\ldots15$ and stays essentially flat: $k=15$ reaches a mean of $-1.34$
logits (4 flips) against a median det margin of 17.8, putting the minimum
hitting set beyond 15 layers. \textbf{Even with all fifteen credited layers
removed at once, Phi's wrong sign does not ablate.} Joint MLP$+$attention ablation at L36--39 flips
2/58 det and 0/39 IBP (cases with a positive baseline lead). And the single
most-credited attention head, L36H33, is a verified null (0 flips, mean
$\Delta$lead $+0.005$ det / $-0.038$ IBP). No small subset of the ensemble is a
bottleneck.

\begin{center}\includegraphics[width=\linewidth]{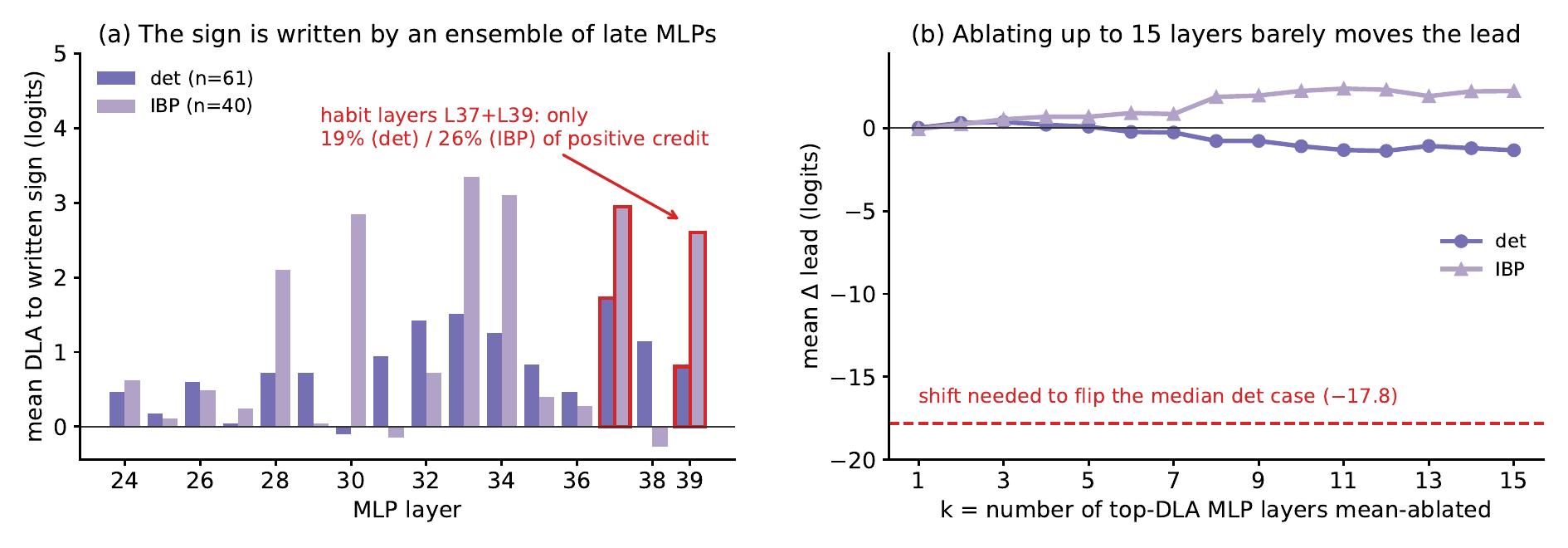}\end{center}

\textbf{Figure K2:} The sign is written by a redundant ensemble. \textbf{(a)}
Mean per-layer MLP direct logit attribution to the written sign, L24--39. The
habit layers L37/L39 (red outline) carry only about a fifth to a quarter of the
positive credit. \textbf{(b)} Cumulative staircase: mean change in the
wrong-sign lead after mean-ablating the top-$k$ DLA layers. The curve is flat
in det and slightly \emph{positive} in IBP (ablations backfire; see K.4) ---
nowhere near the $-17.8$ needed to flip the median det case (dashed).

\textbf{K.3\quad What it takes to flip Phi.} Phi is not dose-immune --- it is
dose-expensive. A single-layer sweep at $\alpha \in \{5, 8, 12\}$ shows that at
$\alpha=12$, L37 flips 12/57 det errors (mean $\Delta$lead $-15.0$) and L39
flips 29/40 IBP errors (mean $-31.3$). Three observations make this
overpowering rather than control. First, diminishing returns: the delivered
lead reduction at high dose is only about half to two-thirds of what a linear
extrapolation from $\alpha=1$ predicts. Second, the IBP push ($-31.3$ mean)
exceeds every baseline in the pool (max 31.1) --- a blanket shove past the
margin, not a targeted correction. Third, specificity survives saturation:
CTRL\_RAND is $\approx$0 at every dose, the mid-network control site L15 is
null at every dose, and L25 contributes at most 1--2 flips even at $\alpha \ge
8$. The sharpest test doses each case individually, $\alpha_i =
\mathrm{baseline}_i / |\mathrm{slope}_i|$ (capped at 15), using each case's own
measured dose response --- information no deployment-time intervention would
have. Even then, joint L37$+$L39 subtraction flips 5/10 det and 2/6 IBP cases;
L39 alone flips 3/10 and 0/6; CTRL\_RAND flips none. This is the precise sense
in which Phi is unsteerable: \emph{unsteerable at calibrated dose}.

\textbf{K.4\quad Checks.} Two families of checks constrain the interpretation.
\emph{The IBP anomaly.} In the IBP domain the wide-window ablations
\emph{worsen} the errors ($+1.79$ / $+2.02$), the staircase is positive
throughout (Figure~K2b), and patching clean late-window activations into error
runs \emph{strengthens} the wrong sign ($+5.44$ for L32--39, over the 39 cases
with a positive baseline lead). Together with the 28--32 floor, this is
consistent with IBP error cases sitting near a saturated attractor: perturbing
the direct path releases opposing signals. We flag it as a target for dedicated
follow-up rather than interpreting it further. \emph{Ruled out.} Pipeline
artifacts (CTRL\_RAND, CTRL\_MAG, and the control layers are clean or
near-clean nulls in every arm; teacher-forcing verified); tokenizer-variant
mismatch (Phi's BPE has two minus tokens, but the error cohorts are 100\% the
standard leading-space variant, and stratifying the correct-side cases makes
the measured lever \emph{stronger}, not weaker); and model scale
(Mistral-Small, half Phi's depth budget, flips 37/91 det at the same standard
dose $\alpha=3$).

\textbf{K.5\quad What this adds to the dissociation.} Phi's circuit is
correctly located --- the same layers, roles, and timing as the models that
steer --- but the decision it writes is over-determined twice over: margins
several times larger than any other model's (Figure~K1), written by an ensemble
too wide for low-rank surgery at the named sites (Figure~K2). Both quantities
are invisible in a localization map and measurable only by intervening. High
dose can overpower the margin and per-case calibration reaches about half the
cases, so the direction is causally real; what fails is precisely the inference
from \emph{where} to \emph{how hard}. Localization names the site; margin and
redundancy set the price of using it.

Every number in this appendix was recomputed from the per-case result files in
the repository (Phi discovery directories, stages s02--s08, plus the
other models' steering files for Figure~K1a); the joint MLP$+$attention
ablation, clean-window patching, and tokenizer stratification results are in
the repository in the same schema.

\end{document}